\documentclass[acmsmall,screen,authorversion,nonacm]{acmart}

\usepackage{float}
\usepackage{dsfont}
\usepackage{multirow}
\usepackage{wrapfig}
\usepackage{rotating}
\usepackage{enumitem}
\usepackage{comment}
\usepackage{adjustbox}
\usepackage{amsmath}
\usepackage{mathtools}
\usepackage{pifont}
\usepackage{xcolor}
\usepackage{physics}
\usepackage{colortbl}
\usepackage{xcolor}

\usepackage[flushleft]{threeparttable}
\newtheorem{definition}{Definition}
\newtheorem{problem}{Problem}
\newcommand{\cmark}{{\color{green}\ding{51}}}
\newcommand{\xmark}{{\color{red}\ding{55}}}

\AtBeginDocument{%
  }

\setcopyright{acmcopyright}
\copyrightyear{2018}
\acmYear{2018}
\acmDOI{XXXXXXX.XXXXXXX}

\acmConference[Conference acronym 'XX]{Make sure to enter the correct
  conference title from your rights confirmation email}{June 03--05,
  2018}{Woodstock, NY}
\acmPrice{15.00}
\acmISBN{978-1-4503-XXXX-X/18/06}

\begin{document}


\title[Robust Computer Vision in an Ever-Changing World]{Robust Computer Vision in an Ever-Changing World: \\A Survey of Techniques for Tackling Distribution Shifts}

\author{Eashan Adhikarla}
\authornote{Includes efforts while working at Lenovo Research.}
\email{eaa418@lehigh.edu}
\affiliation{%
  \institution{Department of Computer Science and Engineering, Lehigh University}
  \streetaddress{113 Research Drive}
  \city{Bethlehem}
  \state{Pennsylvania}
  \country{USA}
  \postcode{18015}
}

\author{Kai Zhang}
\affiliation{%
  \institution{Department of Computer Science and Engineering, Lehigh University}
  \streetaddress{113 Research Drive}
  \city{Bethlehem}
  \country{USA}
\email{kaz321@lehigh.edu}
}

\author{Jun Yu}
\affiliation{%
  \institution{Department of Computer Science and Engineering, Lehigh University}
  \streetaddress{113 Research Drive}
  \city{Bethlehem}
  \state{Pennsylvania}
  \country{USA}
  \email{juy220@lehigh.edu}
}

\author{Lichao Sun}
\affiliation{%
  \institution{Department of Computer Science and Engineering, Lehigh University}
  \streetaddress{113 Research Drive}
  \city{Bethlehem}
  \state{Pennsylvania}
  \country{USA}
  \email{lis221@lehigh.edu}
}

\author{John Nicholson}
\affiliation{%
  \institution{Lenovo Research}
  \streetaddress{}
  \city{Raleigh}
  \state{North Carolina}
  \country{USA}
  \email{jnichol@lenovo.com}
}

\author{Brian D. Davison}
\authornote{Corresponding author}
\affiliation{%
  \institution{Department of Computer Science and Engineering, Lehigh University}
  \streetaddress{113 Research Drive}
  \city{Bethlehem}
  \state{Pennsylvania}
  \country{USA}
  \email{bdd3@lehigh.edu}
}

\renewcommand{\shortauthors}{Adhikarla et al.}

\begin{abstract}
    AI applications are becoming increasingly visible to the general public. There is a notable gap between the theoretical assumptions researchers make about computer vision models and the reality those models face when deployed in the real world. One of the critical reasons for this gap is a challenging problem known as distribution shift. Distribution shifts tend to vary with complexity of the data, dataset size, and application type. In our paper, we discuss the identification of such a prominent gap, exploring the concept of distribution shift and its critical significance. We provide an in-depth overview of various types of distribution shifts, elucidate their distinctions, and explore techniques within the realm of the data-centric domain employed to address them. Distribution shifts can occur during every phase of the machine learning pipeline, from the data collection stage to the stage of training a machine learning model to the stage of final model deployment. As a result, it raises concerns about the overall robustness of the machine learning techniques for computer vision applications that are deployed publicly for consumers. Different deep learning models each tailored for specific type of data and tasks, architectural pipelines; highlighting how variations in data preprocessing and feature extraction can impact robustness., data augmentation strategies (e.g. geometric, synthetic and learning-based); demonstrating their role in enhancing model generalization, and training mechanisms (e.g. transfer learning, zero-shot) fall under the umbrella of data-centric methods. Each of these components form an integral part of the neural-network we analyze contributing uniquely to strengthening model robustness against distribution shifts. We compare and contrast numerous AI models that are built for mitigating shifts in hidden stratification and spurious correlations, adversarial attack shift, and unseen data shifts. Overall, we summarize the innovations and major contributions in the literature, give new perspectives toward robustness, and highlight the limitations of those proposed ideas. We also identify loopholes in other surveys, and propose potential future long- and short-term work directions.
\end{abstract}




\maketitle

\begin{figure}[hbtp] 
    \centering
    \includegraphics[width=\textwidth, angle=0]{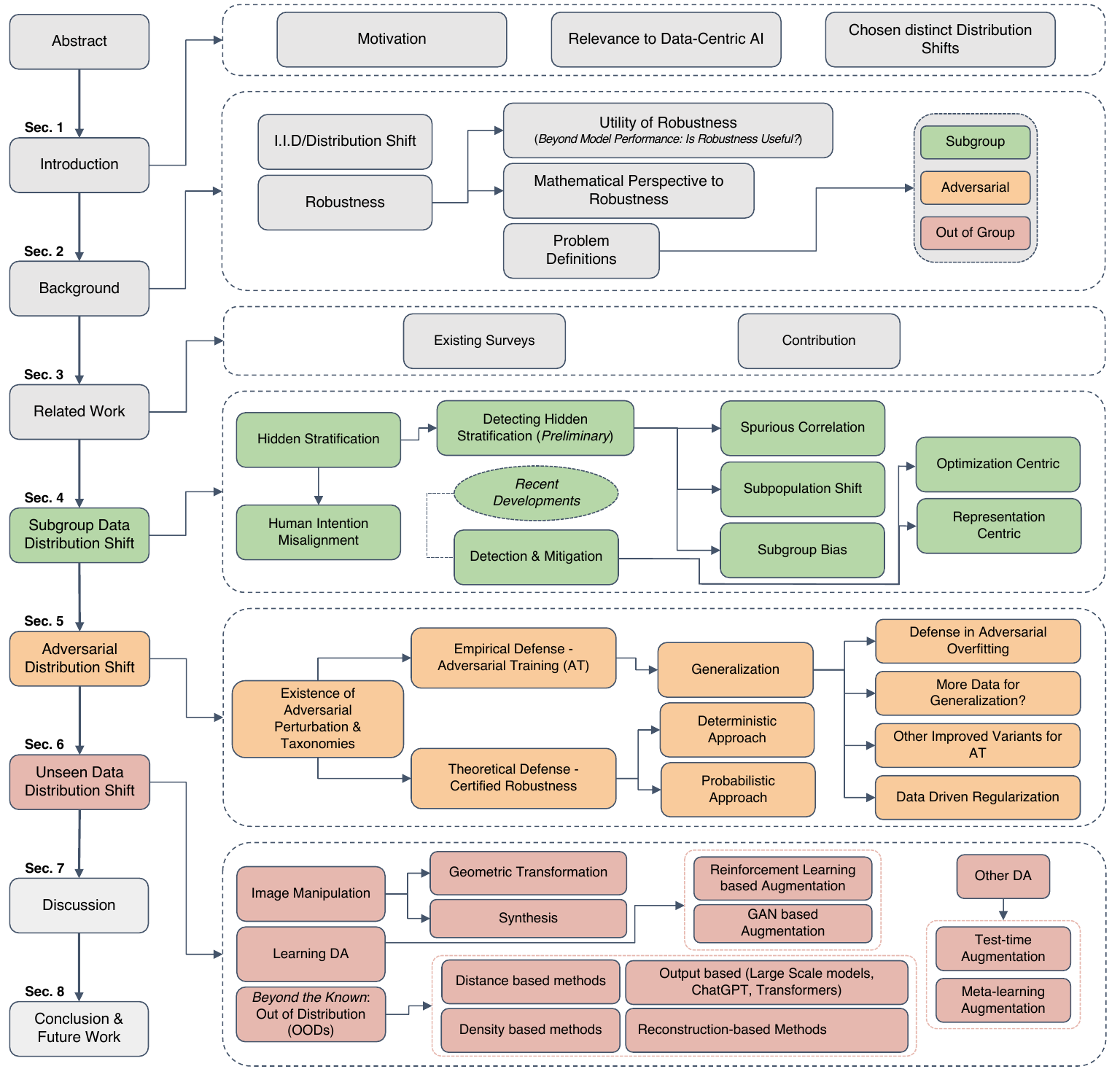}
    \caption{\label{fig:content-summary.png}\textit{Presenting a comprehensive tree diagram showcasing the hierarchical structure of the paper for clear navigation of content.}}
\end{figure}
\section{Introduction}

    {\em Distribution shift} is a widely known problem in the data science community. This common problem arises when the data distribution of an algorithm’s training dataset differs from the distribution it encounters when deployed. Traditionally, we deploy machine learning models in the real world under the assumption that the dataset distributions from testing are very close to that of training data. This assumption rarely holds in real-world scenarios making it risky. While it may appear easier 
    to achieve high accuracy on a test dataset that mirrors that training data, but does not necessarily translate to effective performance in diverse, real-world conditions. However, the problem of data distribution shift becomes more apparent when tested over real-world datasets after deployment. Data distribution shifts can arise due to numerous reasons such as changes in environmental conditions, evolving user behaviors, or alterations in data recording processes. Datasets become more prone to the problem of distribution shift in such several cases. 
    One common scenario is when collecting data in a dynamic environment where the data features change over time/space, e.g., lighting conditions in images, user interaction patterns, sensor/hardware characteristics change over time. 

    \begin{figure}[!ht]
        \centering
        \includegraphics[width=\textwidth]{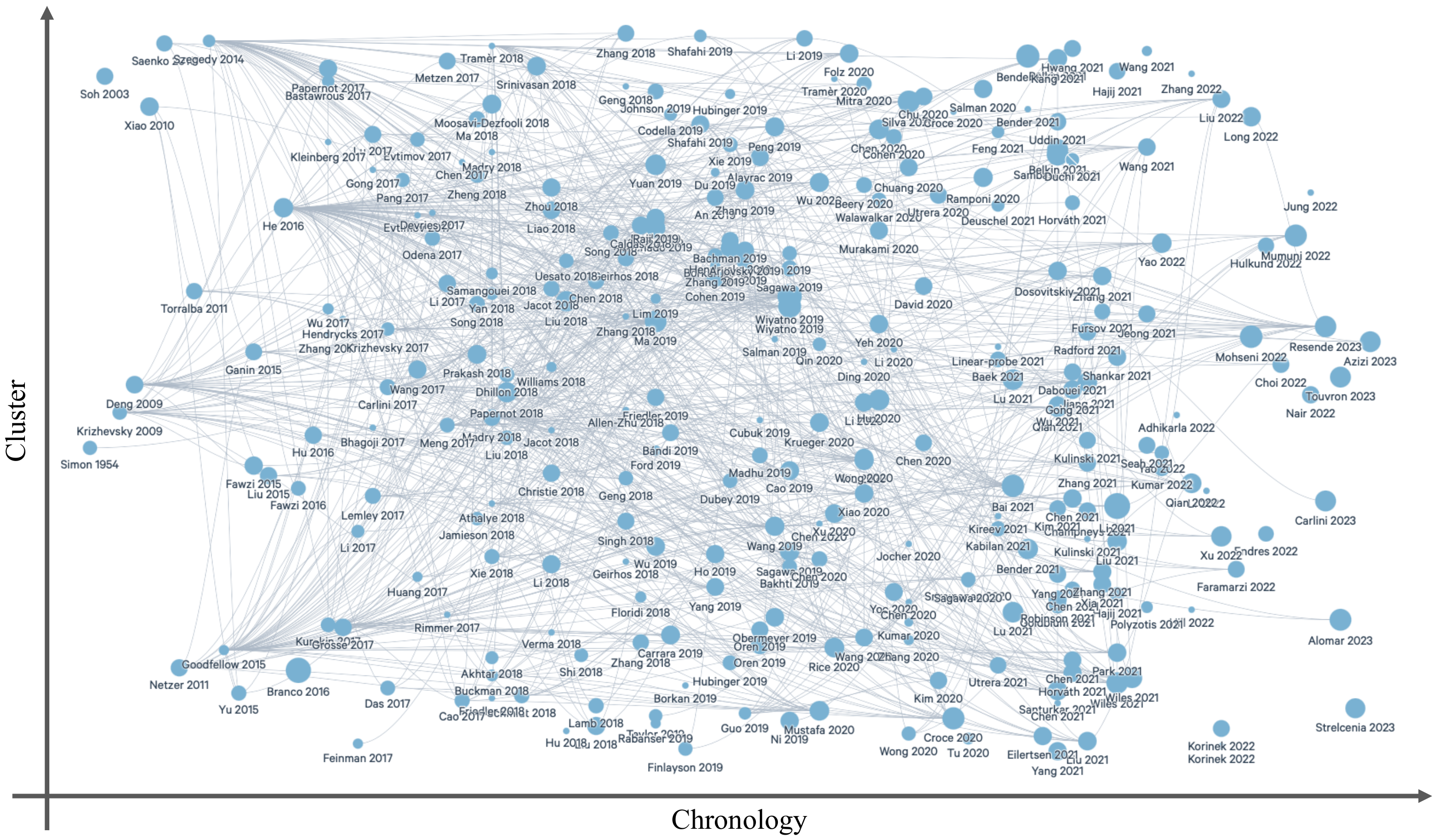}
        \begin{flushleft}
            \scalebox{0.8}{\footnotesize This diagram is drawn using Litmaps\footnote{https://app.litmaps.com} tool, a platform that helps to visualize how the research connects.}
        \end{flushleft}
        \caption{\label{fig:citation-graph}\textit{The figure represents the citation graph of the articles used in this depth study. Each blue node represents a research article, and the size of the blue node directly corresponds to the number of citations. The nodes are chronologically ordered from left to right. The links between the nodes show the connection, demonstrating evidence that papers from different categories relate to one another and this report further highlights those connections and gives new perspective.}}
    \end{figure}

    Such an environment hides the actual distribution of data, causing only a subset of the distribution to be active at any given time. For instance, concept shift (or concept drift) occurs in predictive models where relationships between variables change over time \cite{lu2018learning, gama2014survey}. If the impacted variables are performance-relevant for a machine learning model, data distribution shift may become an issue. For example, consider stock market prediction, where the market can suddenly (and sometimes permanently) shift due to factors that the model may have never encountered. Such environmental changes could cause a non-robust stock market prediction model to suffer significant drops in performance. A second common scenario of a data distribution shift would be to train a model in a controlled environment, such as a synthetic dataset. 
    Similar challenges arise in training models in uncontrolled environments. A prime example is training self-driving car AI models on real-world data \cite{nvidia-self-driving,teng2023motion}. These models may be vulnerable to distribution shifts, especially if the training environment does not include a sufficient range of `sub-environments', such as night driving examples, or varying traffic conditions. Conversely, a second common scenario of a data distribution shift occurs when training a model in a more controlled setting, such as with synthetic datasets. The gap between synthetic and real-world data often causes major performance issues for models that initially performed well during testing.

    Certainly, the overall problem of data distribution shift is wide and holds numerous challenges depending on the type of distribution shift that has taken place. The problem raises many sub-questions. For example, if the model achieves $99$\% accuracy in the test data set, would it be robust to a real-world environment? Did it learn all the subsets of the dataset appropriately, including subsets the dataset authors may not have considered, such as different lightning conditions in the case of object recognition? This question becomes especially pertinent in scenarios where the real-world application involves more diversse conditions than the ones represented in the training and validation datasets. 

    In the past few years, it has become evident that artificial intelligence with big data has benefited a lot from machine learning, specifically deep learning techniques leading to state-of-the-art (SOTA) results in various tasks; computer vision \cite{ViT, SwinViT, dalle2}, natural language processing \cite{clip, llama}, image generative from text (ChatGPT\footnote{A publicly accessible tool developed by openAI (\href{https://chat.openai.com}{https://chat.openai.com})}, GPT-4\footnote{\href{https://openai.com/gpt-4}{https://openai.com/gpt-4}}, Bard\footnote{\href{bard.google.com}{bard.google.com}}, Bing Chat\footnote{\href{https://www.bing.com/search?q=Bing+AI\&showconv=1}{https://www.bing.com/search?q=Bing+AI\&showconv=1}}), medicine \cite{zhang2023biomedgpt,chen2021synthetic,kather2022medical}. 

    The data science community widely refers to data as the `food' for machine learning models, but if data is the 'food', then distribution shift acts like a `poison.' Data distribution shift is widely argued to be addressable via a data-centric paradigm \cite{hazylab}. \citeauthor{dcaiNot} describes this approach builds on re-examining traditional techniques in light of modern AI models, emphasizing the manipulation of data over inventing new models due to established techniques in data reasoning. It prioritizes domain knowledge encoded in data management, setting up processes for monitoring and improving data quality. Central to data-centric AI is the development of theory and norms around data management for AI systems, focusing on foundational areas like weak supervision, data augmentation, and understanding representations. These concepts are crucial for analyzing and addressing data distribution shifts in a variety of domains. While growing data volumes can lead to more generalized models, they also introduce complexities in handling multiple kinds of distribution shifts simultaneously, particularly in real-world scenarios with dynamic and diverse data characteristics.

    Under this nascent data-centric paradigm, there seems a significant potential for training models that are robust to distribution shifts. A major challenge in this problem is that current datasets are insufficiently diverse, thus making it difficult to select a strategy to enhance the resilience and generalizability of models. Another big challenge is reducing the ``distribution gap'' as the gap is made more problematic by the use of such deep models which can be very uncertain and can easily learn patterns outside the soul objective.

    \textbf{Relevance to data-centric AI.}\quad As noted in \cite{dcaiNot}, dealing with data or various kinds of distribution shifts is not new. What's new is that we are now at a stage where we need to develop a standardized procedure that can account for generalizing high-dimensional data and re-examining traditional mathematical \& computational techniques in light of AI deployment. Detecting the distribution shift has always been possible and achieved in a lot of different works as follows: \cite{RabanserGL19, abs-2107-06929, abs-2208-02896}. However, it is much harder to combat, avoid or fix the problem of distribution shifts with growing data. As with the growing data, we also see a variety of changes in behavior and characteristics of the real-world data. Industries have always been focused on increasing as much data as they can to achieve better performance and robustness. Growing data increase the challenges of multiple distribution shift in scale, that is, dealing with numerous kinds of shifts at the same time in a unified framework. The research presented in `Data-centric Artificial Intelligence: A Survey' \cite{zha2023datacentric}, along with related works by Polyzotis et al. \cite{polyzotis2021datacentric}, Jarrahi et al. \cite{jarrahi2022principles}, Jakubik et al. \cite{jakubik2023datacentric}, and Whang et al. \cite{whang2022data}, underscores the transition from a model-centric approach to a data-centric one in AI. This shift involves a systematic approach to data engineering, enhancing AI system construction, and thus moving the focus from models to data \cite{zha2023datacentric}. This paradigm transition is evidenced by collaborative efforts in both academic and industrial sectors, stressing the importance of data-centric methodologies in AI system development \cite{zha2023datacentric}. Techniques such as data augmentation \cite{feng-etal-2021-survey}, feature selection \cite{li2017feature}, and prompt engineering \cite{Liu2023prompt} have been developed, aiming to assure data's quality, volume, and trustworthiness, ensuring that the models function as intended \cite{zha2023datacentric}. In the past decade, much research has focused on individual types of distribution shifts, however, model deployments in the real world need to tackle all kinds of shifts in a unified skeleton on a given dataset in scale. The purpose of this study also includes different perspectives of the problems and exploring different solution spaces existing in the literature.

    \textbf{Motivation.}\quad Categorically, we are interested in the following three kinds of distribution shifts for the purpose of our study: (1) \textit{Subgroup Distribution Shift (Alignment Problem/Hidden Stratification)}, (2) \textit{Adversarial Distribution Shift}, (3) \textit{Unseen Distribution Shift (subset of Domain Shift)}. Starting with the paper's plot, we chose three different sets of distribution shifts although there are plenty of other distribution shifts out in the literature such as co-variate shift, concept shift, label shift (prior-probability shift), model drift caused by concept drift or data Drift and more sub-categories of shifts under them. We think, these distribution shifts are the building blocks of more complex data distribution shifts. Consider the simplest case of two attributes: the label and the nuisance attribute. If we consider the marginal distribution of the label, it decomposes into two terms: the conditional probability and the probability of a given attribute value: $p(y^{\text{label}}) = \sum_{y^{\text{attr}}}p(y^{\text{label}}|y^{\text{attr}})p(y^{\text{attr}})$. The three shifts namely: Subpopulation Shifts, Adversarial Perturbation, and Unseen Data Shift, where Unseen Data Shift and Adversarial Perturbation shifts controlling the $p(y^{\text{attr}})$. However, Subpopulation Shift phenomenon are caused due to Hidden Stratification that controls $p(y^{\text{label}}|y^{\text{attr}})$ as also evidently seen in \cite{wiles2022a}. There could be more combinations of different shifts affecting the fundamental blocks, however, the choice also accounts for the recent trends observed in the literature. 

\begin{table*}
    \centering
    \tiny
    \caption{\label{tab:notations}Summary of basic notations used in this paper.}
    \begin{tabular}{cl} \toprule
        \textbf{Notation}        & \textbf{Description} \\ \midrule
        n, $\mathcal{N} \in \mathbb{R}$ & Scalars are denoted by plain lowercase or uppercase letters. \\
        $x_{i}; i\in\mathcal{N}$        & Vector of data samples ranging from [0,$\mathcal{N}$] \\
        $y_{i}; i\in\mathcal{N}$        & Vector of labels ranging from [0,$\mathcal{N}$] \\
        $\mathcal{X}$                   & A tensor $\mathcal{X}$ with size $I^1\times\dots\times I^{\mathcal{N}}$ \\
        $\mathcal{Y}$                   & A tensor $\mathcal{Y}$ with size $I^1\times\dots\times I^{\mathcal{N}}$ \\
        $g_{i}$                         & A vector representing subgroup cluster inside each class $y_{i}$ \\
        $\mathcal{G}$                   & A tensor $\mathcal{G}$ of multiple subgroups drawn, $\mathcal{G}\in\mathbb{R}$\\
        \textit{l}                      & Loss on a data sample that is a measure of the difference b/w predictions \& actual values.\\
        $\mathcal{L}$                   & Average loss is a measure of the difference b/w predictions \& actual values.\\
        $\mathbb{E}$                    & Expectation/Expected value is an operator that is applied to a random variable.\\
        $\mathcal{D}$                   & Data distribution refers to the probability distribution of the input data.\\
        $\boldsymbol{h}_{\theta}(x)$    & hypothesis function, a function that maps input data $\mathcal{X}$ to output values $\mathcal{Y}$.\\
        w                               & Weights are the learnable parameters of a machine learning model that control signals b/w two neurons.\\
        $z_\textit{h}$                  & Logits layer of model h\\
        T                               & Scaler temperature hyperparameter, where, $T>0$\\
        b                               & Biases               \\
        $\|\boldsymbol{w}\|_1$          & The $\ell_1$ norm of a vector, calculated as the sum of the absolute vector values.\\
        $\|\boldsymbol{w}\|_2$          & The $\ell_2$ norm of a vector, calculated as the square root of the sum of the squared vector values.\\
        $\|\boldsymbol{w}\|_{\infty}$   & The $\ell_{\infty}$ norm of a vector, calculated as the maximum of the absolute vector values.\\
        $a_{i}\in\mathcal{A}$           & Distinct attributes $a_{i}$ within the set of attributes $\mathcal{A}$ for each class, representing unique features.\\
        $\theta$                        & Model parameters     \\
        $\Theta$                        & Parameter Space      \\
        $\eta$                          & Learning rate \\
        $\delta$                        & where delta is the purturbation vector \\
        $\theta_{\mathbb{B}}$           & Parameter space for the biased model, representing the parameter settings leading to biased predictions. \\
        $\theta_{\mathbb{D}}$           & Parameter space for the de-biased model, representing the parameter settings for bias mitigation. \\
        $\|\boldsymbol{W}\|_{0}$        & The $\ell_0$ norm, i.e., cardinality of a matrix, defined as the number of nonzero components.\\
        $\|\boldsymbol{W}\|_{1}$        & The $\ell_1$ norm of a matrix, calculated as the maximum of the $\ell_1$ norm of the column vectors.\\
        $\|\boldsymbol{W}\|_{2}$        & The $\ell_2$ norm of a matrix, calculated as its maximum singular value.\\
        $\|\boldsymbol{W}\|_{F}$        & The Frobenius norm of a matrix, calculated as the  square root of the sum of the squared matrix values.\\
        $\tau$                          & Classification threshold \\
        $\hat{x}$                       & input with adversarial noise \\
        $\xi_{i}$                       & i.i.d samples in $\mathbb{P}*$ with $i = [1,\dots,n]$\\
        $\mathbb{P}*$                   & Empirical measure $\hat{\mathbb{P}_{n}} = \frac{1}{n}\sum_{i=1}^{n} \delta_{\xi_{i}}$ leading to the empirical risk minimization (ERM) problem \\
        $\sigma_{y}^{-1}$               & Covariance matrix for in-distribution data for class $y_{i}$, representing the variability of the data around its mean. \\
        \bottomrule
    \end{tabular}
\end{table*}

\section{Background}

    In this section, we explore the fundamental aspects of the robustness problem in image classification, delving into commonly discussed concepts in the literature. We provide a detailed discussion on the assumption of independent and identically distributed (i.i.d.) data and offer a mathematical framework to understand robustness in this context.

    \subsection{Independent and identically distributed random variables (i.i.d)}
    \label{sec:iid}
    The premise of traditional learning algorithms is rooted in the assumption that both training and testing samples are drawn from distributions that are independent and identically distributed (i.i.d.), denoted as $\mathcal{D}_{train} = \mathcal{D}_{test}$. This assumption forms the basis for the empirical risk minimization (ERM) approach \cite{ERM}, which is centered on minimizing the average loss across training samples to derive an optimal model with strong generalization capabilities for the test distribution. The essence of ERM is encapsulated in the following mathematical expression:
    \begin{equation}
        \mathcal{L} = \frac{1}{n}\sum_{i=1}^{n}\textit{L}(h_{\theta}(\mathcal{X}), \mathcal{Y})
    \end{equation}
    Under this framework, the goal is to achieve a minimized loss function $\mathcal{L}$, which is calculated as the average of the loss function \textit{L} applied to each training example. The function \textit{L} evaluates the discrepancy between the predicted output of the hypothesis function $h_{\theta}(\mathcal{X})$ and the actual label $\mathcal{Y}$, over the entire training set consisting of 'n' samples. This approach is critical for developing models that are not only efficient on the training data but also exhibit a strong predictive performance when exposed to unseen data in the test set.

    \subsection{Robustness}
    \begin{definition}
        \cite{Li_2023, drenkow2022systematic, piratla2023robustness}~\textit{At a high level, robustness in the field of Machine Learning reflects ways one can achieve stable consistency between testing errors and training errors in the variability of the environment that tries to disconnect the test and train datasets, showing similar model performance between the two.}
    \end{definition}

    In addition, it is worth noting that robustness is orthogonal to performance. If a model performs very poorly on both the test and training scores, we can still consider it a robust model as its performance was consistent across datasets. When considering robustness, the performance itself is irrelevant, the only thing that matters is the consistency of performance between test and train sets.
    
    \subsubsection{Beyond Model Privacy: Is Robustness Useful?}\hfill

    Robustness in the context of machine learning extends beyond model resilience. It entails the model's capacity to sustain uniform performance when confronted with shifts in subpopulation distributions, minor variations in data, and situations involving data that the model has not previously encountered \cite{wiles2022a}. Such robustness is crucial; it ensures to some extent that if a model excels in test environments, it is more likely to sustain similar performance in real-world applications but not guaranteed. Moreover, robustness indicates a deeper, more meaningful understanding of the task at hand by the model, making it more reliable and trustworthy, especially in critical applications. This capability to effectively handle distribution shifts implies that the model goes beyond surface-level pattern recognition, achieving a profound comprehension of the underlying task. Philosophically, this could potentially offer insights into cognitive processes of AI systems. Failing to address this category of problems would leave us uncertain about our systems' genuine grasp of the tasks they are designed to perform, particularly in safety-critical applications where reliability is paramount.

    \subsubsection{Mathematical Perspective to Robustness for Classification Models}\hfill

    \label{robustness-through-utility}
    In this section, we will define a mathematical perspective among many other perspectives out in the literature for Robustness. Readers familiar with the relevant literature and a general notion can directly move on to the next section~\ref{sec:HIM}.
    Suppose we choose a prediction problem. We define the input features as $x \in \mathcal{X}$ and their corresponding labels as $y \in \mathcal{Y}$ and the prediction as $\hat{y}$, $\forall ~(\mathcal{X}, \mathcal{Y}) \in \mathbb{R}$, and $\theta \in \Theta$ representing model parameter in parameter space. And the training loss through back-propagation will be;
    \begin{equation} \label{eq:random-variable}
        \textit{l}\{(\hat{y},y);x,\theta\}
    \end{equation}
    Let the data-distribution be represented by $\mathcal{D}$ with subscripts `train' and `test' denoting training and real-world datasets. Given the above definition, our objective will be focusing on minimizing the expectation of the random variable \ref{eq:random-variable} given the input $x$ and the algorithm $\theta$;
    \begin{equation} \label{eq:expected-loss}
        \mathbb{E}_{\mathcal{D}_{train}}(\textit{l}\{(\hat{y},y);x,\theta\})
    \end{equation}
    The Equation~\ref{eq:expected-loss}. shows how we mathematically drive the algorithm to provide a utility to the user. Now to describe robustness in the above scenario, we introduce a utility function \cite{kurtz2018towards,ERM} ``assigning values to the actions that model can provide usefulness". 
    \begin{align}
        U(x, h_{train}) ~;~ \text{where}, h_{train} = \mathbb{E}_{\mathcal{D}_{train}}(\textit{l}\{(\hat{y},y);x,\theta\})
    \end{align}
    Here, the $h_{train}$ is an algorithm that is performing on the given training data $x$ and provides a prediction $\hat{y}$. Whenever predictions are accurate they are considered useful. So a good utility function will acknowledge accurate predictions as usefulness. Hence, $U > 0$ (positive utility is considered better by convention). Shifts in $\mathcal{D}$ can happen due to data corruption, naturally or adversarially. Corruption can occur in all parts of the machine learning pipeline and can take various forms. Additionally, multiple types of corruption can happen at the same time, as described in the reference \cite{UW_Robustness}. We explain in detail about natural corruption causing shifts in sections~[\ref{sec:SG},\ref{sec:US}] and adversarial corruption causing shift in section~\ref{sec:AP}.
    \begin{align} \label{eq:corruption}
        \mathcal{D} & = \mathcal{D}_{train} \pm ~\varepsilon \\
        U(\boldsymbol{h}, x_{\varepsilon}) & \geq 0 \\
        U(\boldsymbol{h}, x_{\varepsilon}) & < U(\boldsymbol{h}, x)
    \end{align}
    In Equation~\ref{eq:corruption}, disturbance/corruption is a general term encompassing any phenomenon that can induce a distribution shift. To estimate the robustness across mean, we recover the true mean ($\mu$) of the distribution ($\mathcal{D}$).
    \begin{problem}
        \cite{UW_Robustness} Let $\mu \in \mathbb{R}$, and let $\boldsymbol\epsilon < \frac{1}{2}$.~ Let $\mathcal{D}$ be a set of $n$ samples and $\mathcal{D} = \mathcal{D}_{train} \cup \mathcal{D}_{test}$, where $\mathcal{D}_{train}$ is a set of i.i.d. samples from Gaussian distribution $\mathcal{N}(\mu,\sigma^2)$, and $\mathcal{D}_{test}$ satisfies $\mid\mathcal{D}\mid < \varepsilon\cdot n$. Given $\mathcal{D},\varepsilon,\sigma$ and output mean distribution $\hat{\mu}$ minimizing $\mid\mu-\hat{\mu}\mid$.
    \end{problem}
    Here, $\mu$ can also be represented by $\mathbb{E}_{{(x,y)} \sim \mathcal{D}} [\Bar{h}(x,y)]$ based on Eq~\ref{eq:expected-loss}, hence we can certify the upper bound of the gap as in Kumar et al., \cite{certi-ds}; 
    \begin{equation}
        \mid \mathbb{E}_{{(x_{train},y_{train})}\sim\mathcal{D}} [\Bar{h}(x_{train},y_{train})] - \mathbb{E}_{{(x_{test},y_{test})}\sim\mathcal{D}} [\Bar{h}(x_{test},y_{test})] \mid ~\leq~ \psi(\varepsilon)
    \end{equation}
    where, $\psi$ is a concave function that bounds the total variance (distribution gap)

    \subsubsection{Problem Definitions}\hfill

    For all the three shifts, let us consider a similar classification task setting as described in Section \ref{robustness-through-utility}, in Madras et al.'s description in \cite{Post_hoc}. Firstly, we introduce the problem definition which includes subgroups in the problem definition. 
    \begin{definition}\label{def:subgroup-shift}
        Given a dataset: $\{(x_{i}, y_{i})\}_{i=1}^{N_{d}}$, where $x \in \boldsymbol{X}, ~y \in \boldsymbol{Y} ~ \forall ~ \boldsymbol{Y} = \{1,...,\boldsymbol{Y}\}$. Also, considering the categorical subgroups inside the groups (for simplicity, let's call them supergroups) of $\{1,\dots, \boldsymbol{Y}\}$ be $\{g_{i}\}_{i=1}^{n}~;~ g \in \{1 \dots \boldsymbol{G}\} ~:~ c_{i} = g$ where, `i' is frequency of data instances inside the subgroup `g'.
    \end{definition}
    The most important assumption in any general subgroup supervised training problem setting is that the subgroup information is only available during training and validation, not during testing. $C$ represents a non-uniform distribution across the training set to mimic uncertainty in the real-world. For each distribution over the set $(X, Y, d) ~:~ \{(x_{i}, y_{i}, d)\}_{i=1}^{N_{d}}$ where, $N_{d}$ represents the number of samples in domain $d$. As $\mathcal{D}$ represents a mixture of domain samples, Yao et al., in \cite{LISA} defines the training distribution as:
    \begin{equation}
        \begin{aligned}
            \rho_{train} = \sum_{d\in\mathcal{D}}\lambda_{d}^{train}\rho_{d}
        \end{aligned}
    \end{equation}
    Hence, the training and testing domains can be formulated as shown below:
    \begin{equation}
        \left.
        \begin{array}{ll}
            \mathcal{D}_{train} = \{d\in \mathcal{D}_{train}~|~\lambda_{d}^{train} > 0\}\\
            \mathcal{D}_{test} = \{d\in \mathcal{D}_{test}~|~\lambda_{d}^{test} > 0\}
        \end{array}
        \right \} \Longrightarrow
    \end{equation}
    The robustness of the model can be calculated by measuring the \textit{worst-group accuracy} for the $0-1$ loss function (binary setting) as minimizing the chance of classifying the subgroup `g' as any other group.

    \begin{gather}
        \text{min}_{g}~\mathbb{E}[ \mathds{1}[\bar{h}(x) = y] | g_{i} = g]\\
        \text{where, ~} \boldsymbol{\bar{h}}(x) \rightarrow ~\textit{argmax}~\boldsymbol{h}(x)~\text{and}~\boldsymbol{h}(x) = \sigma(w^\top{T}x + b) \notag
    \end{gather}

    Secondly, we introduce noise that is adversarially added on a feature vector input. If this notion is relevant, the reader can skip and move on to the next paragraph. The classification neural network model learns a feature hypothesis $h$, given a natural input $x$ with it's label value $y$, an adversarial perturbation on the input image, adversarial image $\hat{x}$, such that $\|x-\hat{x}\|$ the magnitude of their difference is so small that the perturbation on the image is invisible to the naked eye. Typically, when the adversarial image is classified as a label $\{i \neq y; ~h(\hat{x}=i)\}$ it is termed as \textit{untargeted attack}, and when the adversarial image is classified as a label $\{j \neq y; ~h(\hat{x}=j)\}$ where $j$ is the target of the attacker is termed as \textit{targeted attack}. Attacks generated by any of the methods described in this section use a perturbation scheme ($\delta$) that is added to the benign input array ($X$) (Equation \ref{eq:basic_attack_scheme}). In particular, we can use different proxies (distances, instead of human in the loop) to capture the difference between original and adversarial twin example, metrics like $l_0$, $l_1$, $l_2$, $l_{\infty}$ norm. Typically for an attacker, recognizing the precise formulation of these metrics isn't crucial but rather having an understanding that higher the number for a norm, the more weighted that distance is towards being influenced by outliers.
    \begin{align}
        \hat{x} = x + \delta \text{ \quad(where, }\hat{x} \text{ also referred as }x_{adv}\text{ in literature)}
        \label{eq:basic_attack_scheme}
    \end{align}
    The general norm function is described  \cite{magnet} as:
        \begin{align}
            ||w||_{p} = \left( \sum_{i=1}^{n}|w_i|^p\right)^\frac{1}{p}
            \label{eq:norm}
        \end{align}
    where $i$ is the dimension of the $w$ vector and $p$ is the norm. Most researchers do not consider $l_{0}$ a norm.\footnote{mathematically it returns the non-zero elements from the image vector}. The $l_2$ norm measures the short distance matrix, popularly known as the Euclidean distance matrix. \footnote{It calculates the square root of the sum of squared distances between any two points, i.e., it returns the hypotenuse distance between the two points.} Lastly, $l_\infty$ returns the absolute maximum value out of all the sets of values. Now that we have an idea of a single attack, let's dive into adversarial distribution.

    Considering the underlying distribution $\mathcal{D}$ over training and testing datasets consisting of data samples $x\in\mathcal{X}$, the adversarial distribution can be written as the set of adversarial examples, $\mathcal{D}_{adv} := \{x_{adv}\}_{i=1}^\mathcal{N} = \{x_i+\delta_i\}_{i=1}^\mathcal{N}$. 
    We treat the a small region including the set of the perturbation vector $\delta$ as the adversarial distribution shift against natural features. So, the perturbation vector in Equation~\ref{eq:basic_attack_scheme} can be rewritten as $\delta \in \mathcal{D}_\delta := \mathcal{D}_{adv} - \mathcal{D}$.


\begin{table}[htbp]
    \caption{\label{tab:selected-articles}Highlighting the selected crucial research articles in each category of distribution shift.}
    \scalebox{0.76}{
    \begin{tabular}{p{2.2cm}cccccccc}
        \toprule
        \multirow{4}{*}{\textbf{Surveys}} &
        \multirow{4}{*}{\textbf{Venue}} &
        \multirow{4}{*}{\textbf{Year}} &
        \multicolumn{6}{c}{\textbf{Robustness Criteria}} \\ \cline{4-9} 
        &&&
        \multicolumn{2}{c}{\textbf{\begin{tabular}[c]{@{}c@{}}Subgroup\\ Distribution\end{tabular}}} &
        \multicolumn{2}{c}{\textbf{\begin{tabular}[c]{@{}c@{}}Adversarial\\ Distribution\end{tabular}}} &
        \multicolumn{2}{c}{\textbf{\begin{tabular}[c]{@{}c@{}}Unseen\\ Distribution\end{tabular}}} \\ \cline{4-9} 
        &&&
        \begin{tabular}[c]{@{}c@{}}Spurious\\ Correlations\end{tabular} &
        \begin{tabular}[c]{@{}c@{}}Bias \&\\ Incomplete Data\end{tabular} &
        \begin{tabular}[c]{@{}c@{}}Adversarial\\ Training\end{tabular} &
        \begin{tabular}[c]{@{}c@{}}Certified\\ Robustness\end{tabular} &
        \begin{tabular}[c]{@{}c@{}}Data\\ Augmentation\end{tabular} &
        OODs \\
        \midrule

        \citeauthor{imbalanced-domains-survey} &
        ACM CS &
        2016 &
        \xmark &
        \cmark &
        \xmark &
        \xmark &
        \cmark & 
        \xmark \\

        \citeauthor{trustworthy-ai-survey} &
        ACM CS &
        2022 &
        \xmark &
        \xmark &
        \cmark &
        \xmark &
        \cmark & 
        \cmark 
        \\

        \citeauthor{subgroup-and-unseen-shift-survey-3} &
        ACM CS &
        2022 &
        \xmark &
        \xmark &
        \cmark &
        \cmark &
        \cmark &
        \cmark \\

        \citeauthor{adv-robustness-survey-2} &
        IJCAI &
        2021 &
        \xmark &
        \xmark &
        \cmark &
        \cmark &
        \xmark &
        \xmark \\

        \citeauthor{adv-attacks-survey-1} &
        ACM CS &
        2021 &
        \xmark &
        \xmark &
        \cmark &
        \cmark &
        \xmark &
        \xmark \\

        \citeauthor{adv-attacks-survey-2} &
        Comp. Sec. &
        2022 &
        \xmark &
        \xmark &
        \cmark &
        \xmark &
        \xmark &
        \xmark \\

        \citeauthor{evaluating-adversarial-attacks-driving-safety} &
        IEEE IT &
        2022 &
        \xmark &
        \xmark &
        \cmark &
        \xmark &
        \xmark &
        \xmark \\

        \citeauthor{unseen-shift-survey-1} &
        J. Big Data &
        2019 &
        \xmark &
        \xmark &
        \xmark &
        \xmark &
        \cmark &
        \xmark \\

        \citeauthor{MUMUNI2022100258} &
        Array &
        2022 &
        \xmark &
        \xmark &
        \xmark &
        \xmark &
        \cmark &
        \xmark \\

        \citeauthor{unseen-shift-survey-2} &
        Pattr. Recog. &
        2023 &
        \xmark &
        \cmark &
        \xmark &
        \xmark &
        \cmark &
        \xmark \\

        \citeauthor{unseen-da} &
        J. Imaging &
        2023 &
        \xmark &
        \xmark &
        \xmark &
        \xmark &
        \cmark &
        \xmark \\

        \textbf{\textit{Ours}} &
        - &
        2023 &
        \cmark &
        \cmark &
        \cmark &
        \cmark &
        \cmark & \cmark \\ \midrule

    \end{tabular}}

    \raggedright\tiny{Fullforms for abbreviations used: ``ACM CS'' - ACM Computing Survey; ``J. Imaging'' - Journal of Imaging; ``IJCAI'' - International Joint Conference on Artificial Intelligence; ``Comp. Sec.'' - Computer \& Security; ``Pattr. Recog.'' - Pattern Recognition; ``J. Big Data'' - Journal of Big Data}
\end{table}


\section{Related Work on Existing Surveys}

    There are several approaches that have been proposed to address these categories of distribution shifts. We gathered those surveys in this section. BRANCO et al., \cite{imbalanced-domains-survey} is a survey of techniques for handling imbalanced distributions in predictive modeling for classification and regression tasks, with a focus on real-world applications such as fraud detection and stock market prediction. The article provides taxonomy of methods before 2016. Li et al., \cite{trustworthy-ai-survey} provides a theoretical framework covering robustness, generalization, explainability, transparency, reproducibility, fairness, privacy preservation, and accountability. The sections on robustness and generalization talk discusses shift due to adversarial attacks and unseen data in a high-level respectively. Mohseni et al.,'s survey \cite{subgroup-and-unseen-shift-survey-3} discusses ways to make ML algorithms more reliable in open-world applications by addressing vulnerabilities like interpretability and performance limitations. It proposes a taxonomy of ML safety to guide the development of more dependable ML systems. Ramponi and Plank's work in NLP \cite{ramponi-plank-2020-neural} talks about the domain shift problem and methods on domain adaption in NLP domain, and highlights the need for out-of-distribution generalization and learning beyond 1:1 scenarios in future research. The survey \cite{adv-robustness-survey-2} presents recent advances in adversarial training for robustness are presented and analyzed in this paper, including the introduction of a new classification system.  This survey \cite{adv-attacks-survey-2} reviews adversarial attacks in computer vision before 2022, presents a taxonomy, and suggests future research directions. It also analyzes the field's development using citation relationships and a knowledge graph. Machado et al.,'s  \cite{adv-attacks-survey-1} article reviews the latest research on adversarial machine learning in image classification until 2020.

    \textbf{Contribution.}~ Overall, research on the aforementioned three types distribution shifts has been limited, and surveys are targeted in either of the types of shifts, or not deeply explored. In addition, the definition of robustness varies considerably depending on the sub-field to which the applications belong. Nonetheless, it is important to note that a good robust model will incorporate all of the previously mentioned categories. As the deployed model in the real-world can be subjected to either a spurious correlation, a perturbation attack on the deployed model during testing, or an unseen corruption shift. There have been no survey papers yet that summarize work that incorporates all of these distribution shift categories into a single framework. Hence, we believe this report is the first to summarize and going to serve the purpose.

\section{Subgroup Distribution Shift}
\label{sec:SG}

    \begin{wrapfigure}{r}{0.41\textwidth} 
        \centering
        \includegraphics[width=0.38\textwidth]{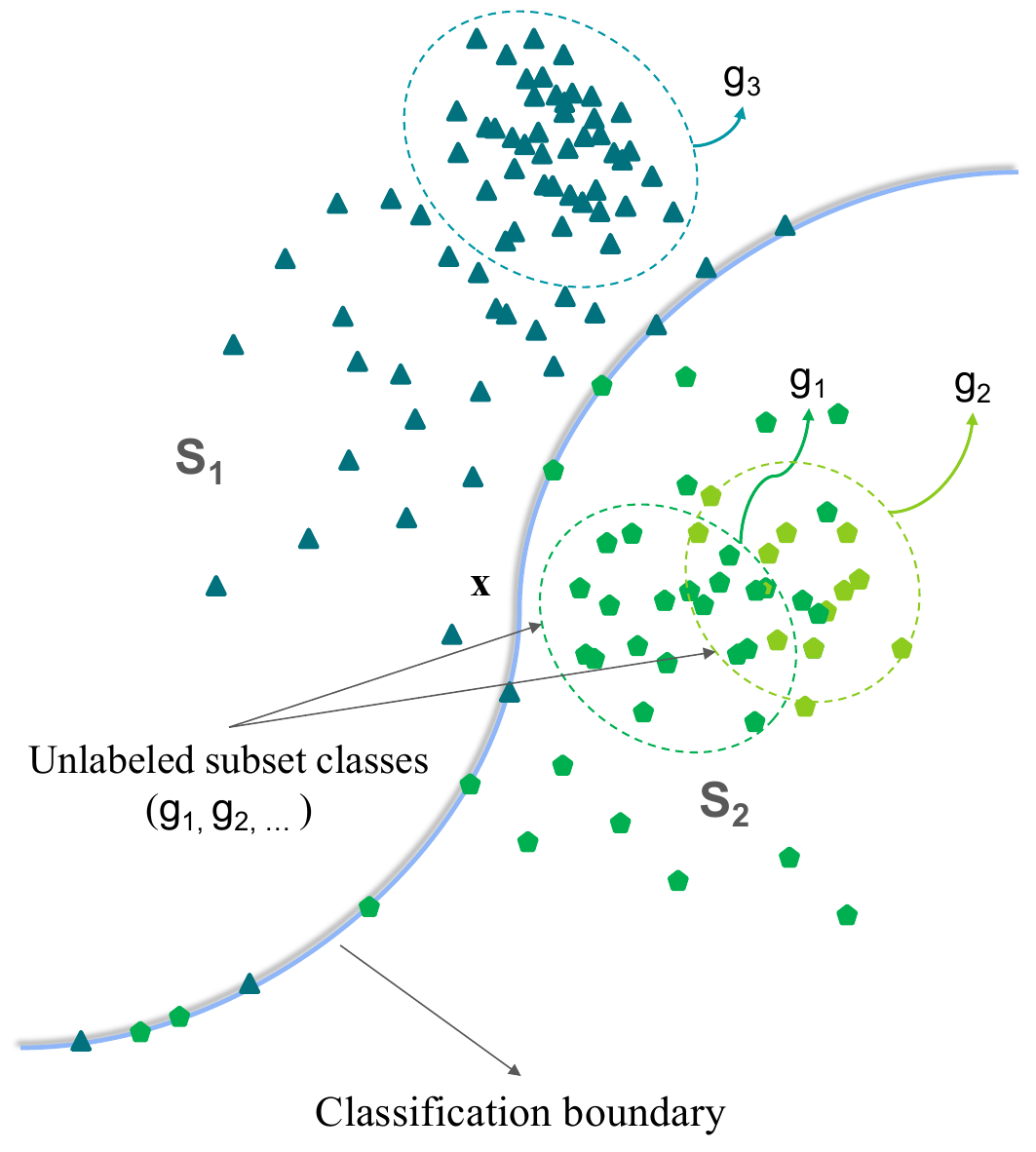} 
        \caption{\label{fig:HS}\textit{Visualizing hidden stratification. $\{S_1,S_2\}\in S$ is a super-class; $\{ c_1, c_2 \} \in C$ is a sub-class.}}
    \end{wrapfigure}
    A machine learning task with subpopulation shift requires a model that performs well on the data distribution of each subpopulation. A common task in machine learning is a classification of different categories of subpopulation in data. For example, differentiating between dogs and cats. In reality, each class of data we come up with, has numerous subclasses, such as the individual breeds of each dog, which can be further divided into more subclasses to reflect intra-class variation. The ImageNet \cite{imagenet} dataset utilizes a hierarchical structure that reflects this well and is visually demonstrated by Santurkar et al. \cite{BREEDS}. 
    However, it is impossible to cover 100\% of all subclasses in a given dataset. When subclasses are not labeled, it can lead to ``hidden stratification'', which can cause numerous issues. In this section of subgroup distribution shift, the paper focuses on the following issues (as in Tab.~\ref{tab:selected-articles}) that lead to hidden stratification: Spurious correlations \cite{Dunnmon2019AssessmentOC}, \cite{HSClinical}, \cite{Seahe053024}, \cite{human_annotation}, \cite{overparameterization}, \cite{spurious_correlation_ood}, \cite{tacl_a_00335}, Subpopulation shift \cite{ERM}, \cite{IRM}, \cite{Rex}, \cite{DRO-example-1}, \cite{GDRO}, \cite{CORAL}, and Subgroup bias/label noise \cite{LFF}, \cite{label_noise}.

    \subsection{Hidden Stratification.}\hfill

    Let's understand a similar scenario where the alignment problem exists fundamentally in the dataset itself. In the case of classification problems, the phenomenon where the model performs well on the overall task but at the same time also under-performs on a subset of the dataset can be termed as hidden stratification. Often the term hidden-stratification is used as a subgroup relation task in machine learning. The phrase was first defined in the article by Oakden-Rayner et al. titled "Hidden Stratification Causes Clinically Meaningful Failures in Machine Learning for Medical Imaging" (HSClinical). Often we only know the superclass labels for the task, but not the subclass labels and models are usually trained to optimize for average performance and so can underperform on important subclasses. Deep learning models that have been trained on medical images for a specific task or a collection of tasks that are quite similar often provide an average model performance in the clinical domain. However, such models hide performances of the model on clinically distinct and meaningful subgroups within these single findings \cite{Seahe053024}. It is typically straightforward to find the problem of hidden stratification on any dataset that can substantially lower the model's performance. However, measuring its effects and finding a general technique to resolve the problem in subgroups is still highly challenging. Recently, researchers have demonstrated a few ways to tackle, generalize, and repair the impact of hidden stratification. One important development was \cite{Seahe053024}, which established a domain expert reviewed dataset for hidden stratification in medical imagery. 
    More broadly, \cite{HSClinical} divided hidden stratification problems into the following three sub-categories. 

    \subsection{Detecting Hidden-Stratification}
    Distinct approaches to measuring hidden stratification effects. 
    \paragraph{Schema completion~} It is a technique where the author of a dataset exhaustively labels every possible subclass, leading to a more complete set of subclasses. This helps with accurate reporting and model development in industry-related challenges. This is a constrained strategy, as it calls for a group of data annotators for quality annotation.
    \paragraph{Error Auditing~} This method analyses model predictions by retrospectively analyzing the accuracy of predictions. This is done by comparing the distribution of a subclass in groups of accurate and inaccurate predictions. In application, Oakden-Rayner et al. \cite{HSClinical} and Seah et al. \cite{Seahe053024} observed the spurious attributes to be chest drains in the Chest-X-ray dataset is deceiving and that it account for 80\% of test dataset.
    \paragraph{Algorithmic Measurement~} The Algorithmic Measurement approach focuses on using automated methods to detect hidden stratification in datasets. This method differs from the other two manual methods and uses algorithms such as k-means to form unsupervised clusters and observe hidden stratification. It may only work well on simple datasets, perhaps for complex datasets there are methods (described in Sec.~\ref{sec:solutions-HS}) to replace k-means and are more suitable for understanding the data respresentation.

    Methods defined in the later part of Hidden-Stratification sub-sections are relatively more efficient, and recent in the literature.

    \begin{wrapfigure}{r}{0.28\textwidth}
        \centering
        \includegraphics[width=0.26\textwidth]{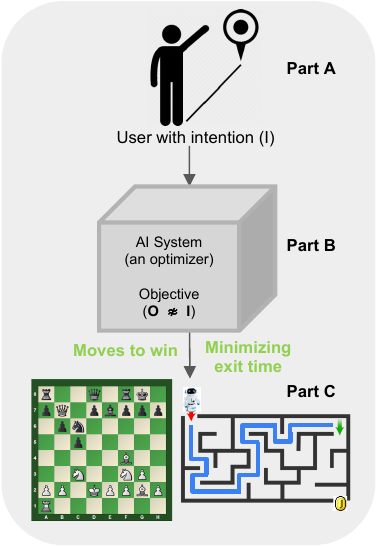}
        \caption{\label{fig:alignment-problem}\textit{Misaligned objectives in AI system optimization.}}
    \end{wrapfigure}
    \subsection{Spurious Correlation}

    \subsubsection{Shift In Data Causes Human Intention Misalignment.} \label{sec:HIM}

    Robustness is a core component of model generalization and ML safety. In a high-level, this section describes relation of spurious-correlation and overall system. As a human, we have an objective. We portray our objective by creating an AI system. Now the AI system has an optimizer which means it has a certain goal to optimize (by min. or max. the objective). As in \textbf{Figure~\ref{fig:alignment-problem}}, we choose the AI for solving the maze. The algorithm can be about finding minimum time or distance to exit the maze. Instead, in the Fig.~\ref{fig:alignment-problem} the model learns to find the green arrow. Part of the reason we have a problem is that ``Part A'' and ``Part B'' will almost certainly end up not being the same \cite{misalignment-example-1,misalignment-example-2}, primarily when the objectives refer to the real world with all its complexity, ambiguity, and uncertainty, so we have this ``\textbf{Alignment Problem}.'' 
    This is why, getting the machine's objective to align with ours is extremely difficult. During deployment, the distribution of inputs "shifts" to something the system is not equipped to handle. It could be that the problem stems from our reliance on data-centric AI, at which point the solution would be to move towards other forms of AI. However, closely observing the data and it's characteristics will help understand how and why such correlations are formed.

    Subgroups can often have correlations in only specific scenarios and usually models are biased in learning such correlations about those classes and explore similar correlations during the testing. Such correlations that do not transfer outside of a subset of data are known as ``Spurious Correlations.'' A model may learn such correlations if they are prevalent in its training data, which will lead the model to perform poorly on testing sets that do not reflect the spurious correlations. 

    In the paper by Bender and Gebru et al., \cite{10.1145/3442188.3445922}, researchers investigate how neural networks can learn spurious correlations in natural language processing tasks. They demonstrate that large pre-trained language models can be over-parameterized and learn spurious correlations in the training data that are not relevant to the task at hand. Another e.g. is Waterbird dataset, a classification dataset for classifying between two birds; $\mathcal{A} \in \{waterbird,landbirds\}$. Here, the challenge is the spurious association of bird images with it's background/environment $e \in \{water, land\}$.

    In the context of image classification, several studies \cite{TerraIncognita, GeirhosRMBWB19, GDRO, varryingCoeffModeling} have shown that neural networks can inappropriately rely on biased data, such as background or texture features, rather than properties of the subject, to obtain high accuracy. The Waterbird dataset \cite{spurious_correlation_ood} is an example of such a dataset, where the challenge is the spurious association of bird images with their background/environment. The neural network may learn to associate waterbirds with water and landbirds with land, rather than learning to classify birds based on their physical features. The problem of spurious correlations has been studied for many years and was first defined by Herbert A. Simon in 1954 \cite{spurious_correlation_1954}. Simon discusses how causal inferences can be made based on observed correlations and highlights the importance of understanding the underlying mechanisms that give rise to the correlations.
    
    In summary, subpopulation shift in machine learning can lead to spurious correlations that models may learn and rely on during testing. This problem has been observed in various domains, including image classification and natural language processing.
    %

    \begin{figure}[!ht]
        \centering
        \includegraphics[scale=0.35]{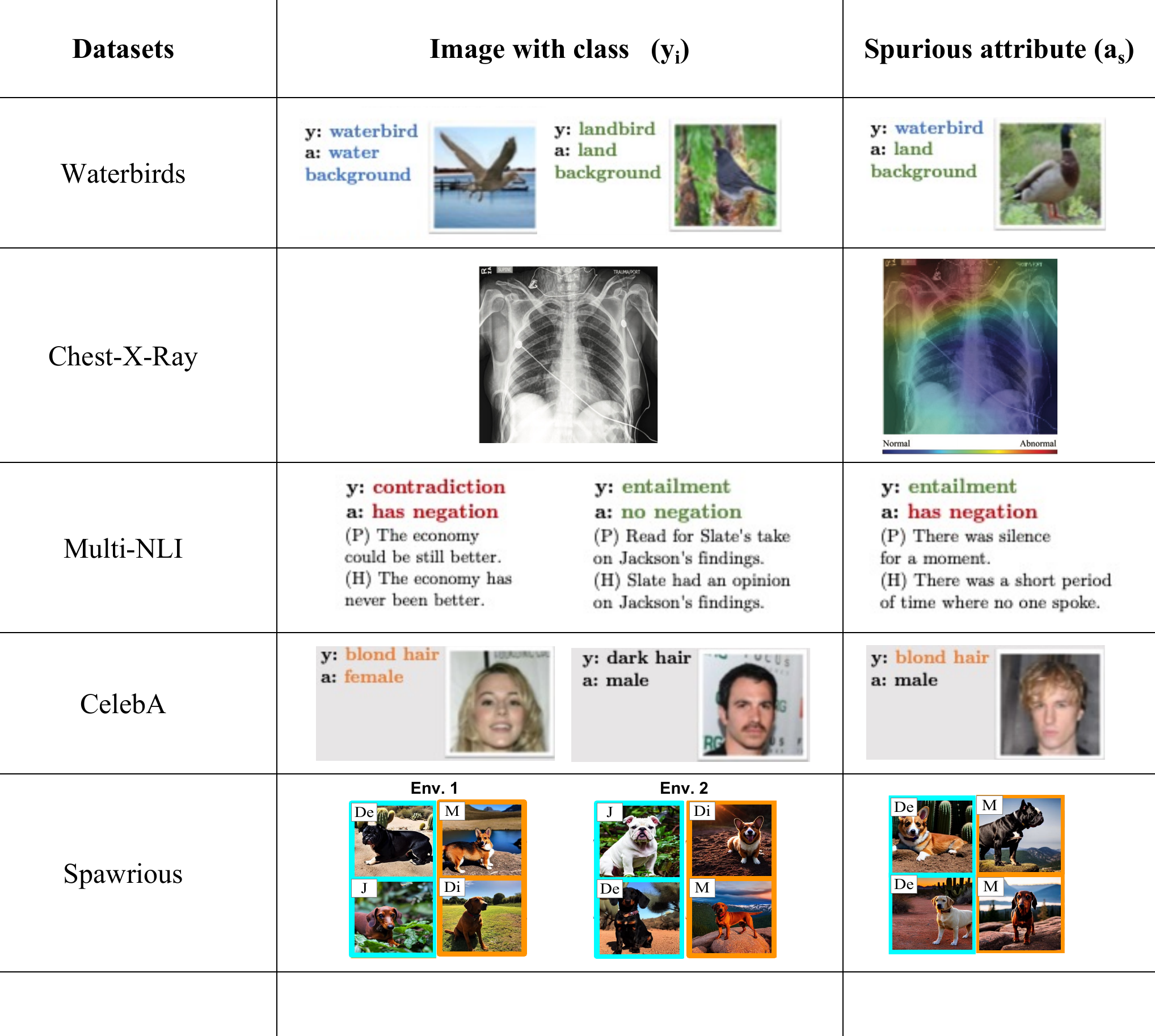}
        \caption{\label{fig:spurious_corr}\textit{
        The table shows the examples of spurious correlation existence and setting. Radiograph Chest X-Ray image shows that the chest drain spuriously aligns with the target: lung disease causing model to learn incorrect patterns with higher weights. This figure is from the CXR-14 dataset, where authors termed the phenomenon as hidden stratification and following authors addressed it too. \cite{HSClinical,Seahe053024,human_annotation,GEORGE,lynch2023spawrious}. Similarly other datasets show how the spurious attributes are causing model to learn wrong intentions behind classifying the target. 
        }}
    \end{figure}
    
    Classifying radiographs chest x-rays as a normal or abnormal is a perfect example of deep learning task. It's been widely known that due to lack of expert observations, tons of radiographs go unread. Dunnmon et al. claims the baseline is very low as no experts look at the image \cite{experts_shortage, unread_radiographs, Mitra2020ASS}. A clear motivation for applying machine learning for the task given the lower baseline. 

    Below we describe a set of clinical research papers where the problem of hidden stratification has been widely noticed and suffered significantly.

    Dunnmon et al. \cite{Dunnmon2019AssessmentOC} initiated a large-scale labeled data collection for chest radiography that has been limited in application. As it's well known in the research community, having ample labeled data always helps the model perform well, improve, and generalize on the dataset. The significant problems at the time were limited labeled data for chest radiographs, and also US National Institute of Health's database (state of the art dataset at the time) included noisy multi-class labels which demonstrated challenges for achieving reliable multi-class thoracic diagnosis prediction with chest radiographs. Authors' hypothesis is about reducing the automated analysis problem to a binary triage classification task would benefit by increasing performance levels in combination to clinician judgment on a task that was relevant to clinical practice using a prospectively labeled data set of a size accessible to many institutions. Dunnmon et al. collected a total of 313k chest radiographic images between the time period of \textit{January 1, 1998 - December 31, 2012} and based on the given classification problem, they trained the popular state of the art convolutional neural network (CNN) models: AlexNet \cite{alexnet}, DenseNet \cite{densenet}, and ResNet \cite{resnet}. 

    \subsection{Subpopulation Shift}
    Subpopulation shift occurs when sub-classes (hidden or otherwise) have different distributions between the training and testing sets for the model. One example could happen in a dog vs. cat classification task. If `Labradors' and `beagles' are most prevalent during the training, while `Chihuahuas' and `pit-bulls' are more common during testing, the model will likely perform poorly, likely to classify Chihuahuas as cats due to strong characteristic similarities. Good datasets involve a balanced distribution of sub-classes in both train and testing sets.

    In order to assess a model's resistance to subpopulation shift, Santurkar et al. \cite{BREEDS}, proposed a method which simulates subpopulation shift in ImageNet and other datasets. Their strategy involves dividing subclasses into two disjoint, random groups and assigning one to the target domain and the other to the source domain. Trains on source domain and tests on target domain as benchmark subpopulation shift large scale dataset wherein the data subpopulations present during model training and evaluation differs. The underlined ideology by creating subpopulation shift in this manner is so the model will struggle under these controlled conditions in a measurable way. \cite{BREEDS} idea can be easily generalized on any dataset with a semantic grouping of classes. ImageNet \cite{imagenet} is easiest to this due to hierarchical nature of the dataset structure. From a traditional statistical setting, the statistical error bounds assumes that the number of samples is much larger than the dimensionality of the problem, but in the recent years, DNNs with more parameters \& algorithmic regularization \cite{belkin2021fit, allen2019convergence, du2019gradient, jacot2018neural} than samples have outperformed traditional models. This discrepancy has received attention from the learning theory community, with studies showing that algorithmic regularization can have negative effects in the context of subpopulation shift. The most popular approaches, such as DRO and sample re-weighting (e.g. SMOTE) can overfit the training data, and empirical studies have shown that DRO overfits in subpopulation shift tasks \cite{GDRO, overparameterization}. In later Sections under subgroup distribution shift, we show how this implicit bias is either removed, converged or mitigated under various settings.\\

    \subsubsection{Dataset-Independent Benchmark to Measure Robustness.}\hfill
    
    Santurkar et al. proposed the benchmark dataset approach that measures a family of distribution shifts. The author also points out that in most cases, literature also talks about three distribution shifts that the literature has distinctly addressed to some extent. They are as follows: (1) Data corruptions: applying family of transformation to portray real-world noise from the source distribution \cite{fawzi2015manitest, fawzi2016robustness, ford2019adversarial, kang2019testing, shankar2021image}, (2) Difference in data sources: more than one independently collected data about the same subject of interest \cite{SaenkoKFD10, TorralbaE11} w.r.t geographic location \cite{TerraIncognita}, time \cite{KumarML20}, or user population \cite{abs-1812-01097} (3) Subpopulation representation: mostly literature talks about if the model performs equally on any subpopulation \cite{HuNSS18, abs-1810-08750, abs-1812-01097, OrenSHL19, GDRO}. However, above mentioned approaches to the problems cover only single facet of the problem, and it is not necessarily true that the robustness would translate one facet to other kinds. Hence, authors approach is to target multiple types of distribution shifts through a single benchmark dataset that encompasses multi-facet distribution shifts. Using any dataset of interest, such as ImageNet for example, the method first defines the superclasses based on semantic hierarchy. Semantic grouping is used mainly because the existing class hierarchy that is mostly found in visual classification tasks are grouped opposed to visual characteristics (For example: ``Umbrella'' \& ``roof'' are both ``coverings'') or classy hierarchy has different level of granularity between different classes. Also, in some cases, hierarchy might not represent a tree (for example: ``bread'' can be found in both ``baking goods'' \& ``starches''). However, BREEDs benchmark carefully validates the benchmark dataset via human annotations i.e., they make sure therein superclasses contain visually coherent similarities among them so that a model can easily generalize on cross-subpopulation. Authors randomly select two superclasses and then present some images to the annotators to be labelled one among the two superclasses, basically leveraging human annotators to measure the robustness of humans to subpopulations.


    \subsection{Subgroup Bias}
    A third major category of issues that can arise from hidden stratification is subgroup bias. If a specific subgroup has biases attached to it, it may lead to a noisy, imbalanced label distribution. In facial recognition datasets, Asians may have this issue as they appear younger than other races, therefore their estimated ages tend to be lower \cite{buolamwini2018gender, raji2019actionable}. The authors found that the algorithm was less accurate for black defendants than for white defendants in the criminal justice system, and that it was also less accurate for younger defendants of all races (that included Asians in majority) \cite{kleinberg2016inherent}.
    One may consider subgroup bias and spurious correlations to be similar classes of problems. Indeed, they are closely related. Subgroup bias can stem from spurious correlations between a subgroup and some criteria, for example. These two categories approach the same issue from different perspectives. Label noise \cite{{label_noise}}, for example, could help us to understand how a portion of noisy labels in the data can be particularly attention-grabbing for a model, which can in turn mislead a model to assume spurious correlations.
    \par In \cite{label_noise} by Deuschel et al., the authors devise a benchmark to test such natural distribution shifts. The benchmark tests for resistance to spurious correlations by curating younger (or older) faces from each race (in face dataset), and training only on those instances. By doing this, spurious correlations are artificially introduced in a controlled manner. The benchmark evaluates model robustness by testing its performance under various controlled conditions including noise in age labels and subpopulation shifts by omitting races from the training data. 

    Research studies have demonstrated the impact of hidden stratification on subgroup bias. For instance, in a study by Obermeyer et al. \cite{obermeyer2019dissecting}, the authors showed that a widely-used algorithm for prioritizing patients for extra care was biased against black patients, and this bias could be traced back to the prevalence of biased training data that ignored differences in healthcare utilization between different racial groups. Similarly, Buolamwini and Gebru \cite{buolamwini2018gender} found that facial recognition algorithms from leading technology companies showed higher error rates for darker-skinned individuals and women, which was attributed to the lack of representation of these groups in the training data.

    One way to address the problem of hidden stratification is to identify and mitigate the effects of confounding variables that are correlated with both the features and the target variable \cite{obermeyer2019dissecting}. Another approach is to increase the diversity and representativeness of the training data by collecting more data from underrepresented subgroups and balancing the distribution of features and labels across the subgroups \cite{buolamwini2018gender}. In addition, it is important to evaluate models on multiple subgroups and measure their performance across different dimensions of fairness and bias \cite{friedler2019comparative}.
    
    In conclusion, hidden stratification in the dataset can lead to subgroup bias in machine learning models. To address this issue, it is important to identify and mitigate confounding variables, increase diversity and representativeness of the training data, and evaluate models across multiple subgroups and dimensions of fairness.
    
    Nam et al., in \cite{LFF} provides a possible solution to the subgroup bias problem by observing that models learn biases only when they are ``easier'' to learn than the target criteria. The authors exploit this fact by training two models in parallel. One is trained on a subset of the data known to product bias, and the other is trained on a more challenging subset of the data. Using ``Relative Difficulty'', the system determines what causes biases in the biased model, and re-weights the de-biased model accordingly. Equation~\ref{eq:LFF} defines the biased and de-biased model updation respectively, where GCE is a generalized cross entropy and CE is the cross-entropy (GCE making a assumption when considering a noisy labeled dataset, that the noise is conditionally independent given the true dataset. \cite{GCE}). GCE helps in amplifying the biases through the biased model which is used for re-weighting training samples using the \textit{relative difficulty score} (W) in Equation~\ref{eq:W}.

    \begin{equation}
        \left.
        \begin{array}{ll}
        f_{\mathbb{B}}(x;\theta_{\mathbb{B}}) ~\text{by } ~\theta_{\mathbb{B}} \leftarrow \theta_{\mathbb{B}} - \eta \delta_{\theta_{\mathbb{B}}} \sum_{(x,y) \in \mathbb{B}} GCE(f_{\mathbb{B}}(x),y)\\\\
        f_{\mathbb{D}}(x;\theta_{\mathbb{D}}) ~\text{by } ~\theta_{\mathbb{D}} \leftarrow \theta_{\mathbb{D}} - \eta \delta_{\theta_{\mathbb{D}}} \sum_{(x,y) \in \mathbb{D}} W(x) \cdot CE(f_{\mathbb{D}}(x),y)
        \end{array}
        \right \}
        \label{eq:LFF}
    \end{equation}

    \begin{equation}
        \begin{aligned}
            W(x) = \frac{CE(f(\mathbb{B}), y)}{CE(f(\mathbb{B}), y) + CE(f(\mathbb{D}), y)}
        \end{aligned}
        \label{eq:W}
    \end{equation}

    \subsection{Solutions for Mitigating Hidden Stratification}
    \label{sec:solutions-HS}

    \begin{figure}[!ht]
            \centering
            \includegraphics[width=\textwidth]{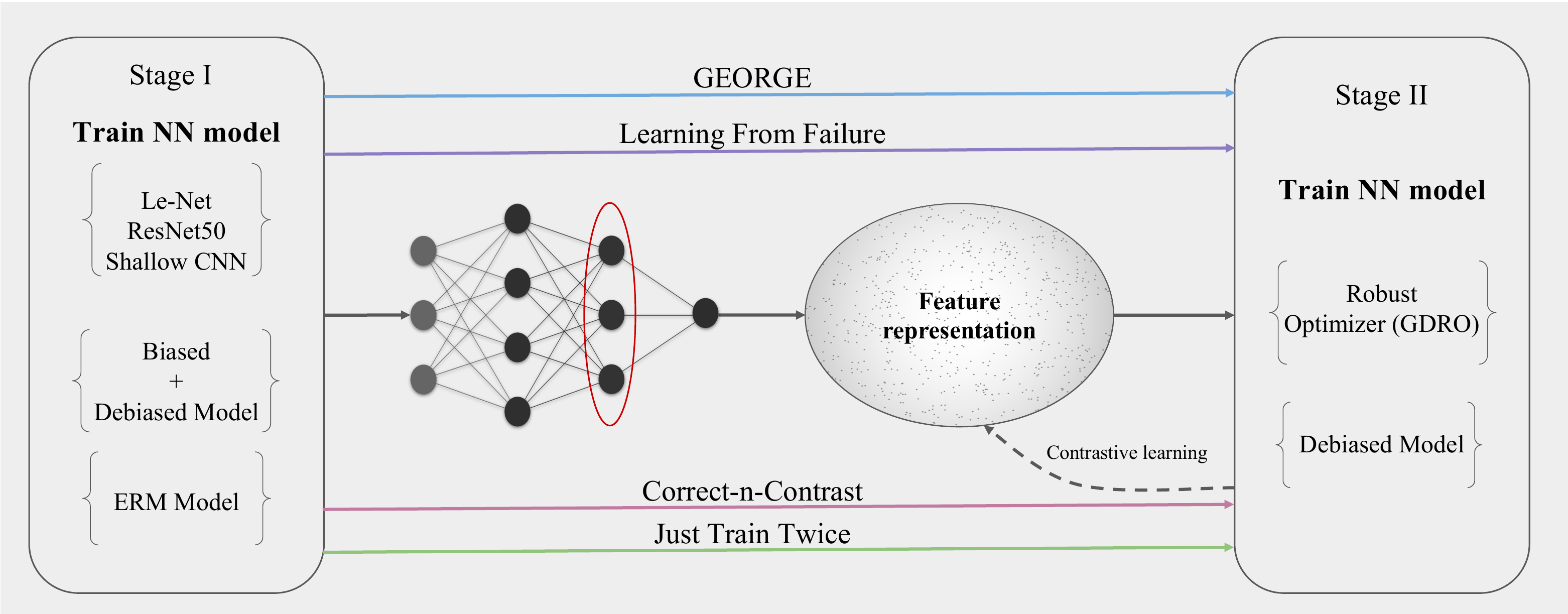}
            \caption{\label{fig:HSRemedies}\textit{A common block diagram to demonstrate different methods that are detecting, tackling Hidden Stratification, Spurious Correlations in a two stage process. 
            }}
    \end{figure}
    
    \subsubsection{Optimization Centric}\hfill
    
    As Chen Dan \cite{dan2021statistical} covers in their dissertation, In a domain-oblivious setting, it is not possible to compute the worst-case risk for minimization as we won't have access to the different domains $\mathcal{D}^{0}, \mathcal{D}^{1}, \mathcal{D}^{2}, ..., \mathcal{D}^n$. Given a subgroup domain $d \in \mathcal{D}$, the DRO tries to maximize the worst-case subgroup performance.

    \paragraph{Distributionally Robust Optimization (DRO)}~ As the model can easily latch onto spurious correlations, and the models that are on average performing well have a high worst group error. The goal of the Group-DRO is to lower the worst group error. Considering the waterbird-landbird (as explained above) and human hair examples from \textbf{Figure~\ref{fig:spurious_corr}}, where the person having hair color type (ex. blonde,dark), model can pick up spurious associations with demographics (ex. gender). Methods that branch out form GDRO \cite{GDRO} such as (\cite{overparameterization}, \cite{GEORGE}, \cite{pmlr-v139-liu21f}, \cite{selective-classification}) aims to directly minimize the worst-group loss as in Equation~\ref{eq:GDRO}.
    \begin{equation}
        \begin{aligned}
            \mathcal{R}_{\text{GDRO}}(\theta) = \text{max}_{\hat{g}\in G}\mathbb{E}_{g,x,y}~[l(\theta;(x,y))|g=\hat{g}]
            \label{eq:GDRO}
        \end{aligned}
    \end{equation}
    In case of Parametric Distributionally Robust Optimization (P-DRO) \cite{michel2021modeling}, it approximates the distribution $P$ with a parametric family $\hat{Q}_\theta$, striking a balance between complexity and model error (\ref{eq:DRO-obj}). It improves generalization under distribution shift and serves as a hedge against model mis-specification. While suitable for GDRO's worst-group loss objective, P-DRO requires dealing with computation challenges arising from discretization or sampling of continuous distributions.
    \begin{equation}\label{eq:DRO-obj}
        \min_{x \in \mathcal{X}} \left\{ \hat{Z}(x) := \max_{\P \in \mathcal{A}} \mathbb{E}_{\P}[h(x;\xi)]\right\}
    \end{equation}

    where $\mathcal{A}$ is commonly set to $\mathcal{A}(\hat{Q}\theta; d, \epsilon) = {P: d(P, \hat{Q}\theta) \leq \epsilon}$ and $d$ refers to Wasserstein distance. The worst-case nature provides a hedge against model mis-specification. By balancing parametric complexity and model error, P-DRO can improve generalization. Recently, a new approach called RP-DRO (Ratio-based Parametric DRO) \cite{rp-dro}. This new approach modifies the P-DRO \cite{michel2021modeling} algorithm by parametrizing the likelihood ratio between the training distribution and the worst-case distribution. This significant change eliminates the need for an unwieldy generative model, making the method more versatile and applicable to a broader range of scenarios.

    Moreover, another recent method, Bayesian Distributionally Robust Optimization (DRBO) \cite{pmlr-v108-kirschner20a} combines Bayesian estimation of a parametric family $f(\theta)$ with DRO to handle model uncertainty. By minimizing the worst-case expected cost over an ambiguity set centered on the Bayesian posterior, DRBO provides a principled way to estimate the distribution from limited data. However, solving the nested robust optimization problem in DRBO poses computational challenges for its application in GDRO.
    \begin{equation}
        \min_{x \in \mathcal{X}} \mathbb{E}_{\theta_N} \left[ \sup_{Q \in \mathcal{M}^\theta} \mathbb{E}_{Q|\theta}[G(x, \xi)] \right],
    \end{equation}
    where $\mathcal{M}_\theta$ is an ambiguity set around $f(\theta)$. Bayesian-DRO trades off posterior mean and variability.

    \paragraph{Leaving no subclasses behind}~ Sohoni et al. \cite{GEORGE} proposed a method \texttt{GEORGE} that helps in measure and mitigate hidden stratification. The strongest novelty of the proposed method is that even when the subclass labels are unknown it will help mitigate hidden stratification during training. Authors test their proposed training strategy on four different image classification datasets that are a mix of real-world and benchmark datasets. Authors explains how an hidden stratification might occur in a dataset. Typically, there are subclasses that are unlabeled in a labelled superclasses that are slightly deviated from the main superclass distribution. They don't contribute to the loss for that superclass and hence the model easily misclassify them. For such cases, we typically always minimize the worst group loss to get everything correctly classified. For a simple ``dog'' vs. ``cat'' classification, these coarse class labels might not capture other salient characteristics (color, size, breed, etc.); these characteristics might be seen as latent variables signifying other subclasses. So, as the latent space of the superclasses and their subclasses is not known, they leverage the feature respresentation (with gaussian assumption; $P(V_{i}|Z)~\text{where,~} V \rightarrow \text{latent vector, i} \rightarrow \text{attribute,} ~\& ~Z \rightarrow \text{subclass}$) by training a neural network, now using the feature respresentation GEORGE \cite{GEORGE} forms clusters. Since subclasses are not known, each cluster is treated as a different subclass. Finally, using the robust optimization from GDRO \cite{GDRO} train a new classifier with better worst case performance.\newline

    \paragraph{Magnifying Disparities by selectively classifying Groups}~ Jones et al. \cite{selective-classification} find that selective classification magnifies accuracy discrepancies between subclasses in the presence of spurious correlations. Authors point out how group disparities are problematic in high-stakes areas like medicine \cite{Chen2021-oq} and criminal justice \cite{hill_2020}, where we would want to consider deploying selective classification. The analysis in this work uses the margin distribution of selective classifiers with thresholds for abstaining. By computing the remaining percentage of the distribution that is correct, the authors show that the model simply abstains from an alarming number of incorrect predictions. A model's accuracy at ``full coverage'' (Meaning the model never abstains) is a strong indicator of how drastic the disparity between subclasses will be. The authors suggest using models that are unbiased and do not discriminate between subclasses to mitigate this issue, and they use ``Group Distributionally Robust Optimization'' (GDRO) \cite{GDRO} to achieve this.

    \subsubsection{Representation Centric}\hfill
    
    Representation learning in subgroup distribution shift refers to learning a transformation of the input data that is robust to changes in the subgroup distribution. The goal is to learn a representation that captures the underlying structure of the data across different subgroups while also being invariant to changes in subgroup-specific properties. Discovering such a representation is essential because, in real-world scenarios, the data may come from different subgroups with different statistical properties. A model that is not invariant to these changes may not generalize well.\newline
    
    \paragraph{Adapting to Latent Sub-group Shift via Concept \& Proxies}~ Recent work has explored ways to make predictive models more interpretable and robust to distribution shifts. For unsupervised domain adaptation where source and target domains diverge due to a latent subgroup distribution shift, if subgroup confounds all observed data, covariate and label shift assumptions become inapplicable \cite{alabdulmohsin2022adapting}. \citeauthor{alabdulmohsin2022adapting} propose an approach for unsupervised domain adaptation that handles shifts stemming from changes in the distribution of a latent subgroup between domains. Their method relies on having access to high-level concept variables $C$ and proxy variables $W$ in the source domain that provide information about the latent subgroup $g \in G$. It identifies the optimal predictor in the target domain $q(Y|X)$ by expressing it in terms of source domain distributions and the marginal input distribution in the target:
    \begin{equation}
        q(Y|X) \propto \sum^{k_{g}}_{i=1}~p(Y|X,~g = i)\cdot p(g=i|X)~\frac{q(g=i)}{p(g=i)}
    \end{equation}
    Alternatively, \citeauthor{pmlr-v119-koh20a} introduced concept bottleneck models that first predict human-defined concepts $C$ from the input $X$ before predicting the target $Y$: 
    \begin{equation}
        \left.
        \begin{array}{ll}
        \hat{q} ~=~ \textit{argmin} \sum_{i} L_{y}(f(q(x^{i}));~ y^{i}) \\
        \hat{f} ~=~ \textit{argmin} \sum_{i} L_{y}(f(\hat{q}(x^{i}));~ y^{i})
        \end{array}
        \right \}
    \end{equation}
    where, $q \rightarrow$ maps $x$ to concept space, and $f \rightarrow$ maps concept to final prediction. This allows test-time intervention on concepts to improve accuracy. Concept bottlenecks can match standard models in accuracy while being more interpretable. Both papers aim to leverage auxiliary information and structure to make models more useful in practice. The concept bottleneck paper demonstrates strong empirical performance, while \citeauthor{alabdulmohsin2022adapting} provide theoretical identification results. An interesting direction would be to explore combining these approaches, using concept bottlenecks that are robust to subgroup shifts via the adaptation approach of \cite{alabdulmohsin2022adapting}. Moreover, \citeauthor{oikarinen2023labelfree} present a method to create concept bottleneck models without needing annotated concept data. They generate an initial concept set using GPT-3, filter the concepts, then learn a projection to align backbone model activations with concepts using CLIP embeddings. The resulting label-free concept bottleneck models achieve high accuracy and interpretability on ImageNet scale datasets.

    \paragraph{Overparameterization causing Spurious Correlation}
    \begin{enumerate}
        \item Complex models \& Simpler data: When the model is overparamaterized with respect the data it is trained on, it can easily memorize the training samples.
        \item Imbalance in subclass: A dataset for a superclass downstream task having an imbalanced ratio of subclass inside superclasses.
    \end{enumerate}
    Sagawa et al., \cite{overparameterization} shows that worst-group error is negatively impacted by overparameterization on real datasets with spurious correlations. The key observation of this paper is that overparameterized models tend to learn spurious correlations, and simply ``memorize'' the minority groups that contradict the spurious correlations, thus obtaining 0\% training loss. To further demonstrate this concept, the authors construct a synthetic dataset containing a majority group with spurious correlations, and a minority group that contradicts. The authors' experiments show that relative size of majority and minority groups is a major factor in the poor performance of overparameterized models. Underparameterized models are also affected by this issue, though less so. When the majority and minority groups are roughly equivalent in size, overparameterized models actually outperform all others. Thus, balanced data is a possible solution in the presence of spurious correlation-related issues. 
    Another important factor is the informativeness of the spurious correlation features. If the features used by the spurious correlations coincide with the same labels more, the overparameterized models will tend to rely on them more, causing testing errors. Unlike the majority/minority ratios, underparameterized models are largely unaffected by spurious correlation informativeness. 
    Lastly, the authors explore two remedies for spurious correlations with overparameterized models: Sub-sampling and re-weighting. The authors determine that while both of these techniques are generally seen as equivalent, sub-sampling is more effective with overparameterized models than re-weighting. This is because sub-sampling reduces the majority-minority ratio, which impacts overparameterized models more strongly.\newline

    \paragraph{Contrastive learning}~ The aforementioned methods: \texttt{GEORGE} \cite{GEORGE} and Selective Classification \cite{selective-classification} leverage the group-DRO in their pipeline to mitigate spurious correlation. However, gathering such data could be costly, and we might not be aware of the spurious attributes in advance in a specific dataset. On the other hand, there are approaches under constrastive learning that intuitively learn the representation by contrast. Fundamentally, the approach works by learning to predict whether two inputs are ``similar'' or ``dissimilar'' by training an encoder to simultaneously maximize the similarity between the feature representations of anchor and positive data-points, while minimizing the similarity between anchor and negative representations \cite{phuc2020khac}. Unsupervised CL typically employs uniform sampling to obtain ``negatives'' \cite{bachman2019learning}, and generates ``positives'' by augmenting the input data with various views of the same object (e.g., via data augmentation) \cite{chen2020simple}. The proposed model CNC \cite{pmlr-v162-zhang22z} focuses on robustness to spurious correlations. However, this approach sets the initial loss objective (preliminary setting) similar to GDRO. The works' focus is upon aiming to close the robustness gap of a trained ERM model by improving the learned representations of the model in the second stage. It accomplishes this by ``drawing together'' data points of the same class and ``pushing apart'' data points of different classes, independent of their individual groupings or spurious features. Similar to manifold representation learning for adversarial attacks in \cite{adhikarla2022memory}. Contrasting learning (eq. \ref{eq:contrastive-learning}) defines the central objective for the correct-n-contrast (CNC) approach. 
    \begin{equation}
        \begin{aligned}
            \mathcal{L}_{\text{con}}^{\text{sup}}(x;h_{\text{enc}}) = 
            \mathbb{\hat{E}}_{\{x_{i}^{+}\}_{i=1}^{M},\{x_{i}^{+}\}_{j=1}^{N} }
            -\text{log}\left[\frac{exp(z^{\top}z_{i}^{+}/\top)}{\sum_{m=1}^{M}exp(z^{\top}z_{m}^{+}/\top) \sum_{n=1}^{N}exp(z^{\top}z_{n}^{-}/\top)}\right]
        \end{aligned}
        \label{eq:contrastive-learning}
    \end{equation}
     Authors prove that the gap b/w worst group and average loss is upper bounded by class respective alignment loss of the representation alignment. Overall, contrastive learning overlaps to the idea of triplet loss. However, this approach suffers from hard negatives which arises when the selected negative examples are too easy to distinguish from the positive examples, resulting in a loss that does not effectively update the model's parameters \cite{NEURIPS2020_63c3ddcc, wu2021conditional, robinson2021contrastive, chen2022incremental}.\newline

    \paragraph{Tuning Empirical Risk Minimization (ERM)}~ Liu et al. \cite{pmlr-v139-liu21f} introduce a technique known as ``Just Train Twice'' (JTT) for improving performance in the face of spurious correlations. The authors use a two-step training procedure: First, a simple Empirical Risk Mimization (ERM) model is trained on a dataset known to contain spurious correlations. Then, the authors highlight errors the model makes during testing, and use an upweighting function ($\lambda_{\text{up}}$ in the equation shown below) to force the classifier to learn not to rely on spurious correlations.
    \begin{equation}
        \begin{aligned}
            \text{J}_{\text{UP-ERM}}(\theta, Er) = 
            \left(\lambda_{\text{up}}\sum_{(x,y) \in Er} l(x,y;\theta) + \sum_{(x,y) \notin Er} l(x,y;\theta)\right)
        \end{aligned}
        \label{eq:jtt}
    \end{equation}
    JTT is evaluated on both NLP and Computer Vision tasks, using Waterbirds and CelebA for vision, and MultiNLI and CivilComments-WILDS for NLP \cite{wilds}. CVaR DRO \cite{yang2021distributionally} is a conceptually similar technique to JTT, but Liu et al. show that JTT outperforms CVaR DRO significantly. The key difference is that JTT holds the error points fixed for the second training phase, whereas CVaR DRO recomputes them continuously.

    Lastly, the authors experiments all have a focus on spurious correlations. However, they suggest that due to JTT's simplicity, it may be worthwhile to experiment with different types of distribution shifts. It is possible that JTT's applications generalize beyond solely spurious correlations.



\begin{table}
    \centering
    \tiny 
    \caption{\label{tab:subgroup-dataset-summary}The following table summarizes the datasets used by different research articles for experiments on subgroup distribution shifts.}
    \begin{threeparttable}
    \resizebox{\columnwidth}{!}{ 
    \begin{tabular}{p{1.2in} p{0.5in} p{2.4in} p{1in} p{1in} c p{0.5in}}
    \toprule
    \rowcolor{gray!15}

    \textbf{DATASETS} & \textbf{SOURCE} & \textbf{DESCRIPTION} & \textbf{TYPE} & \textbf{TRAIN/TEST IMAGES} & \textbf{YEAR} & \textbf{LINK} \\ \midrule

    \rowcolor{gray!5}
    CXR14 \cite{Wang2017ChestXRay8HC} \tnote{1}
      & CVPR
      & Frontal-view chest X-ray PNG images in 1024$\times$1024 resolution.
      & 14 classes
      & 112,120 (30,805 unique patients)
      & 2017 
      & \href{https://nihcc.app.box.com/v/ChestXray-NIHCC}{link} \\ \midrule

    CelebA \cite{liu2015faceattributes} 
       & ICCV
       & A facial attribute dataset containing information of 10,177 celebrities. Images of the size 178×218. 
       & 40 attributes, 5 landmark
       & 202,599 
       & 2015 
       & \href{http://mmlab.ie.cuhk.edu.hk/projects/CelebA.html}{link} \\ \midrule

    \rowcolor{gray!5}
    CheXpert-device \cite{selective-classification} \tnote{2}
      & AAAI 
      & Presents both frontal \& lateral views. The task is to do automated chest-x-ray interpretation.
      & 14 observations 
      & 224,316 (65,240 unique patients)
      & 2019 
      & \href{https://stanfordmlgroup.github.io/competitions/chexpert/}{link} \\ \midrule

    MIMIC-CXR \cite{mimic-cxr} \tnote{3}
      & Sci. Data
      & Extra dataset from MIT that provides frontal \& lateral view of Chest-X-Ray images. ChestXRay \& MIMIC-CXR datasets share a common annotator tool.
      & 14 labels
      & 371,920
      & 2019 
      & \href{https://stanfordmlgroup.github.io/competitions/chexpert/}{link} \\ \midrule
    
    \rowcolor{gray!5}
    Waterbirds derived from CUD \& Places \cite{GDRO}
      & ICLR 
      & 95\% of all waterbirds against a water background and the remaining 5\% against a land background and vice-a-versa. 
      & 2 classes \tnote{4} 
      & 4795 
      & 2020 
      & \href {https://nlp.stanford.edu/data/dro/waterbird_complete95_forest2water2.tar.gz}{link} \\ \midrule

    Camelyon \cite{bandi2018detection} \tnote{5}
      & IEEE TMI
      & 1399 lymph node WSIs w/ \& w/o tumor cell labels. Contains 4 different kinds of tumor cells, both with and without metastases.
      & 3 classes\tnote{6}
      & 1399
      & 2017
      & \href{https://camelyon17.grand-challenge.org/Data/}{link} \\ \midrule

    \rowcolor{gray!5}
    Patch-Camelyon \cite{bandi2018detection}
      & MICCAI
      & (96x96)px extracted from histopathology scans of lymph node sections.
      & 2 classes 
      & 262,144/32,768/32,768
      & 2017
      & \href{https://patchcamelyon.grand-challenge.org/}{link} \\ \midrule

    DomainNet \cite{peng2019moment}
      & ICCV
      & The dataset consists of images from six distinct domains, photos, sketch, clipart, painting, quickdraw, and infograph.
      & 345 classes 
      & 600K (48K - 172K per domain)
      & 2019 
      & \href{http://ai.bu.edu/DomainNet/}{link} \\ \midrule

    \rowcolor{gray!5}    
    Undersampled-MNIST \cite{GEORGE}
      & NeurIPS 
      & Modified MNIST with task to classify ``<5'' \& ``$\leq 5$'' with digits 0 to 9 representing subclasses.
      & 10 classes 
      & 60,000/10,000 
      & 2020 
      & - \\ \midrule 

    ISIC \cite{isic} 
       & NeurIPS 
       & Skin cancer dataset for classifying skin lesions as `malignant' or `benign'. Images are provided in 1024x1024px.
       & 2 classes
       & 33,126/10,982
       & 2019
       & \href{https://challenge.isic-archive.com}{link} \\ \midrule

    \rowcolor{gray!5}
    BREEDs \cite{BREEDS} 
       & ICLR
       & BREEDs provides a range of shifts, including expected shifts such as domain adaptation and data imbalance, and unexpected shifts such as weather and camera changes.
       & 1K classes
       & 1,281,167/50K/100K
       & 2021
       & \href{https://github.com/MadryLab/BREEDS-Benchmarks}{link} \\ \midrule

    CIFAR-10 \cite{cifar10-cifar100}
       & U. of Toronto
       & CIFAR-10 is a benchmark image classification dataset of 32x32 RGB images with 6K images per class.
       & 10 classes
       & 50,000/10,000
       & 2009 
       & \href{https://www.cs.toronto.edu/~kriz/cifar.html}{link} \\ \midrule

    \rowcolor{gray!5}
    CIFAR-100 \cite{cifar10-cifar100}
       & U. of Toronto
       & A harder version of CIFAR-10 dataset consisting of 32x32 RGB images with 600 images per class.
       & 100 classes
       & 50,000/10,000
       & 2009 
       & \href{https://www.cs.toronto.edu/~kriz/cifar.html}{link} \\ \midrule

    LSUN (C), LSUN (R) \cite{yu15lsun} \tnote{7}
       & ECCV
       & Dataset of scenes for scene recognition and detection tasks.
       & 10 classes
       & 3,000,000/100,000
       & 2016 
       & \href{https://www.yf.io/p/lsun}{link} \\ \midrule

    \rowcolor{gray!5}
    iSUN \cite{xu2014isc}
       & CVPR
       & Large scale eye-tracking dataset of natural scene images from SUN database \cite{Xiao2010SUNDL}
       & 2333 classes\tnote{8}
       & 20,608
       & 2014
       & \href{https://www.cs.cmu.edu/~hrs/datasets.html}{link} \\ \midrule

    SVHN \cite{netzer2011reading}
       & NeurIPS W.
       & Real-world dataset with order of magnitude higher than MNIST for recognizing digits and numbers in natural scene images.
       & 10 classes
       & 73,257/26,032
       & 2011 
       & \href{http://ufldl.stanford.edu/housenumbers/}{link} \\ \midrule

    \rowcolor{gray!5}
    TinyImageNet \cite{tiny-imagenet}
       & ImageNet
       & Subset of ImageNet with 30 classes
       & 200 classes
       & 100K/10K/10K
       & 2015
       & \href{https://tiny-imagenet.herokuapp.com/}{link} \\ \midrule

    ImageNet-30 \cite{NEURIPS2019_a2b15837}
       & Univ. NC
       & Subset of ImageNet with 30 classes
       & 30 classes
       & 1200/50/1500
       & 2020
       & \href{http://vision.cs.unc.edu/datasets/imagenet30/}{link} \\ \midrule

    \rowcolor{gray!5}
    WILDS-iWildCam \cite{beery2020iwildcam}
       & CVPR
       & Multi-class classification of animal species from camera trap images with domain shifts in time, geographic location, and camera type
       & 11 classes 
       & 1,771,687/25,000/62,562
       & 2020 
       & \href{https://wilds.stanford.edu/datasets/\#iwildcam}{link} \\ \midrule

    WILDS-Camelyon 17 \cite{bandi2018detection}
       & IEEE T MED
       & Binary classification of metastatic tissue in whole-slide histopathology images of lymph nodes
       & 2 classes 
       & 270/70/130
       & 2018
       & \href{https://wilds.stanford.edu/datasets/\#camelyon17}{link} \\ \midrule

    \rowcolor{gray!5}
    WILDS-RxRx1 \cite{taylor2019rxrx1}
       & ICLR
       & The dataset provides the study of shift induced by batch effects on a variant of RxRx1 dataset.
       & -
       & -
       & 2019 
       & \href{https://wilds.stanford.edu/datasets/\#rxrx1}{link} \\ \midrule

    WILDS-FMoW \cite{christie2018functional}
       & CVPR
       & Multi-class classification of aerial imagery with domain shifts in image resolution, zoom level, time of day, and weather
       & 63 object classes
       & 1,372,266/17,153/68,995
       & 2018
       & \href{https://wilds.stanford.edu/datasets/\#fmow}{link} \\ \midrule

    \rowcolor{gray!5}
    WILDS-PovertyMap \cite{yeh2020using}
       & Nature
       & Binary classification of poverty in satellite imagery with domain shifts in geographic location, season, and year
       & 2 classes 
       & 5,000,000/50,000/100,000
       & 2020 
       & \href{https://wilds.stanford.edu/datasets/\#povertymap}{link} \\ \midrule

    WILDS-GlobalWheat \cite{david2020global}
       & Plant Phenomics
       & Global Wheat Detection (GWD) is an object detection dataset consisting of aerial images of wheat crops captured around the world.
       & Multiple wheat heads
       & 3315/1000
       & 2020
       & \href{https://wilds.stanford.edu/datasets/\#globalwheat}{link} \\ \midrule

    \rowcolor{gray!5}
    WILDS-OGB-MoIPCBA \cite{hu2020open}
       & NeurIPS
       & It is a molecular property prediction dataset with a binary classification task. The molecules are represented as graphs with nodes and edges.
       & 2 classes 
       & 841,050/105,131/105,156
       & 2020 
       & \href{https://wilds.stanford.edu/datasets/\#ogb-molpcba}{link} \\ \midrule

    WILDS-CivilComments \cite{borkan2019nuanced}
       & WWW
       & Text classification task to predict upvote count for comments on the Civil Comments platform.
       & \textit{N.A}
       & \textit{N.A}
       & 2019
       & \href{https://wilds.stanford.edu/datasets/\#civilcomments}{link} \\ \midrule

    \rowcolor{gray!5}
    WILDS-Amazon \cite{ni2019justifying}
       & EMNLP
       & The goal is to predict whether a review is classified as "positive" or "negative", while also ensuring that model performance does not significantly vary across different demographic subgroups.
       & \textit{N.A}
       & \textit{N.A}
       & 2020 
       & \href{https://wilds.stanford.edu/datasets/\#amazon}{link} \\ \midrule

    WILDS-Py150 \cite{lu2021codexglue}
       & arXiv
       & WILDS-Py150 is a code classification dataset consisting of 150 programming languages for multi-class classification tasks.
       & 150 classes 
       & \textit{N.A}
       & 2021 
       & \href{https://wilds.stanford.edu/datasets/\#py150}{link} \\ \midrule

    \rowcolor{gray!5}
    Biased Action Recognition (BAR) 
       & NeurIPS
       & Real-world image dataset of six action classes (eating, smoking, taking selfie, talking on the phone, reading, and working on a computer) biased to distinct places.
       & 6 classes
       & 7861/970/970
       & 2020 
       & \href{https://github.com/alinlab/BAR}{link}
       \\ \midrule
    
    MultiNLI \cite{N18-1101}
       & NAACL-HTL
       & The MultiNLI dataset is a compilation of 433K sentence pairs that have been annotated with information on textual entailment and sourced from multiple genres.
       & 3 classes
       & 392,702/20K/20K
       & 2018
       & \href{https://cims.nyu.edu/~sbowman/multinli/}{link} \\ 
       \bottomrule
    \end{tabular}}

    \begin{tablenotes} 
        \item[1] The image labels are NLP extracted so there would be some erroneous labels but the NLP labelling accuracy is estimated to be >90\%.
        \item[2] On-going competition. Datasets collected b/w Oct. 2002 and Jul. 2017. 
        \item[3] Beth Israel Deaconess Medical Center between 2011 - 2016.
        \item[4] Waterbird~\{seabird:11;~waterfowl:6\} or Landbird
        \item[5] Part of second grand challenge Camelyon17, 2017 by Radbound University Medical Center.
        \item[6] Classes each w/ 8 stages; Staging of breast cancer is based on the 7th edition of the TNM staging criteria.
        \item[7] A 32x32 image crop that is represented by ``(C)'' and a resizing of the images to 32x32 pixels that termed by ``(R)''. 
        \item[8] 397 categories.
    \end{tablenotes}
    \end{threeparttable}
\end{table}

\section{Adversarial Distribution Shift}
\label{sec:AP}

    In this section, we summarize different kinds of data-centric approaches that have gained higher impact (performance and popularity) in the literature and the comparison of model-centric approaches with data-centric approaches. 
    Adversarial distribution shift is another crucial problem in the field of data science and machine learning, when the distribution of the training data is maliciously made different from the distribution of the test data by the attackers, causing adversarial shift or misrepresentation of the actual data, which confuse model to make misleading prediction. In recent years, deep neural network (DNN) models have outperformed humans in various tasks, but they are vulnerable to adversarial perturbations, which are carefully crafted noise added to natural datasets \cite{SzegedyZSBEGF13}. Adversarial distribution shift on sensitive data, such as finance \cite{fursov2021adversarial, goldblum2021adversarial}, autonomous  \cite{robust-physical-world-attack, adv-semantic-image-segmentation, adv-sensor-attack, evaluating-adversarial-attacks-driving-safety}, and healthcare \cite{champneys2021vulnerability, an2019longitudinal, finlayson2019adversarial}, can have serious consequences, even when it affects only a subset of the data. The susceptibility of deep learning models to adversarial distribution shifts, which can be triggered by data harvesting bots collecting faulty data and modifying datasets, emphasizes the need for robust models. This raises concerns about the credibility of deployed models, as there is a risk of compromised system outputs.Conventional security measures, such as firewalls, passwords, and encryption, are not sufficient to protect against adversarial attacks.
    \begin{figure}[htbp!]
        \centering
        \includegraphics[width=0.75\textwidth]{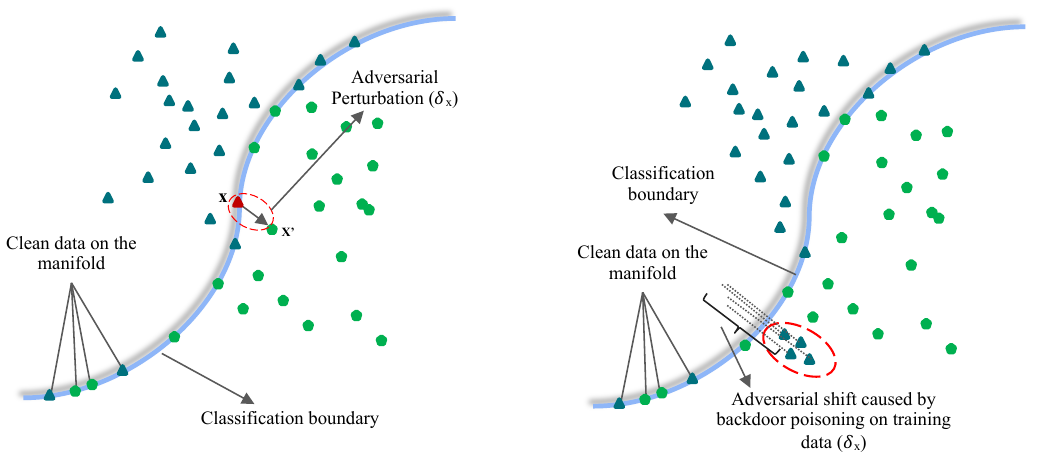}
    \caption{\label{fig:attack}\textbf{[Left]}~\textit{The adversarial attack objective for infusing a perturbation $\delta$ in a benign input $x$.}\textbf{[Right]}~\textit{Malicious actor influences the training data to cause the adversarial shift, and to cause the model to underperform during deployment.}}
    \end{figure}

    \subsection{The Existence of Adversarial Perturbations and Their Taxonomies}

    In this subsection, we define the fundamentals of the attackers perspective, however to keep the survey concise, we will exclude detailed summary of the attack section, and we will mainly focus on the defense scenarios aimed at achieving robustness against adversarial shift. There are other well summarized surveys to go deeper in the deep learning model attack section of research. We have curated a list of some popular surveys (\cite{adv-attacks-survey-1,adv-attacks-survey-2,evaluating-adversarial-attacks-driving-safety,adv-attacks-survey-4,adv-attacks-survey-5,adv-attacks-survey-6}) for the attack summaries that will be helpful to get started. 

    According to Machado et al. \cite{adv-attacks-survey-1}, there exist various types of  \textit{perturbation scopes}, \textit{perturbation visibility}, \textit{perturbation limitation}, and \textit{perturbation measurement metrics} and other additional variables that we think should be a part of variables that an adversarial attacker may consider:

    \begin{enumerate}
        \item \textit{Scope.} [\textcolor{orange}{\textbf{Data-centric}}]\quad The scope of the perturbation can be defined in two ways: first, by attacking a single image at a time (individually scoped), and second, by perturbations that are agnostic to the input images and are produced independently from any input sample. Overall, the goal of both kinds of scopes is to pertain the ability to cause models to miscalssify. In the real world circumstances, universally produced perturbations are much easier.

        \item \textit{Visibility.} [\textcolor{orange}{\textbf{Data-centric}}]\quad Perturbations here can be again divided into two categories: First, human imperceptible perturbations. It depends on the attacker's goal. The perturbations can be imperceptible and be insufficient to misclassify a model. On the other hand, perturbations can be optimally tuned to misclassify and be human imperceptible. Second, physically visible perturbations that are physically added to the image/object. This type of perturbation is mainly found in fooling object detection models \cite{visible-perturbation-1,visible-perturbation-2}. 
    
        \item \textit{Limitation.} [\textcolor{purple}{\textbf{Model-centric}}]\quad The limitations on perturbations can be set based on the given goal by the attacker. The optimization perturbation will have a optimization goal to misclassify the input example. Moreover, a constraint perturbation would place constraints over the optimization problem.
    
        \item \textit{Metrics.} [\textcolor{purple}{\textbf{Model-centric}}]\quad As we describe in Equation~\ref{eq:norm}, it is measuring the magnitude of the perturbation using the distance metric. 
        
        \item \textit{Data availability.} [\textcolor{purple}{\textbf{Model-centric}}]\quad Attackers may have different levels of access to the training data and/or the model architecture, which can impact the type and effectiveness of their attacks. For example, a white-box attacker has full access to the model and its parameters, whereas a black-box attacker only has access to the model's input and output.

        \item \textit{Attack goal.} [\textcolor{orange}{\textbf{Data-centric}}]\quad The goal of an attacker can also vary, and can impact the type of perturbations they generate. For example, an attacker may aim to cause a model to misclassify an image in a specific way (targeted attack), or simply to cause any kind of misclassification (untargeted attack).
    
        \item \textit{Adversarial examples transferability.} [\textcolor{purple}{\textbf{Model-centric}}]\quad An adversarial example is said to be transferable if it can fool multiple models trained on different datasets or architectures. Attackers may consider transferability when crafting their perturbations to increase the scope and impact of their attacks.
    
        \item \textit{Adversarial examples perceptibility.} [\textcolor{orange}{\textbf{Data-centric}}]\quad An attacker may consider the perceptibility of their perturbations, as overly noticeable perturbations may be detected and flagged by security measures. A popular approach is to generate imperceptible perturbations that are still able to fool the model.
    
        \item \textit{Adversary's knowledge.} [\textcolor{purple}{\textbf{Model-centric}}]\quad The level of knowledge an attacker has about the target model can also impact the type of attacks they generate. For example, an attacker with little knowledge of the model may opt for a black-box attack, whereas an attacker with full knowledge of the model may opt for a more sophisticated white-box attack.\newline
    \end{enumerate}

    \begingroup
        \leftskip=0cm plus 0.5fil \rightskip=0cm plus -0.5fil
        \parfillskip=0cm plus 1fil
        \textit{\textbf{How can we train deep neural networks that are robust to adversarial shifts?}}\par
    \endgroup

    \begin{figure}[!ht]
        \centering
        \includegraphics[scale=0.75]{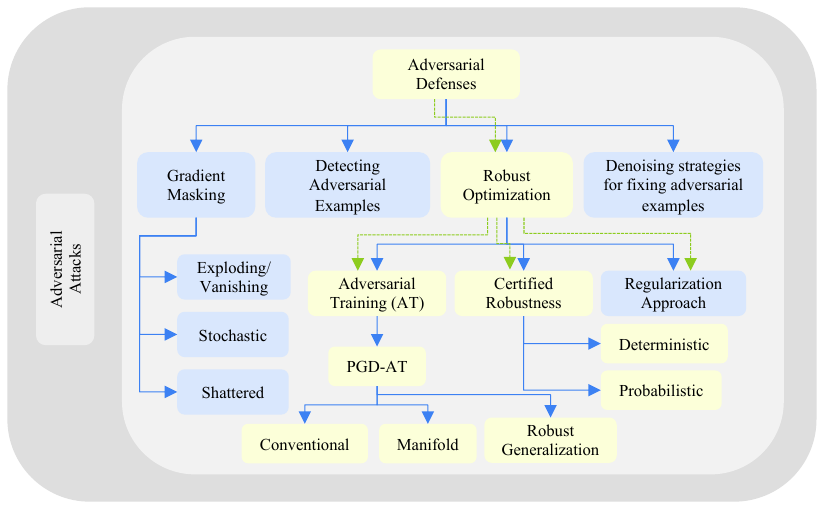}
        \caption{\label{fig:defense-flow}\textit{Outlining the recent literature categorization for adversarial defenses.}}
    \end{figure}

    \textbf{Figure~\ref{fig:defense-flow}} demonstrates the classification of popularly studies sub-fields towards achieving adversarial defense (an updated version of \cite{adv-robustness-survey-1}). It has been seen in most of the literature that methods that perform adversarial training are able to sustain model robustness against adversarial examples. Below we talk about AT in detail.

    \subsection{Empirical Defense - Adversarial Training (AT)}\hfill
    \begin{center}
        \textit{Minimizing the worst-case loss}
    \end{center}
    Over the last decade, there has been tremendous research conducted for creating defense against adversarial examples \cite{SzegedyZSBEGF13}. This has indeed proven extraordinarily a difficult task. Adversarial training is now used a default training strategy for defense mechanism. It is currently considered one of the most effective methods for improving the adversarial robustness of the neural network.

    In case of adversarial training, the adversarial examples are used during the training phase of the neural network. Mathematically, the task is to solve the inner-maximization problem on Equation~\ref{eq:adversarial-training} and outer minimization problem. The solution to the problem is achieved by approximating inner-maximization
    ${max}~\mathcal{L}_{\text{ce}} (h_{\theta} (x+\delta), y)$
    with the heuristic search, trying to find the optimal lower-bound. The method has shown promising results, even models with complicated settings and deeper networks such as ImageNet, ImageNet-V2, and Vision Transformers. From the perspective of distribution shift, we model the adversarial perturbations around each natural example $x_i$ by a distribution $p(\delta_i)$, whose support is contained in $\mathcal{D}_{\delta}$. 

    \begin{equation}
        \begin{aligned}
            \theta^{*} = \underset{\theta}{min}~\mathbb{E}_{p(\delta_i)}\left[ \underset{p(\delta_i) \in \mathcal{D}_{\delta}}{max} {max}~\mathcal{L}_{\text{ce}} (h_{\theta} (x_i+\delta_i), y_i) \right]\\
        \end{aligned}
        \label{eq:adversarial-training}
    \end{equation}
    
    The min-max game aims to make neural networks more robust to adversarial attacks. The principle idea is to find the perturbation that maximizes the loss function, which means that the added perturbation should ideally confuse the neural network as much as possible. Once this perturbation is found, the neural network is trained using a minimization formula to reduce the loss on the training data, while keeping the perturbation fixed. This way, the model is optimized to be more resilient against future adversarial perturbations. In other words, the goal of this approach is to improve the model's robustness and ability to adapt to perturbations, thus enhancing its overall performance.

    \textbf{PGD universal adversarial training?}~Models developed utilizing projected gradient descent (PGD) adversarial training were demonstrated to be resistant to the most powerful known attacks. Contrary to other defense mechanisms that have been compromised by newly implemented attack strategies. Such as Rice et al. \cite{10.5555/3524938.3525687} showing effectiveness of adversarial training for DNNs to the point where the original PGD based method can achieve comparable robust performance to state-of-the-art methods with Vanilla PGD-based training attaining robust test error rate of 43.2\% against PGD adversary on CIFAR-10 with $l_{\infty}$ radius 8/255 and TRADES \cite{trades} shown similar 43.4\% robust test error for the same adversary.  
    
    Although PGD adversarial training for models have proven to be de facto, there is a serious issue in this training mechanism \cite{croce2020robustbench}.   Equation~\ref{eq:adversarial-training} draws $x, y$ both from the same distribution $\mathbb{D}$, hence the model will be highly robust to attacks produced by that method with which it is trained but has poor generalization for other unseen attacks under the same threat model \cite{songimproving}. For example, recent methods \cite{zhang2019defense, xiaoenhancing} can achieve the state-of-the-art robustness against PGD, but they can still be defeated by others. It indicates that these defenses probably cause gradient masking \cite{tramerensemble, athalye2018obfuscated, uesato2018adversarial}, and can be fooled by stronger or adaptive attacks \cite{adaptive-attacks}. Hence, AT's are computationally expensive, especially when using a large number of iterations or a high-dimensional input space. Additionally, while PGD-AT is robust to known attacks, it may not be effective against novel or sophisticated attacks that have not been seen before. Still, PGD-AT does comparatively better than most of the available training methods. This has lead the community to also focus on addressing issues with PGD-AT. Qian et al., \cite{qian2022survey} also points out, multiple directions in Conventional-AT (Overfitting \cite{10.5555/3524938.3525687}, SLAT \cite{slat}, Free \cite{free}, Fast-AT \cite{DBLP:conf/iclr/WongRK20}), Manifold-AT (TRADES \cite{trades}, MAT \cite{ZhangHZL21}, ATLD \cite{DBLP:journals/corr/abs-2107-04401}), \& Robust Generalization (AWP \cite{WuX020}, SCR \cite{ZhangQHW0Y21}) been investigated. 

    \begin{figure*}
        \centering
        \includegraphics[width=0.9\textwidth]{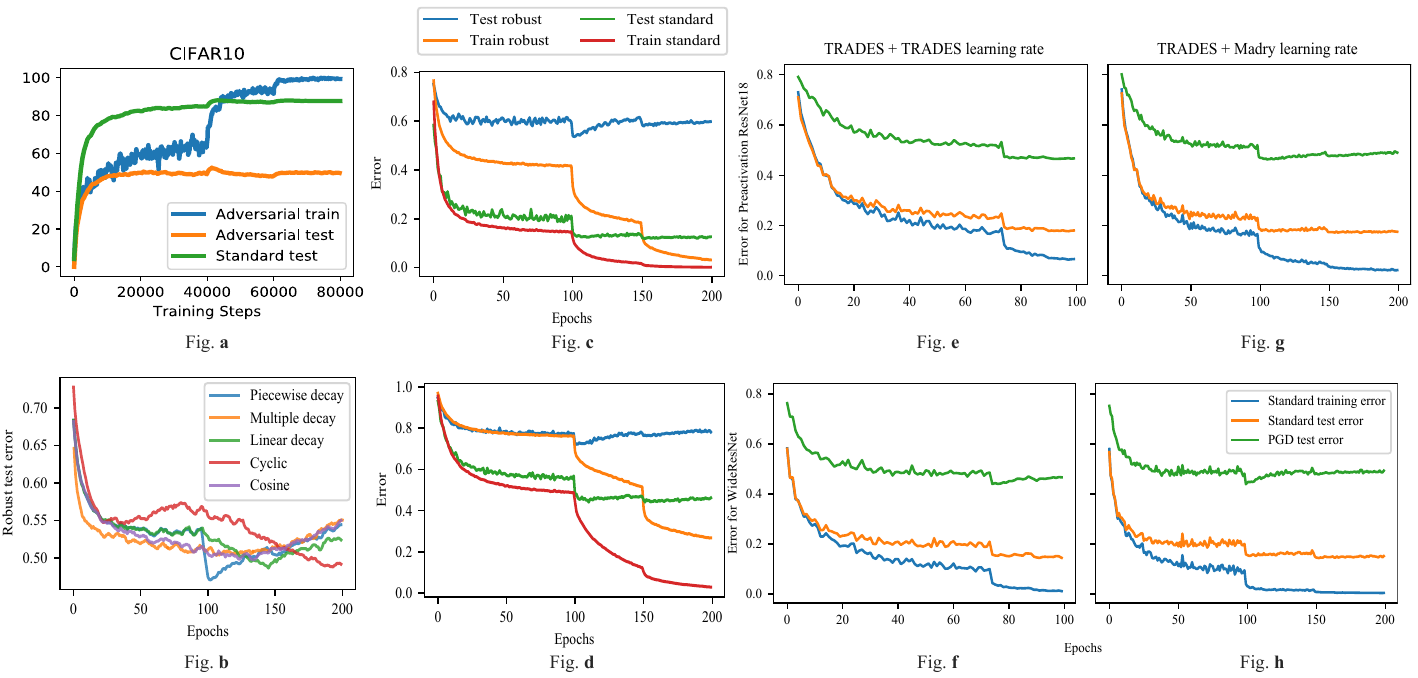}
        \caption{\label{fig:adv-overfitting-trades-regularization}
            \textbf{Subfigures [a-d]}: \textit{Adversarial training cause overfitting, and poor generalization \& worsen standard error.} 
            \textbf{Subfigures [e,f]}: \textit{Figures show the decision boundary during natural training (e) and learned through TRADES adversarial training (f) (taken from \cite{SchmidtSTTM18,10.5555/3524938.3525687,trades}).}
            }
    \end{figure*}

    \subsubsection{Generalization}\hfill

    \textbf{Defenses in Adversarial Overfitting (AOT)}\hfill

    Overfitting has been studied for a very long time from the traditional machine learning to the current deep learning approaches. Researchers have come up with effective countermeasures and model designs to mitigate overfitting. In this subsection, we are looking back into the problem of overfitting in the case of adversarial training. The problem of generalization in adversarial training leads to adversarial overfitting (AOT). \textbf{Figure~\ref{fig:adv-overfitting-trades-regularization}} [a-h] demonstrates the model overfitting in different cases reported by several authors. \textbf{Sub-figure~\hyperref[fig:adv-overfitting-trades-regularization]{(a)}} from Schmidt et al., \cite{SchmidtSTTM18} compares ``How does the sample complexity of standard generalization compare to that of the adversarially robust generalization?''. The figure shows the classification accuracies for the robust optimization of CIFAR-10 trained to be robust for $l_{\inf}$ perturbation. Author in \textbf{Sub-figure~\hyperref[fig:adv-overfitting-trades-regularization]{(a)}} also points out the generalization gap for the robust accuracy which when compared with MNIST got significantly increased for CIFAR-10. To verify that the problem doesn't exist because of a certain learning rate, In \textbf{Sub-figure~\hyperref[fig:adv-overfitting-trades-regularization]{(b)}} from Rice et al., \cite{10.5555/3524938.3525687}, authors provide various learning rate schedules on CIFAR-10. Closely after first decay of learning rate the model is 43.2\% robust error and degrades robustness afterwards. Concluding that none of the learning rate is able to achieve a peak performance to that of standard piece-wise decay learning. \textbf{Sub-figure~\hyperref[fig:adv-overfitting-trades-regularization]{(c, d)}} from Rice et al., \cite{10.5555/3524938.3525687} shows the re-implementation results of Madry et al., \cite{madry2018towards} on CIFAR-10 and CIFAR-100, both showing more robust test error of ~47\% and ~73\% than what is attained at the end of training. Lastly, in \textbf{Sub-figures~\hyperref[fig:adv-overfitting-trades-regularization]{(e-h)}}, authors use TRADES \cite{trades} technique as it has proven to outperform adversarial training. The sub-figure~\hyperref[fig:adv-overfitting-trades-regularization]{(h)} confirms that for achieving robustness they mitigate robust overfitting by early stopping. In order to achieve the reported TRADES result of 43.4\% robust error as it degrades after first decay on PGD-$l_{\inf}$ with 8/255 norm radius. This figure uses Wide-ResNet model on CIFAR-10. As Schmidt et al., \cite{SchmidtSTTM18} suggests, it is crucial to explore and investigate the intersections and gaps between robustness, classifiers and data distribution. Some open problems can be found in \cite{SchmidtSTTM18}. 

    \textbf{More data for Generalization of Adversarial Training?}\hfill

    Rice et al.~\cite{10.5555/3524938.3525687}, Schmidt et al.~\cite{SchmidtSTTM18}, Song et al.~\cite{songimproving}, Madry et al.~\cite{madry2018towards}, have shown the problem of generalization in adversarial training. Generalization can be defined as a vital feature in improving out-of-distribution/unseen examples and in turn also improving robustness. In fact higher sample complexity is the root cause of the difficulty in training robust classifiers, at least partially Schmidt et al.'s, \cite{SchmidtSTTM18} postulate. Their work provides a theoretical proof in order to understand how properties of data affect the number of samples needed for robust generalization. They demonstrated with a Gaussian model setting that the lower bound offers transferable adversarial examples and applies to worst-case distribution shifts without a classifier-adaptive adversary. They also demonstrated with a Bernoulli's model setting that an explicit thresholding layer also significantly reduces the sample complexity when training a robust neural network on MNIST. Using the combination of theoritical proofs and experiments, 
    This means in general, sample complexity gap in a self-supervised model setting with robust self-training (RST) is consistently beneficial. Their experiments on CIFAR-10 and SVHN attained sustainable and consistent improvements in robust accuracy. Equation~\ref{eq:robust-metric-carmon} shows the general error metrics for a self-supervised classification task.
    \begin{gather}
    \label{eq:robust-metric-carmon}
        \text{err}_{\text{standard}}(f_{\theta})~:=~\mathbb{P}_{(x,y)\sim P_{x,y}}(f_{\theta}(x)\neq y)\\
        \text{err}_{robust}^{p,\epsilon}(f_{\theta})~:=~\mathbb{P}_{(x,y)\sim P_{x,y}}(\exists\hat{x}\in\mathcal{B}_{\epsilon}^{p}(x)f_{\theta}(x)\neq y)\} \notag
    \end{gather}
    Equation~\ref{eq:robust-metric-carmon} represents the robust probability error, where $\mathcal{B}_{\epsilon}^{p}(x):=\{\hat{x}\in X| ||x-\hat{x}||_{p}\leq \epsilon$. Overall self-training works in two steps: when algorithm $M$ maps to the dataset (X,Y), to the parameter $\theta$. firstly, capture the intermediate parameters $\hat{\theta}_{intermediate} = M(X,Y)$ that generates pseudo-labels $\hat{y} = f_{\theta}(x_{i})$. Secondly, combine the data and pseudo-labels to obtain a final model $\hat{\theta}_{final} = A([X,\hat{X}][Y,\hat{Y}])$. Noting that hyperparameter tweaking make the task challenging to determine how performance is affected to data volume. Utrera et al. \cite{utrera2021adversariallytrained} in their work closed the gap between transfer learning and adversarial training by proving that adversarially trained deep networks transfer better than non-adversarial/standard trained model. They demonstrate that robust DNNs can be transferred faster and more accurately with fewer images on the target domain. Utrera et al. \cite{utrera2021adversariallytrained} concluded that adversarial training biases learned representations to retain shapes rather than textures, reducing the transferability of source models.

    \subsubsection{Other Improved Defense Variants of AT} \hfill

    Dynamic AdveRsarial Training (DART) et al., \cite{DART} proposes a criterion called First-Order Stationary Condition (FOSC) to evaluate the convergence of adversarial training. FOSC measures the convergence quality of adversarial examples in the inner maximization problem and is well-correlated with adversarial strength, making it a good indicator of robustness.
    
    Friendly Adversarial Training (FAT) et al., \cite{FAT}, the method went in all directions of drawbacks in standard-AT, and offered including better standard accuracy for natural data while maintaining competitive robust accuracy for adversarial data, computational efficiency due to early-stopped PGD. Mainly searching for the least adversarial data among confidently misclassified data
    
    Misclassification Aware adveRsarial Training (MART) \cite{Wang2020Improving}, their method focuses on misclassified examples and trains the model to be more robust to them. ``Misclassification-aware Adversarial Training'' (MAT). The method revisits misclassified examples by generating adversarial examples from them and adding them to the training set.

    Unsupervised Adversarial Training (UAT).
    The labeled data for adversarial robustness is always not possible and expensive too. \cite{UAT} that unlabeled data can be used as a competitive alternative to labeled data to train robust adversary models. Authors show that, in a statistical framework, the complexity of the sample to learn a robust adversary model from unlabeled data corresponds to the fully supervised case down to constant factors i.e., unlabeled data can be used to estimate the regularity of the classifier around natural images, which can improve the adversary robustness of the model. This basically also reduces the natural generalization and adversarial generalization gap.
    
    Guided Adversarial Training (GAT) \cite{GAT}.
    \begin{equation}
        \mathcal{L} = -\textbf{h}_{y}(\hat{x}) + \underset{i\neq y}{\text{max}}~\textbf{h}_{i}(\hat{x}) + \lambda \cdot \| \textbf{h}(\hat{x}) - \textbf{h}(x) \|_{2}^{2}
    \end{equation}
    
    Max-Margin AT \cite{Ding2020MMA}.
    Adaptive selection of the correct margin to achieve optimal results. Instead of using a fixed perturbation magnitude, MMA enables adaptive selection of the correct margin for each data point individually. Making it less sensitive to hyperparameter settings than traditional adversarial training methods.
    \begin{equation}
        \underset{\theta}{\text{min}}
        \left\{\sum_{i\in\mathcal{S}_{\theta}^{+}} \text{max}\{0, (d_{\text{max}}-d_{\theta}(x_{i}, y_{i}))\} + \beta \sum_{j\in\mathcal{S}_{\theta}^{-}} \textbf{J}_{\theta}(x_{i}, y_{i})\right\}
    \end{equation}

    \subsubsection{Data-driven Regularization} \hfill

    Data-driven regularization refers to the incorporation of additional terms or techniques into the objective loss of adversarial training to promote robustness against adversarial shift, whereas the gradient-based methods discussed above involves specific algorithms or modifications that utilize gradients to improve the model's robustness. Moreover, models can be achieve more robustness on combining adversarial training for making models robust against adversarial shift with a regularization approach as seen in multiple surveys \cite{adv-robustness-survey-1, adv-robustness-survey-2}, can be viewed as a data-driven or data-centric regularization. The adversarial regularization was first seen in the \cite{goodfellow2014explaining} paper. The paper proposed a single step attack FGSM. The regularization was an additional term in their objective loss. The results were promising and the method had long list of improvement papers throughout. Qin et al., \cite{local-linearization} included a regularizer that penalizes gradient obfuscation while promoting robustness. This was achieved by encouraging the loss to behave linearly close to the training data. They calculated the first-order taylor expansion on the loss $l(x+\delta)$ and calculated the absolute difference between adversarial loss and it's Taylor expansion. Assumption is that if the loss surface is smooth and approximately linear, then Taylor can easily approximated as well. Mainly replacing FGSM regularizer with local linearity regularizer. Their method achieved 5x training speed compaired to regular adversarial training. \citeauthor{madry2018towards} introduced PGD attack and adversarial training to achieve robustness against stronger PGD attacks. For a couple of years the PGD training has been the default choice for robustness against perturbation attacks. However, TRADES \cite{trades} outperforms PGD adversarial training. In order to provide a differentiable upper bound using the theory of classification-calibrated loss, the authors decompose the prediction error for adversarial examples, also known as the robust error, as the sum of the natural (classification) error and boundary error. It is demonstrated that this upper bound is the tightest bound that is consistent over all probability distributions and measurable predictors. Subfigure~\hyperref[fig:adv-overfitting-trades-regularization]{(e, f)} visually proofs that although both decision boundaries achieve zero natural training error, however TRADES approach achieves better robustness. 


    All the empirical defenses discussed in this section are the defenses based on heuristics to make the model empirically robust. Model parameters are updated by minimizing the worst-case loss in the input vicinity to these models. These defenses are promising to some extent, claim higher robustness but are broken by adaptive attacks. These defenses appear robust, however do not provide any theoretical guarantee that they are not ``breakable''. The next section covers different state-of-the-art models, providing a verification certificate for model robustness.

    \begin{figure}
        \centering
        \includegraphics[width=\textwidth]{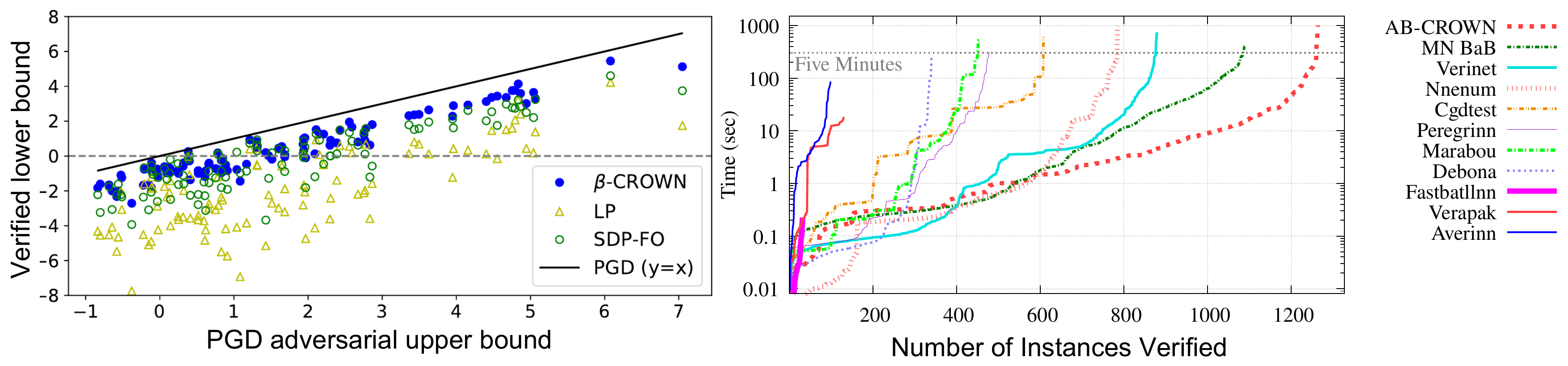}
        \caption{Graph is showing $\alpha,\beta$-CROWN outperforming verifiable neural network methods, taken from formal verification of neural-network competition (VNN-COMP'23) report \cite{Brix2023}.}
        \label{fig:verification-methods}
    \end{figure}

    \subsection{Certified Robustness}
    To guarantee the robustness of DNN models, certified defenses can potentially look for distance or probability certifications. There are numerous defense based models that are published every year, and that claim about being robust against adversarial attacks or adversarial distribution shifts. However, a theoretical guarantee confirms the worst-case test errors and the risks that a deep learning model holds while the it is deployed.

    \begin{wrapfigure}{l}{0.325\textwidth}
        \centering
        \includegraphics[width=0.32\textwidth]{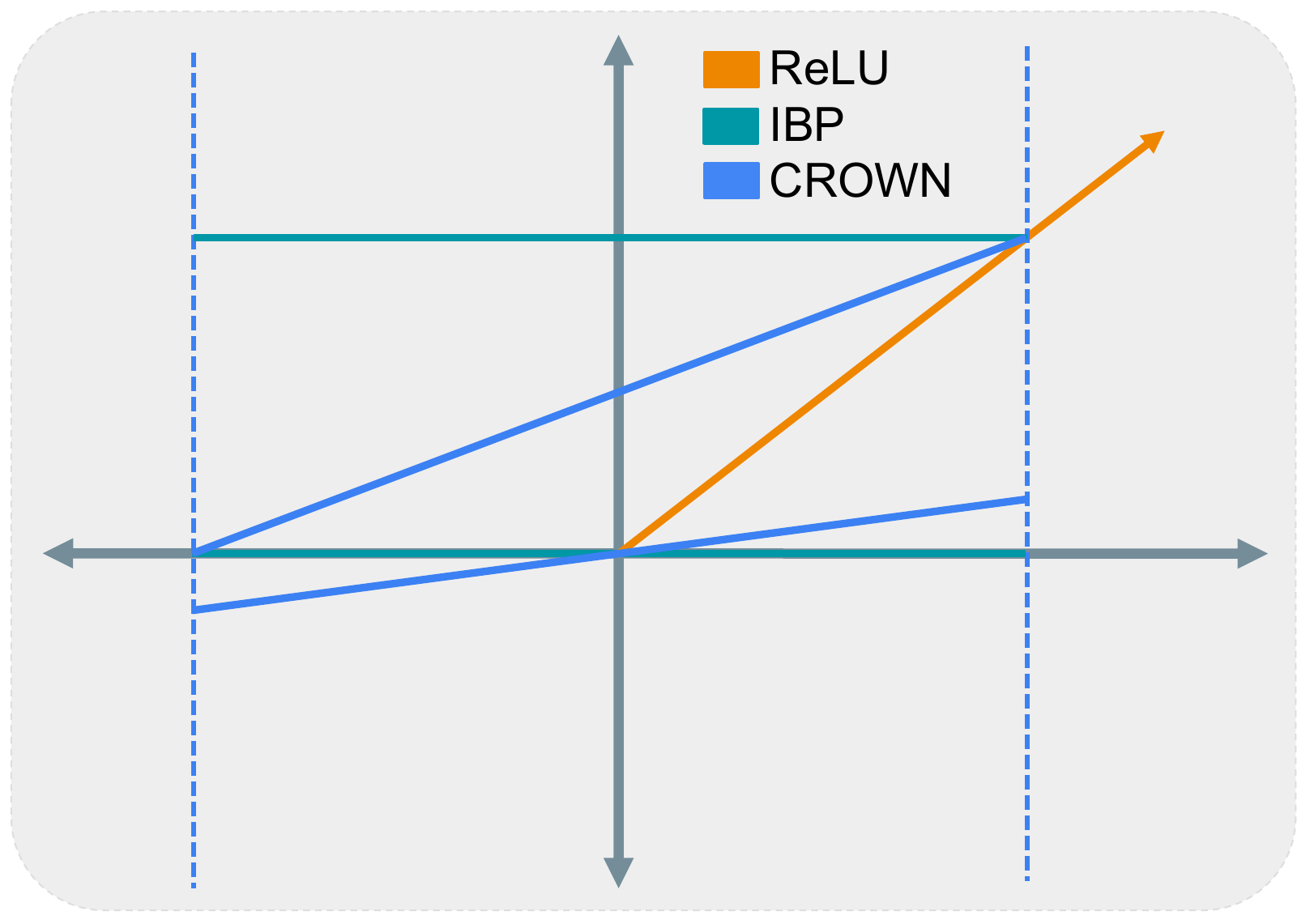}
        \caption{\label{fig:cr}\textit{Upper bound and lower bound non-linear units \cite{lirpa}.}}
    \end{wrapfigure}

    A joint confirmation through a training model and a verification model is necessary to label certified robustness for a model. For instance, if there is a $\epsilon$ perturbation on an image, we would want to prove that the within a given decision boundary the $\epsilon$ norm ball is also going to be classified homogeneously within it as the center point of that boundary. One way, researchers have achieved certifiability is by Lipschitz Constrained over the neural network model. In short, Lipschitz constrains are a property of a function, such that if the input of that function changes then the output of that function is going to change in some bounded way. In we assume the same classification task as earlier, and add a perturbation $\delta$ such that $||delta||_{p} \leq \epsilon$ and that adding this $\delta$ to the input image $\boldsymbol{h}(x+\delta) = y$ holds for correct target. Then any certifiable robustness framework will try to minimize this lower bound ball.
    
    \subsubsection{Deterministic Approaches} \hfill

    Gowal et al., \cite{gowal_IBP} shows how to achieve state-of-the-art in verified accuracy on MNIST for all perturbation radii and CIFAR-10 on 8/255. Authors only focus on $l_{\inf}$ norm bound and conducted evaluations using MIP solver. Authors propose interval bound propagation (IBP) to obtain the certified robustness against adversarial examples. Using a neural network with linear layers and monotonically increasing activation functions as a starting point, a set of permitted perturbations is propagated to determine upper and lower bounds for each layer. These result in constraints on the network's logits, which are used to check/verify if the network modifies its prediction in response to the permitted perturbations. 
    
    \begin{equation}
        \left.
        \begin{array}{ll}
            \underline{z}_{~ki}(\epsilon) = \text{min}_{\underline{z}_{k-1}(\epsilon)\leq z_{k-1} \leq \overline{z}_{k-1}(\epsilon)} ~e_{i}^{T}\boldsymbol{h}_{k}(z_{k-1})\\
            \overline{z}_{~ki}(\epsilon) = \text{max}_{\underline{z}_{k-1}(\epsilon)\leq z_{k-1} \leq \overline{z}_{k-1}(\epsilon)} ~e_{i}^{T}\boldsymbol{h}_{k}(z_{k-1})
        \end{array}
        \right \}
        \label{eq:ibp}
    \end{equation}
    \begin{equation}
        \begin{aligned}
            \mathcal{L} = \underbrace{\kappa l(z_{K}, y_{true})}_{L_{fit}} + (1-\kappa) \underbrace{l(\hat{k}_K(\epsilon), y_{true})}_{L_{spec}}
            \label{eq:IBP-loss}
        \end{aligned}
    \end{equation}
    
    Equation~\ref{eq:IBP-loss}, formulates the training loss, where $l$ is the cross-entropy, and $\kappa$ is a hyperparameter that governs the relative weight of satisfying the specification versus fitting the data. Authors have shown that the proposed approach results in quite tight bounds on robustness and outperforms competing techniques in terms of verified bounds on adversarial error rates in image classification problems, while also training faster. 
    Moreover, Xu et al., \cite{lirpa} stated the goal that given a bounded input for a NN, can we provably bound its output? Their method is to find linear coefficient matrices $\overline{A}$ and $\underline{A}$ and biases w.r.t $x$:
    \begin{equation}
        \begin{aligned}
            \underline{y}: = \underline{A}x+\underline{b} \leq \overline{A}x+\overline{b}: = \overline{y},\qquad ||x-x_{0}||_{\inf} \leq \epsilon 
        \end{aligned}
    \end{equation}
    
    Linear relaxation requires to find the upper and lower bound of the non-linear units. IBP by Gowal et al., \cite{gowal_IBP} is faster in computing the linear relaxation bounds however the bounds are loose. CROWN by Zhang et al., \cite{zhang2018crown} and DeepPoly by Singh et al., \cite{deeppoly} are much better as can be seen in the Figure~\ref{fig:cr}, they can use any lower bound with slope in [0,1]. However, IBP \cite{gowal_IBP} and CROWN \cite{zhang2018crown} are mostly for sequential neural networks with perturbations only on input (x). Requires manual implementation on new network architectures, especially with irregular structure. In spite of the fact that the \cite{lirpa} focus has been on examining known perturbation and offering gaurantees to it. However, an adversary may not always agree the author's presumption in the real-world scenario. In the recent work by Wang et al., \cite{wang2021betacrown}, they developed a new bound propagation technique $\beta-$CROWN that uses parameters that can fully encode neuron splits via optimizable parameters $\beta$ constructed from either primal or dual space. $\beta-$CROWN can handle the split constraints in branch and bound (BaB), allowing combining bound propagation based method with BaB for very efficient GPU based complete verification. It also allows optimization of bounds to potentially achieve tighter bounds that Linear Programming (LP). Figure~\ref{fig:verification-methods} shows the higher quality of bound using $\beta-$CROWN taking about 1 min average data points compared to SDP-FO taking about 3 hours per datapoint. However, the new SOTA is $\alpha,\beta-$CROWN, it operates using a potent framework for linear bound propagation combined with branch and bound techniques. This tool is optimized for rapid performance on GPUs, enabling it to handle fairly extensive convolutional networks, including those with parameter counts in the millions.

    \begin{table}
    \centering
    \tiny
    \caption{\label{tab:adv-defense-summary}Summary of model-centric and data-centric adversarial defense techniques proposed in the last couple of years against malicious adversarial attacks causing adversarial distribution shift.}
    \begin{threeparttable}
    \resizebox{\columnwidth}{!}{
        \begin{tabular}{c p{0.65in} c c p{2.5in} p{1in} c c c} 
        \toprule
        \rowcolor{gray!15}
        Type & 
        Paper & 
        Year & 
        Approach & 
        Focus & 
        Datasets & 
        Adversary's Knowledge & 
        \multicolumn{2}{c}{Approach Type} \\ \cline{7-9} \rowcolor{gray!15}
        & 
        & 
        & 
        & 
        & 
        & BB   GB   WB 
        & \textcolor{orange}{\textbf{DC}} & \textcolor{purple}{\textbf{MC}} \\ \midrule 

        \multirow{65}{*}{\begin{sideways}Proactive\end{sideways}} &

        TRADES \cite{trades} &
        2019 &
        Certification &
        - &
        MNIST, CIFAR-10 &
        \fcolorbox{black}{black!40!black}{\null} \fcolorbox{black}{white!40!white}{\null} &
        \xmark & \cmark \\ &
        IBP \cite{gowal_IBP} &
        2018 &
        Certification &
        Computes an upper bound on the worst-case loss over all possible adversarial perturbations. &
        MNIST, CIFAR-10, SVHN, ImageNet (downscaled) &
        - &
        - & - \\ &
        CROWN \cite{zhang2018crown} &
        2018 &
        Certification &
        Use a combination of interval analysis and quadratic approximations to compute bounds on the output of a neural network for a given input. &
        - &
        - &
        - & - \\ &
        $\beta$-CROWN \cite{wang2021betacrown} &
        2021 &
        Certification &
        Bound propagation based method that can fully encode per-neuron splits. &
        MNIST, CIFAR-10 &
        - &
        - & - \\ &
        \citeauthor{SchmidtSTTM18} \cite{SchmidtSTTM18} &
        2018 &
        Certification &
        The sample complexity of robust learning can be significantly larger than that of "standard" learning. &
        MNIST, CIFAR-10, SVHN &
        - &
        - & - \\ &
        \citeauthor{10.5555/3524938.3525687} \cite{10.5555/3524938.3525687} &
        2020 &
        Preprocessing &
        Early stopping can match the performance gains of recent algorithmic improvements upon adversarial training. &
        SVHN, CIFAR-10/100, ImageNet &
        - &
        \cmark & \xmark \\ &
        Thermometer Encoding \cite{thermometer} &
        2018 &
        Preprocessing &
        Modify neural network architectures via discretizing the input data into a set of binary values that represent temperature levels. &
        MNIST, CIFAR-10/100, SVHN &
        \fcolorbox{black}{black!40!black}{\null} \fcolorbox{black}{white!40!white}{\null} &
        \cmark & \xmark \\ &
        VectorDefense \cite{kabilan2021vectordefense} &
        2021 &
        Preprocessing &
        Use image vectorization as an input transformation step to map adversarial examples back into the natural manifold  &
        MNIST &
        \fcolorbox{black}{gray!40!gray}{\null} \fcolorbox{black}{white!40!white}{\null} &
        \cmark & \xmark \\ &
        PixelDefend \cite{pixeldefend} &
        2018 &
        Preprocessing, Proximity &
        Purify a maliciously perturbed image by moving it back towards the distribution seen in the training data &
        Fashion MNIST, CIFAR-10 & 
        \fcolorbox{black}{white!40!white}{\null} &
        \cmark & \xmark \\ &
        \citeauthor{MustafaKHSS20} \cite{MustafaKHSS20} &
        2020 &
        Preprocessing &
        Use deep image restoration networks to bring adversarial samples onto the natural image manifold. &
        ILSVRC-DEV \tnote{1}, NIPS-DEV &
        \fcolorbox{black}{black!40!black}{\null} \fcolorbox{black}{white!40!white}{\null} &
        \cmark & \xmark \\ &
        \citeauthor{prakash2018deflecting} \cite{prakash2018deflecting} &
        2018 &
        Preprocessing &
        Corrupt the image by redistributing pixel values. &
        1000 images from ImageNet &
        \fcolorbox{black}{white!40!white}{\null} &
        \cmark & \xmark \\ &
        SAP \cite{DhillonALBKKA18} &  
        2018 &
        Gradient Masking &
        A random subset of activations are pruned and the survivors are scaled up to compensate. &
        CIFAR-10 &
        \fcolorbox{black}{white!40!white}{\null} &
        \xmark & \cmark \\ &
        D3 \cite{moosavi2018D3} & 
        2018 & 
        Preprocessing & 
        Reconstruct input images with the denoised patches after input separation. &
        - &  
        \fcolorbox{black}{black!40!black}{\null} \fcolorbox{black}{gray!40!gray}{\null} \fcolorbox{black}{white!40!white}{\null} &
        \cmark & \xmark \\ &
        RRP \cite{Xie2018} & 
        2018 & 
        Preprocessing & 
        Randomize the input images at inference time via resizing and padding. &
        5000 images from ImageNet &
        \fcolorbox{black}{white!40!white}{\null} &
        \cmark & \xmark \\ &
        RSE \cite{LiuCZH18} & 
        2018 & 
        Preprocessing, Ensemble & 
        Add random noise layers to the neural network and ensembles the prediction over random noises. &
        CIFAR-10, ImageNet &
        \fcolorbox{black}{black!40!black}{\null} \fcolorbox{black}{white!40!white}{\null} &
        \cmark & \cmark \\ &
        \citeauthor{BhagojiCM17} \cite{BhagojiCM17} & 
        2017 & 
        Preprocessing & 
        Apply principla component analysis (PCA) to reduce dimensionality. &  
        MNIST, HAR &  
        \fcolorbox{black}{white!40!white}{\null} &
        \cmark & \xmark \\ &
        \citeauthor{LiL17} \cite{LiL17} & 
        2017 & 
        Preprocessing & 
        Use statistics on outputs from convolutional layers to design a cascade classifier. &  
        ILSVRC-2012 &  
        \fcolorbox{black}{white!40!white}{\null} &
        \cmark & \xmark \\ &
        ReabsNet \cite{chen2017reabsnet} & 
        2017 & 
        Preprocessing & 
        Add guardian network to an existing classification network to detect if an image has been adversarially perturbed. &
        MNIST &
        \fcolorbox{black}{white!40!white}{\null} &
        \cmark & \xmark \\ &
        DeT \cite{li2019det} & 
        2019 & 
        Preprocessing, Ensemble & 
        Reconstruct the input example by denoising it. &  
        MNIST, CIFAF-10 &  
        \fcolorbox{black}{black!40!black}{\null} \fcolorbox{black}{gray!40!gray}{\null} &
        \cmark & \xmark \\ &
        Deep Defense \cite{yan2018deep} & 
        2018 & 
        Gradient Masking & 
        Integrates an adversarial perturbation-based regularizer into the classification objective.  &  
        MNIST, CIFAR-10, ImageNet &  
        \fcolorbox{black}{white!40!white}{\null} &
        \xmark & \cmark?\\ &
        \citeauthor{RegClass_Cao} \cite{RegClass_Cao} &
        2017 &
        Proximity &
        The classifier ensembles information in a hypercube centered at the testing example to predict its label. &
        MNIST, CIFAR-10 &
        \fcolorbox{black}{white!40!white}{\null} &
        - & - \\ &
        Feature Distillation \cite{liu2019feature} &  
        2019 &
        Preprocessing &  
        Employ defensive quantization to filter harmful features and refining benign ones to preserve classification accuracy. &  
        ImageNet &  
        \fcolorbox{black}{black!40!black}{\null} \fcolorbox{black}{gray!40!gray}{\null} \fcolorbox{black}{white!40!white}{\null} &
        \cmark & \xmark \\ &
        BAT \cite{wang2019bilateral} & 
        2019 & 
        Gradient Masking & 
        Perturbs both the image and the label during training. &  
        CIFAR-10, SVHN, ImageNet &  
        \fcolorbox{black}{white!40!white}{\null} &
        \xmark & \cmark? \\ &
        S2SNet \cite{folz2020adversarial} & 
        2020 & 
        Gradient Masking & 
        Induce a shift in the distribution of gradients propagated through them, stripping autoencoders from class-dependent signal.  & 
        ImageNet &
        \fcolorbox{black}{gray!40!gray}{\null} \fcolorbox{black}{white!40!white}{\null} &
        - & - \\ &
        \citeauthor{das2017keeping} \cite{das2017keeping} & 
        2017 &
        Preprocessing &
        JPEG compression selectively blurs images by removing high frequency components, effectively eliminating additive perturbations. &
        CIFAR-10, GTSRB &
        \fcolorbox{black}{white!40!white}{\null} &
        - & - \\ &
        MagNet \cite{Meng2017} & 
        2017 & 
        Proximity, AADF & 
        Employ detector networks to identify adversarial examples and a reformer network to align them with normal instances. &  
        MNIST, CIFAR-10 &  
        \fcolorbox{black}{black!40!black}{\null} \fcolorbox{black}{gray!40!gray}{\null} &
        \xmark & \cmark \\ &
        MultiMagNet \cite{machado_iceis2019} & 
        2019 & 
        Ensemble & 
        Randomly incorporates multiple defense components at runtime to introduce non-deterministic behavior. &  
        MNIST, CIFAR-10 &  
        \fcolorbox{black}{black!40!black}{\null} \fcolorbox{black}{gray!40!gray}{\null} \fcolorbox{black}{white!40!white}{\null} &
        \xmark & \cmark \\ &
        ME-Net \cite{yang2019me} & 
        2019 & 
        Preprocessing & 
        Randomly dropping pixels from the image and then reconstructing the image using matrix estimation. &  
        MNIST, CIFAR-10, SVHN, Tiny-ImageNet &  
        \fcolorbox{black}{black!40!black}{\null} \fcolorbox{black}{white!40!white}{\null} &
        - & - \\ &
        Defense-GAN \cite{SamangoueiKC18} &
        2018 & 
        Preprocessing & 
        Train a generative model on clean images to produce close outputs, filtering out adversarial changes. &
        MNIST, Fashion-MNIST &
        \fcolorbox{black}{black!40!black}{\null} \fcolorbox{black}{white!40!white}{\null} &
        \cmark & \xmark \\ &
        HGD \cite{liao2018HGD} & 
        2018 & 
        Preprocessing & 
        Compare clean and denoised outputs, reducing standard denoiser error amplification. &  
        ImageNet &  
        \fcolorbox{black}{black!40!black}{\null} \fcolorbox{black}{white!40!white}{\null} &
        \cmark & \xmark \\ &
        Fortified Networks \cite{lamb2019fortified} & 
        2019 & 
        Preprocessing & 
        Strengthens hidden layers by detecting off-manifold hidden states and mapping them back to well-performing areas of the data manifold. &  
        MNIST, Fashion-MNIST, CIFAR-10 &  
        \fcolorbox{black}{black!40!black}{\null} \fcolorbox{black}{white!40!white}{\null} &
        \cmark & \xmark \\ &
        \citeauthor{xie2019FeatureDenoisingBlock} \cite{xie2019FeatureDenoisingBlock} & 
        2019 & 
        Preprocessing & 
        Add blocks to the network that denoise the features extracted from the images. &
        ImageNet &  
        \fcolorbox{black}{black!40!black}{\null} \fcolorbox{black}{white!40!white}{\null} &
        \cmark & \xmark \\ &
        DDSA \cite{bakhti2019ddsa} &
        2019 & 
        Preprocessing & 
        Remove adversarial noise from input samples before feeding them to a classifier. &  
        MNIST, CIFAR-10 &  
        \fcolorbox{black}{black!40!black}{\null} \fcolorbox{black}{white!40!white}{\null} &
        \cmark & \xmark \\ &
        ADV-BNN \cite{liu2018advbnn} &
        2018 & 
        Gradient Masking & 
        Incorporate randomness via adding noise to all layers in BNN, and perform adversarial training.&  
        CIFAR-10 &  
        \fcolorbox{black}{black!40!black}{\null} \fcolorbox{black}{white!40!white}{\null} &
        \xmark & \cmark \\ &
        DkNN \cite{papernot2018deepnearest} & 
        2018 & 
        Proximity & 
        Incorprate KNN to provide a measure of confidence for inputs outside the model's training manifold. &  
        MNIST, CIFAR-10, GTSRB &  
        \fcolorbox{black}{white!40!white}{\null} &
        \xmark & \cmark? \\ &
        WSNNS \cite{dubey2019WMM} & 
        2019 & 
        Proximity & 
        Project the adversarial images back onto the image manifold.  &  
        IG-N-*, YFCC-100M, IN-1.3M &  
        \fcolorbox{black}{black!40!black}{\null} \fcolorbox{black}{gray!40!gray}{\null} &
        - & - \\ \midrule
        
        \multirow{35}{*}{\begin{sideways}Reactive\end{sideways}} &

        \citeauthor{FeinmanCSG17} \cite{FeinmanCSG17} & 
        2017 & 
        Statistics & 
        Utilize Bayesian uncertainty estimates and density estimation in deep features for implicit adversarial detectio. &  
        MNIST, CIFAR-10 &
        \fcolorbox{black}{white!40!white}{\null} &
        \xmark & \cmark \\ &
        \citeauthor{carrara2018adversarial} \cite{carrara2018adversarial} & 
        20xx & 
        Proximity & 
        Hidden layer activations can be used to reveal incorrect classifications caused by adversarial attacks. &  
        CIFAR-10 &  
        \fcolorbox{black}{white!40!white}{\null} &
        - & - \\ &
        \citeauthor{zheng2018robust} \cite{zheng2018robust} &  
        20xx &  
        Statistics, Proximity &  
        Use the output distributions of the hidden neurons in a DNN classifier presented with natural images to detect adversarial inputs. &  
        MNIST, F-MNIST &
        \fcolorbox{black}{black!40!black}{\null} \fcolorbox{black}{gray!40!gray}{\null} &
        \xmark & \cmark? \\ &
        \citeauthor{GrosseMP0M17} \cite{GrosseMP0M17} &  
        2017 &  
        Statistics &  
        Statistical tests identify non-conforming inputs, while an additional output in the ML model classifies adversarial inputs. &  
        MNIST, DREBIN, MicroRNA &  
        \fcolorbox{black}{black!40!black}{\null} &
        - & - \\ &
        RCE \cite{PangDZ17} &  
        2017 &  
        Gradient Masking &  
        Minimize the reverse cross-entropy during training to encourage the deep network to better distinguish adversarial examples from normal ones. &  
        MNIST, CIFAR-10 &
        \fcolorbox{black}{black!40!black}{\null} \fcolorbox{black}{white!40!white}{\null} &
        - & - \\ &
        NIC \cite{ma2019nic} &  
        2019 &  
        ADM, Proximity &  
        Analyze exploitation channels and extracting invariants for runtime detection under various attacks. &  
        MNIST, CIFAR-10, ImageNet, LFW &  
        \fcolorbox{black}{black!40!black}{\null} \fcolorbox{black}{white!40!white}{\null} &
        - & - \\ &
        \citeauthor{Hendrycks2017} \cite{Hendrycks2017} &
        2017 &
        Preprocessing &
        Detect adversarial images by comparing inputs and their reconstructions using an auxiliary decoder. &
        Tiny-ImageNet, CIFAR-10, MNIST &
        \fcolorbox{black}{white!40!white}{\null} &
        \cmark & \xmark \\ &
        LID \cite{Ma0WEWSSHB18} &  
        2018 &  
        Proximity &  
        LID measures space-filling capability around a reference example, using neighbor distances to characterize adversarial region dimensional properties. &  
        MNIST, CIFAR-10, SVHN &  
        \fcolorbox{black}{white!40!white}{\null} &
        - & - \\ &
        \citeauthor{CohenSG20} \cite{CohenSG20} &  
        2020 &
        Proximity &  
        Influence functions measure training sample impact on validation data, identifying the most supportive samples for each validation example. &  
        CIFAR-10/100, SVHN &  
        \fcolorbox{black}{white!40!white}{\null} &
        - & - \\ &
        \citeauthor{madry2018towards} \cite{madry2018towards} & 
        2018 & 
        Gradient Masking & 
        The notion of security against a first-order adversary as a natural and broad security guarantee. & 
        MNIST, CIFAR-10 &
        \fcolorbox{black}{black!40!black}{\null} \fcolorbox{black}{white!40!white}{\null} &
        \cmark & \xmark \\ &
        MALADE \cite{counterstrike} &
        2018 & 
        Preprocessing & 
        Adversarial samples are relaxed onto the target class manifold using the Metropolis-adjusted Langevin algorithm, considering perceptual boundaries. & 
        MNIST, CIFAR-10, Tiny-ImageNet &  
        \fcolorbox{black}{black!40!black}{\null} \fcolorbox{black}{white!40!white}{\null} &
        \cmark & \xmark \\ &
        Gong et al. \cite{GongWK17} & 
        2017 & 
        - & 
        The binary classifier is trained to identify specific features that are present in adversarial samples but not in clean data. &  
        MNIST, CIFAR10, SVHN &  
        \fcolorbox{black}{white!40!white}{\null} &
        - & - \\ &
        SafetyNet \cite{lu2017safetynet} & 
        2017 & 
        AADF & 
        Comprise an original classifier and adversary detector, analyzes later layers to identify and reject adversarial examples. &  
        CIFAR-1O, ImageNet-1000, SceneProof &  
        \fcolorbox{black}{black!40!black}{\null} \fcolorbox{black}{white!40!white}{\null} &
        - & - \\ &
        \citeauthor{NingHao2018ModelInterpretation} \cite{NingHao2018ModelInterpretation} &
        2018 & 
        Gradient Masking & 
        Leverage ML model interpretation to understand prediction mechanisms, enabling deeper insights for crafting adversarial samples. &  
        Twitter, YelpReview &  
        \fcolorbox{black}{white!40!white}{\null} &
        \xmark & \cmark \\ &
        Memory Defense \cite{adhikarla2022memory} & 
        2022 & 
        AADF & 
        Restrict inter-manifold learning of different classes by masking other classes during the reconstruction of an image.  &  
        Fashion-MNIST, CIFAR-10 &
        \fcolorbox{black}{black!40!black}{\null} \fcolorbox{black}{gray!40!gray}{\null} &
        \xmark & \cmark \\ 
        \\ \bottomrule

    \end{tabular}}
    \begin{tablenotes} 
        \item[1] 5000 images from the ImageNet dataset.
    \end{tablenotes}
    \end{threeparttable}
        \footnotesize
            \begin{enumerate}
                \item The symbols [\cmark]~ and [\xmark]~ are used to indicate the category of the method being described. With [\cmark]~ representing a common practice in the research community, and [\xmark]~ indicating that the method is not applicable due to technical restrictions.
                \item BB $\rightarrow$ Black-box, GB $\rightarrow$ Gray-box, WB $\rightarrow$ White-box.
                \item DC $\rightarrow$ Data-Centric, MC $\rightarrow$ Model-Centric.
            \end{enumerate}
    \end{table}

    \subsubsection{Probabilistic Approaches}\hfill

    Probabilistic Approaches are based on Randomized Smoothing. Apart from applying relaxations to the non-linearity in the neural networks, one can apply randomized smoothing in cases where robustness issues arises due to lack of smoothness. Salman et al., \cite{smooth-adv-training}, Cohen et al., \cite{randomized-smooth}, Mikl{\'o}s Z et al., \cite{boosting-randomized-smooth} are some popular probabilistic approaches that are recently in use. As also discussed in a GitHub repository\footnote{\url{https://github.com/eashanadhikarla/data-centric-ai/blob/main/robustness.md\#probabilistic-approaches}}, probabilistic approaches are more flexible than deterministic methods because they do not require knowledge of detailed neural network structure. Indeed, probabilistic approaches have been extended to defend against various threat models or improved to be deterministic. According to Silva et al., \cite{adv-robustness-survey-1}, randomized smoothing is based on differential privacy and investigates the relationship between DP and robustness against norm-bounded adversarial examples. Assuming a classifier function $\textbf{\textit{h}} : R^{d} \rightarrow [0,1]^{k}$ mapping inputs $x_{i}\in X$ with logits $[0,1]$ can be said to be $\epsilon-$radii robust at $x$ if:
    \begin{equation}
        \begin{aligned}
            H(x+\delta) = c(x), \forall: ||\delta|| \leq \epsilon
        \end{aligned}
    \end{equation}
    In addition, when \textbf{\textit{h}} is a L-Lipschitz constraint, then \textbf{\textit{h}} classifier is $\epsilon$-radii robust at $x$ with:
    \begin{equation}
        \begin{aligned}
            \epsilon = \frac{1}{2L}(p_{a}-p_{b})\\
            \text{where}, ~p_{a} = max_{i\in[k]}f_{i}(x)
        \end{aligned}
    \end{equation}
    Cohen et al., in \cite{randomized-smooth} proposed a smoothed classifier for generating tight certification bounds. Their second theorem defines the certification radius (r) as:
    \begin{equation}
        \begin{aligned}
            r = \frac{\sigma}{2}(\phi^{-1}(\underline{p_{a}}) - \phi^{-1}(\overline{p_{b}}))
        \label{eq:radium-randomized-smooth}
        \end{aligned}
    \end{equation}
    Figure~\ref{eq:radium-randomized-smooth} shows the decision boundaries of the different classes with different colors and the certification radius as defined in the equation \ref{eq:radium-randomized-smooth}. As the $sigma$ noise goes higher, the certification radius increases and so the probability of the class A. 

    Table~\ref{tab:adv-defense-summary} summarizes the adversarial defense techniques. Defensive techniques can be broadly classified into two categories, namely \textit{proactive} and \textit{reactive} defenses, based on their primary objectives. \textit{Proactive} defenses are geared towards enhancing the robustness of classification models against adversarial distribution shifts by reducing their vulnerability to such shifts. This can be achieved by either training the model on a dataset of adversarial examples generated under various distributions or by utilizing adversarial training techniques. Conversely, \textit{reactive} defenses concentrate on detecting adversarial distribution shifts after they have been created and before they are utilized to target a classification model, thereby preventing any potential damage. AADF (Auxiliary Adversarial Defense Filter), is a defensive technique that uses adversarial training to create a secondary binary model. This model acts as a filter by checking whether an input image is authentic or adversarial before passing it to the primary application classifier.

\section{Unseen Distribution Shift}
\label{sec:US}
    \begin{figure}[!ht]
            \centering
            \includegraphics[scale=0.50]{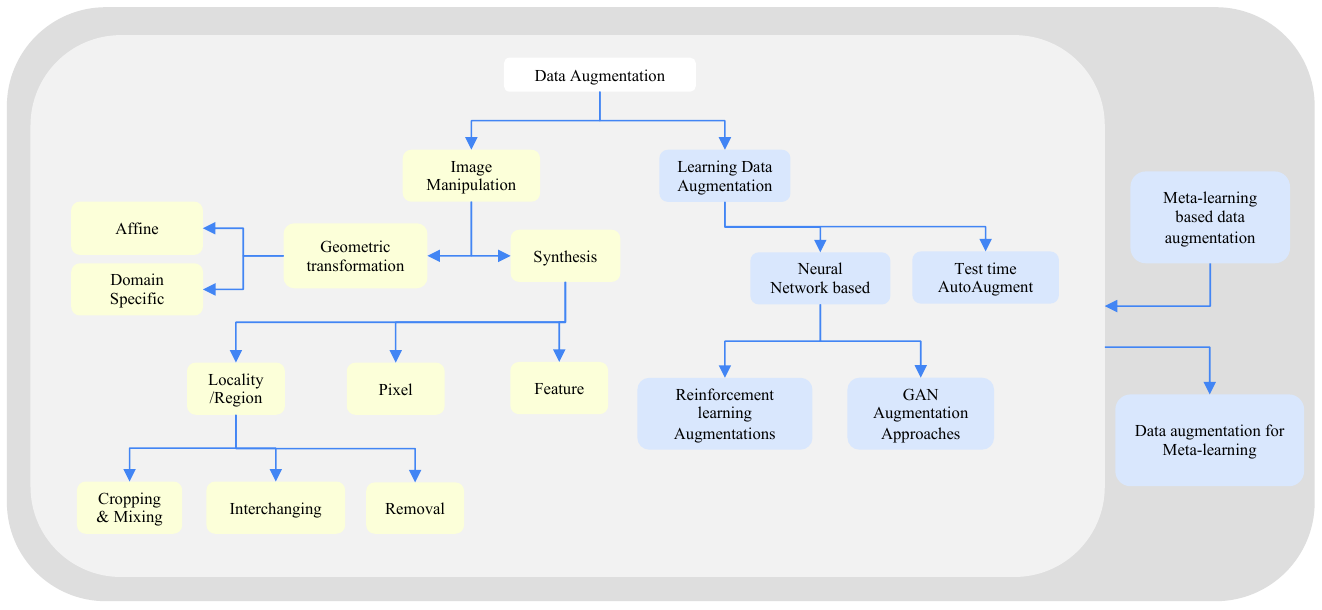}
            \caption{\label{fig:unseen-shift}\textit{Outlining the recent literature categorization on data augmentation techniques for unseen distribution shifts.}}
    \end{figure}
    Data augmentation is a powerful technique for improving the robustness of machine learning models to distribution shifts. By artificially creating new data points from existing data around the distribution cloud, data augmentation can help models to learn to generalize to unseen data distributions. This is especially important for applications that contain outdoor scenarios, which are often subject to unusual conditions such as image corruptions, snowfalls, and object renditions. For such cases, it is often infeasible to collect enough real-world data to train a robust model. Larger data will certainly lead to better performances. However, collecting and labeling large amounts of data can be time-consuming and expensive, and in some cases, there may be limited data available for a particular task. Data augmentation can help to fill this gap by providing models with a wider range of data to learn from. \citeauthor{make5010019} explains the significance of large quantity of data has been fulfilled by addressing challenges in lack of training data, lack of relevant data, model overfitting, imbalanced data. A lot of them have been studied through the lens of GAN based augmentation techniques (more details in Section~\ref{gan-aug})
    
    In addition to being effective, data augmentation is also a relatively simple and inexpensive technique to implement. This makes it a valuable tool for any machine learning practitioner who is looking to improve the robustness of their models to distribution shifts. In Fig~\ref{fig:unseen-shift}, we categorize data augmentations through \textit{Image Manipulation} and \textit{Learning Data augmentation}.

    \subsection{Image Manipulation}
    The process of altering an image's appearance is widely known as image manipulation. This can involve various reasons, such as enhancing image quality, editing images, and in the current era of machine learning, it is heavily used for data augmentation.

    \subsubsection{Geometric Transformation}
    A transformation method that can change the location of pixels without altering their value/intensity is referred to as a geometric transformation on an image. The primary objective of applying geometric transformation on image datasets for machine learning is to make the model invariant to position and orientation changes, which helps it learn more generally for specific domains\footnote{\url{https://www.sciencedirect.com/topics/computer-science/geometric-transformation}}. Moreover, \citeauthor{unseen-shift-survey-1} \cite{unseen-shift-survey-1} explains that a non-label preserving transformation could result in low-confidence predictions. Constructing refined labels during post-augmentation for achieving robustness is computationally expensive. Therefore, it is crucial to evaluate a data augmentation technique's potential impact on the data and the resulting machine learning model's accuracy before applying it to a given dataset. This can include testing the technique on a subset of the data, comparing the results with the original dataset, and analyzing the potential risks and benefits of using the technique.
    \begin{definition}\label{def:geometric-transformation}
        A geometric transformation is defined by a pair of function $x' = \phi(x,y)$, ~$y' = \psi(x,y)$ that map the old coordinates $(x,y)$ into the new ones $(x',y')$.
    \end{definition}
    The process of geometric data augmentation, also known as pixel-based or image-based data augmentation, involves applying a range of spatial transformations to the input data, including flipping, rotating, scaling, or cropping an image. These transformations are relatively straightforward and do not cause significant alterations to the image's content.

    \subsubsection{Synthesis}
    In contrast to geometric transformation, synthesis data augmentation involves the creation of new artificial samples by synthesizing them from existing data through various techniques such as image blending, texture transfer, or generative models. Synthesis data augmentation enables the production of new images that can have more intricate and varied content, including images that merge features from multiple input images or images that feature new objects or textures that were not present in the original data. One of the advantages of synthesis data augmentation is that it can generate more realistic and diverse data that better represents the distribution of the underlying data. This can help machine learning models to generalize better to new, unseen data, which can significantly enhance their performance.

    \begin{figure}[!ht]
        \centering
        \includegraphics[scale=0.35]{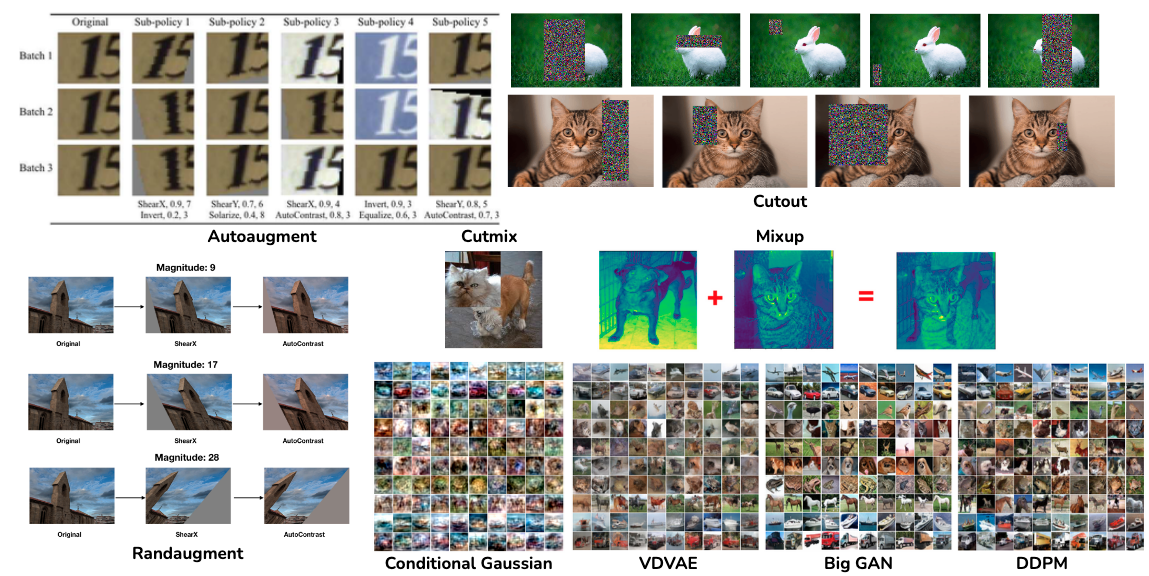}
        \caption{\label{fig:UDS}\textit{Illustrating different kinds of augmentation techniques found in the literature.}}
    \end{figure}

    \paragraph{Locality/Region Level}\hfill

        \textbf{Cropping \& Mixing}.\quad In the context of cropping and mixing cropped patches of images, several data augmentation methods have been developed that aim to improve the performance and robustness of image classification models. One such method is SmoothMix \cite{smoothmix}, which combines the strengths of existing techniques called mixup and cutmix. SmoothMix blends the features of two images in a smooth manner, which helps to preserve spatial information rather than simply replacing them, which helps to preserve the spatial information in the images and  hence improve the generalization of the model. SaliencyMix \cite{uddin2021saliencymix} by \citeauthor{uddin2021saliencymix}, which uses saliency maps to create more informative and challenging training examples. By combining two images with their corresponding saliency maps, SaliencyMix masks out less important areas of each image, resulting in more difficult but more informative training examples. SuperMix \cite{supermix} is another such augmentation technique that combines MixUp, CutMix, and CutOut in a novel way. This method has been shown to be effective in improving the performance and robustness of image classification models. Finally, Attentive CutMix \cite{attentive-cutmix}, a method that uses attention to improve the performance of CutMix. It is done by masking out irrelevant parts of images before they are cut and pasted, Attentive CutMix reduces the introduction of artifacts into the images, resulting in more discriminative and informative training examples. Overall, these data augmentation methods provide a promising approach to improving the performance and robustness of image classification models.

        \textbf{Interchanging}.
        \quad CutMix \cite{yun2019cutmix} is a powerful data augmentation technique that uses two images to improve the performance of deep neural networks. This method blends two images, replacing a portion of one with a portion of the other, to increase the network's robustness to small changes in input data. The algorithm randomly selects a rectangle from each image, masks out the selected rectangles, and blends them together. This technique has been shown to be effective in improving the performance of deep neural networks in a variety of tasks, including image classification, object detection, and segmentation. CutBlur \cite{cutblur} is a more advanced version of CutMix that combines cutmix and blur. It first performs cutmix on two images, blending them together, and then blurs the resulting image. This technique improves the network's robustness to noise and occlusions. CutBlur has been shown to be effective in improving the performance of deep neural networks on a variety of tasks. ResizeMix \cite{resizemix} is another data augmentation technique that combines resize and mixup. It first resizes two images to the same size, and then performs mixup on the resized images, blending their features together. This technique improves the network's robustness to different input sizes. GridMix \cite{GridMix} is a data augmentation technique that combines grid and mixup. It divides each image into a grid of cells and performs mixup on the cells of the two images. This technique improves the network's robustness to different object positions in the input image. GridMix has been shown to be effective in improving the performance of deep neural networks on a variety of tasks.

        \textbf{Removal}.
        \quad Several effective data augmentation techniques have been proposed for improving the generalization performance of convolutional neural networks (CNNs) on various tasks. These techniques include a CutOut \cite{devries2017cutout}, which involves removing contiguous sections of input images, effectively augmenting the dataset with partially occluded versions of existing samples, and Random-Erasing \cite{random-erasing}, which randomly erases a rectangular patch from an image with a probability $p$ during training. Hide-and-seek \cite{singh2018hide}, a weakly supervised localization approach that hides a target object in an image with a goal to effectively predict category label and bounding-box, while GridCut and Mix \cite{gridcut-mix} cuts out a grid of patches from an image and mixes them with patches from other images. GridMask \cite{chen2020gridmask} randomly masks out a grid of patches from an image with different sizes and aspect ratios. Initially, the algorithm detects image regions that exhibit uniform distribution and subsequently arranges them into a grid. Consequently, the pixel distribution within the image serves as a foundation for information omission. AttributeMix \cite{AttributeMix} mixes the attributes of two images, while SuperMix \cite{supermix} mixes the features of two images. Finally, the FenceMask \cite{li2020fencemask} randomly masks out a fence-shaped patch from an image, and the KeepAugment \cite{gong2021keepaugment} randomly masks out a rectangular patch from an image while ensuring that the masked-out patch does not contain any vital information about the object in the image. These data augmentation techniques have been shown to be effective at improving the performance of CNNs on various tasks and are more efficient than other data augmentation techniques, such as cutout.

    \paragraph{Pixel Level}
        Pixel-level augmentations can help with the problem of unseen shifts by applying random transformations to the pixels of the images, such as noise, blurring kernels, and patch crops. This helps to ensure that the model is exposed to a wider range of input data. Mixup-based \cite{zhang2017mixup} strategies have proven more effective, as they combine two data points by linearly interpolating their features, resulting in a virtual feature-target vector. 
        ROIMix \cite{roimix} extends the mixup technique to object detection tasks by mixing proposals from multiple images, making the model more adaptable to variations in object appearance. The authors of PuzzleMix \cite{pmlr-v119-kim20b} and Co-Mixup \cite{kim2021comixup} have suggested incorporating aspects of both pixel-based and region-based techniques to enhance adversarial resilience and improve the generalization accuracy. Co-Mixup \& Puzzle Mix first computes saliency maps for each data point. Saliency maps highlight the most critical regions of an image, which can be used to guide the mixing process. Co-Mixup then mixes two data points by first creating a joint saliency map. The joint saliency map is then used to mix the two images such that the most salient regions of each image are preserved. Whereas Puzzle Mix then mixes two data points by creating a puzzle of the two images. The puzzle is then solved by placing the most salient regions of each image together. RegMix \cite{regmix} is a new data augmentation technique designed explicitly for regression tasks. RegMix first computes a regression model for each data point. The regression models are then used to mix the two data points such that the output of the mixed data point is a weighted average of the outputs of the two original data points. Finally, AdaMixUp \cite{adamix} is an auxiliary network that recognizes MixUp as a form of ``out-of-manifold regularization,'' imposing ``local linearity'' constraints on the model's input space beyond the data manifold. This analysis highlights the limitation of MixUp, referred to as ``manifold intrusion.'' Machine learning models can benefit from mixup approaches, which are widely used techniques to enhance their generalization performance. Nevertheless, mixup can also lead to the inclusion of OOD samples, which are not part of the original dataset and can negatively affect the model's ability to learn accurate correlations and perform well on new data. 

    \paragraph{Feature Level}
        One important aspect to consider is the diversity of extracted feature maps, even when collecting a large amount of data. This can become problematic for tasks like image classification, where the model must be able to generalize to unseen data. Recent research has shown that expanding training data by expanding extracted feature maps is a promising solution. Several works, such as PointMixup \cite{point-mixup} and Manifold Mixup \cite{manifold-mixup} both work by interpolating between two different data points, while PatchUp \cite{patchUp} and Feature Space Augmentation \cite{feature-space-aug} work by randomly modifying the input data. Smart Augmentation \cite{smartAug} is a more general approach that can be used to learn an optimal data augmentation strategy for any deep learning model. PointMixup and Manifold Mixup are simple and effective methods that can be easily implemented, but they may not be as effective for tasks where the data is noisy or has a lot of outliers. PatchUp and Feature Space Augmentation are more robust to noise and outliers, but they can be more difficult to implement and may not be as effective for tasks where the data is very sparse. Smart Augmentation is the most general approach, but it can be more computationally expensive to train the reinforcement learning agent. Recently, transformer \cite{ViT} based methods such as Token Mixup \cite{token-mixup} method for transformer-based models that works by first detecting salient tokens in the input data, and then interpolating between the hidden states of those tokens. 
 
    However, synthesis data augmentation can be more computationally expensive and requires more sophisticated techniques than geometric data augmentation. It also requires careful tuning to ensure that the synthesized data is realistic and representative of the original data.

    \subsection{Learning Data Augmentation}\hfill

    Incorporating potential invariances into the data is more reliable than hard-coding the variance into the data. However, improving network design has been a major area of focus for the machine learning and computer vision communities. Finding improved data augmentation techniques that incorporate more invariances has received less focus. In the literature, we can see widely used methods to add additional knowledge to the dataset while training machine learning vision models. While data augmentation is widely used to enhance training datasets, it's true that developing effective and scalable techniques that capture relevant invariances is challenging. Designing application-specific data augmentation methods can yield good results but may not generalize well to different domains or tasks. Scaling data augmentation techniques to new applications requires careful consideration of the specific problem at hand. Some invariances may be more universally applicable, such as geometric transformations like rotations or translations. Still, others might be more specific to certain domains or tasks, such as medical imaging or satellite imagery analysis. 

    \subsubsection{Neural Network based Learning Data Augmentation}
        \paragraph{Reinforcement Learning Augmentation}\hfill

            \textbf{AutoAugment}\hfill

                \begin{figure}
                    \centering
                    \includegraphics[scale=0.25]{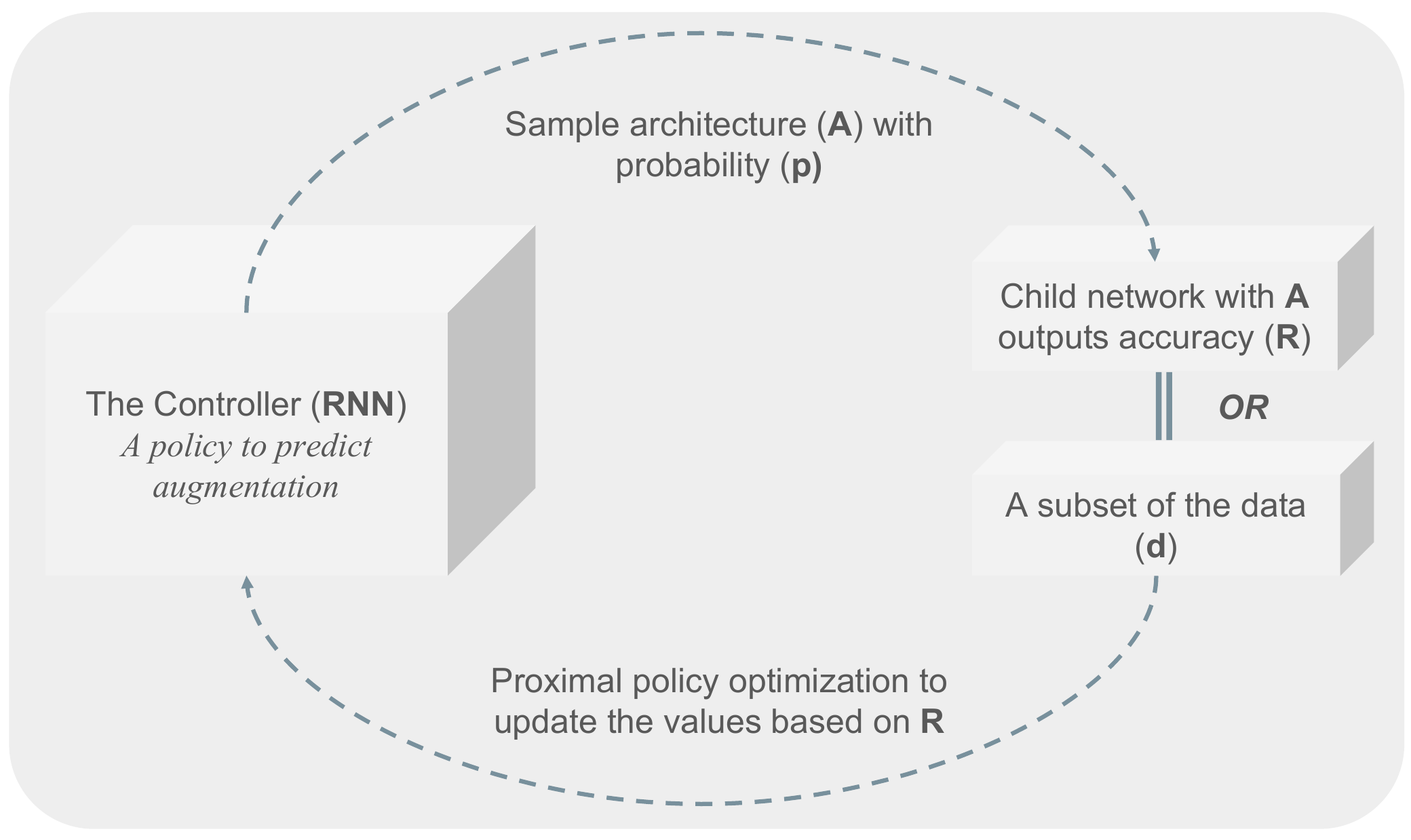}
                    \caption{\label{fig:autoaugment-principle}Core principle behind training a RNN for Autoaugment and other reinforcement learning based augmentations.}
                \end{figure}
                Cubuk et al., \cite{cubuk2019autoaugment} proposed AutoAugment approach for data augmentation using reinforcement learning search algorithms to search through a possible set of data augmentations on images. Compared to other methods such as \cite{zhang2017mixup,devries2017cutout,yun2019cutmix} where they apply the augmentation in a user-specified manner with a fixed hyperparameter for those augmentations. In contrast, the authors in AutoAugment approach with the fundamental principle of searching augmentations and define a discreet search space as shown in Figure~\ref{fig:autoaugment-principle}. In total they implement 16 simple data augmentation techniques each with 10 magnitudes (for ex. if there is color brightness, then there can be 10 magnitudes for strength of brightness), 11 Values for probability of applying operations, 5 sub-policies (each containing two operations). Combining all these into one exhaustive search space, there can be $(16x10x11)^10 \sim 2.9 x 10^{32}$ combinations of augmentations. Figure~\ref{fig:UDS} shows that in a given sub-policy there are two operations each associated with two numerical values, and for every image in a mini-batch, they choose a sub-policy uniformly at random to generate a transformed image to train the neural network. Authors show in Table~3 \cite{cubuk2019autoaugment}, autoaugment makes a improvement of ~6\% improvement from baseline ResNet-50 with top-1 and top-5 accuracies of 83.5\%/96.5\%. This improvement is remarkable given that the best augmentation policy was discovered on 5,000 images. However, another extended work of Geng et al., \cite{ARS} shows that autoaugment is constraint with limited performance due to discreet search space with operation type and magnitude. On the other hand, Geng et al., provided a continuous search space instead of discreet to fix the inherent flaw in Autoaugment.

            \textbf{RandAugment}\hfill

                Cubuk et al., \cite{cubuk2020randaugment} claims that randaugment is much more practical then previous automated data augmentation techniques like \cite{cubuk2019autoaugment} or Population Based Augmentation \cite{PBS}. Unlike autoaugment and Nueral Architecture Search uses RNN as shown in Figure~\ref{fig:autoaugment-principle} that does a seperate search for data augmenation, randaugment has drifted away from this idea and rather dramatically reducing the search space for automated data augmentation. As authors target Autoaugment \cite{cubuk2019autoaugment} as their baseline, they share numerous critics about using Autoaugment, most importantly about the sub-policy taking two operations each with two magnitude params. Notably, using a small proxy task isn't a good indicator of performance when we scale up the autoaugment method to a larger network and larger dataset. On a low level comparison with \cite{cubuk2019autoaugment}, autoaugment tries to find the probability of the operation among 16 different operations, the maginitude for each of the operations, and trying to find five sub-policies to optimize the search space. However, randaugment on the other side, only uses two arguments. `N': number of augmentation transformations to apply sequentially, `M': Magnitude for all the transformations. In the Figure~\ref{fig:UDS}, Randaugment uses N:2 and varying M vertically with [9,17,28] as magnitude of the transformation. Figure~1 of Randaugment \cite{cubuk2020randaugment}; bottom two graphs show an increasing trend that means if the model complexity increases or if the training dataset size increases, the child network requires higher magnitude in the search space. This is clearly infeasible to achieve with Autoaugment. Randaugment experiments compared Autoaugment \cite{cubuk2019autoaugment}, Fast AA \cite{faa}, PBA \cite{PBS} as their baselines on four different datasets: CFAIR-10, CIFAR-100, SVHN (core-set), SVHN. It outperformed the baselines with Shake-Shake, PyramidNet, WideResnet-28-10. Although the performance is not too significant compared with Autoaugment, but it significantly a lot faster and easily scalable.

        \textbf{Deep AutoAugment}\hfill

            \begin{figure}[!ht]
                \centering
                \includegraphics[scale=0.25]{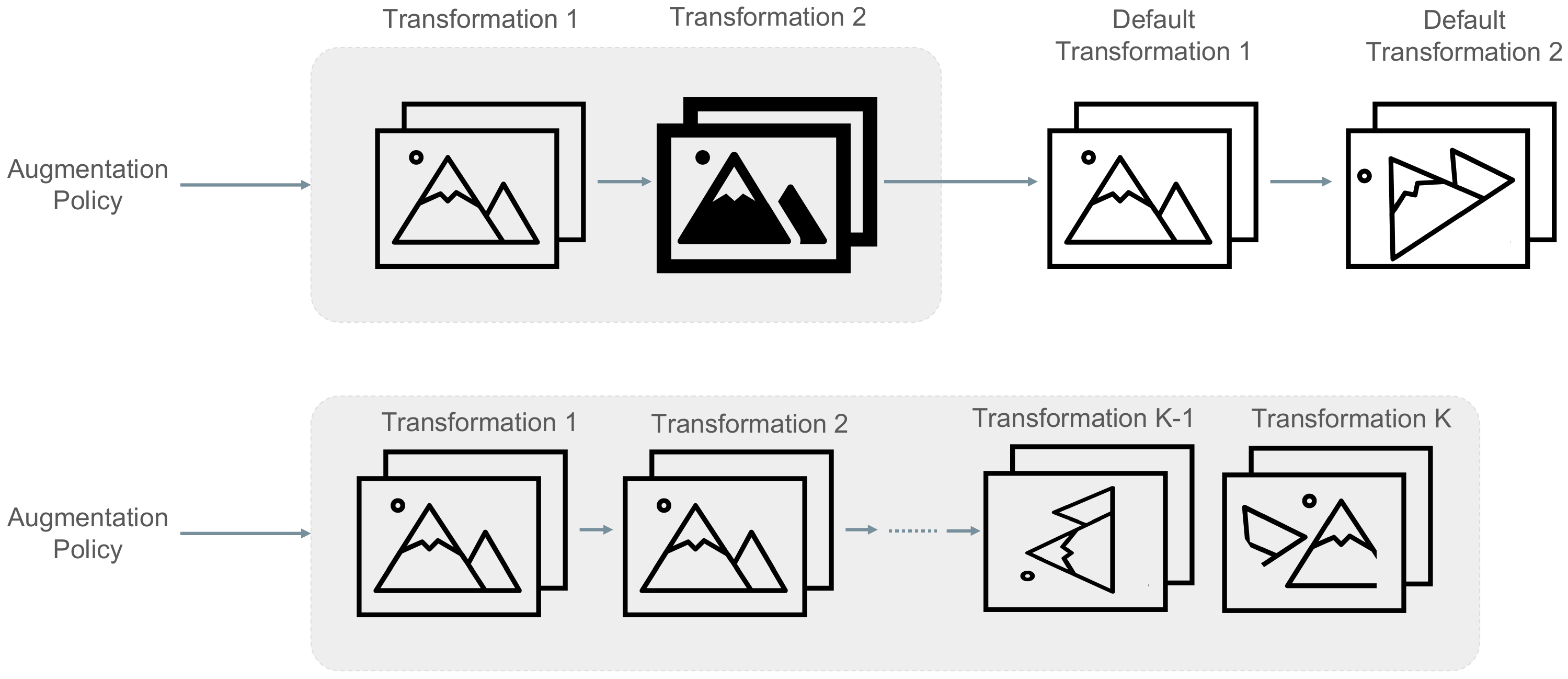}
                \caption{\label{fig:deepauto}The figure contrasts the augmentation pipeline structure of prior methods versus the proposed DeepAA.~[\textbf{Top}]~\textit{Approaches use shallow policies with only 1-2 learned transformations, followed by fixed default augmentations.}, [\textbf{Bottom}]~\textit{DeepAA eliminates any handpicked defaults and instead learns a deeper pipeline with multiple augmentation layers optimized from scratch.}}
            \end{figure}
            Deep AutoAugment (DeepAA) \cite{zheng2022deep}, a fully automated method for searching data augmentation policies without using any handpicked default augmentations (as shown in Figure~\ref{fig:deepauto}). The key idea is to formulate augmentation search as a regularized gradient matching problem that maximizes the cosine similarity between gradients of augmented and original data along low variance directions. DeepAA builds a deep augmentation pipeline by progressively stacking layers and optimizing each conditioned on previous layers, avoiding exponential growth in search space.

        \paragraph{Generative Models for Data Augmentation}\hfill

        \label{gan-aug}
            \begin{figure}[!ht]
                \centering
                \includegraphics[width=0.75\textwidth]{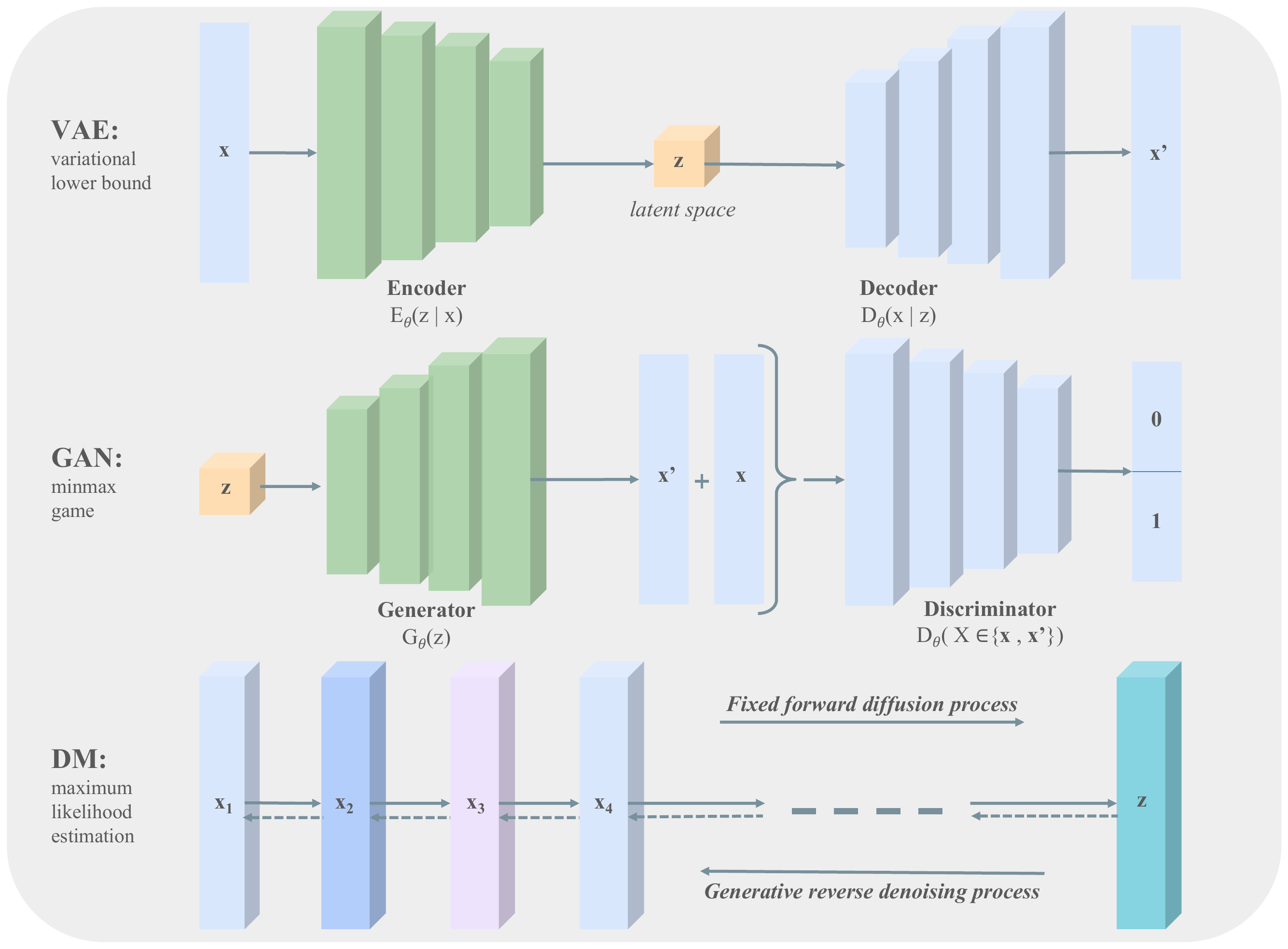}
                \caption{\label{fig:gm}\textit{Framework comparison of Variational Autoencoders (VAE), Generative Adversarial Networks (GAN), and Diffusion Models (DM) for generative modeling. The diagram illustrates the architectural components and training objectives of each model, highlighting their unique approaches to generating synthetic data.}}
            \end{figure}

            %
            \textbf{GAN based Augmentation}

            Previously, we have seen in literature such as \cite{make5010019}, how GAN based augmentation technique that are comprised of a generator $G(z)$ whose task is to generate synthetic data samples (images, texts, audio, etc) from a latent distribution $z \in Z$ have been studied and became widely popular. Traditional GAN Methods such as AC-GAN \cite{ac-gan}, where AC-GAN, adds an auxiliary classifier to the discriminator network to stabilize the training of discriminator, DA-GAN \cite{da-gan}, where generate augmented versions of training images. The augmented images are then used to train a classifier, and the classifier is evaluated on the original training images. The results show that DA-GAN can improve the performance of classifiers on unseen data, T-GAN \cite{t-gan} where, the generator is trained to generate images from one domain, the discriminator is trained to distinguish between real and fake images, and the translator is trained to translate images from one domain to another, and in-turn generating more realistic examples. Ensemble methods have been incorporated into recently introduced GAN techniques \cite{ensemble-gan} to enhance the quality of synthetic data. The authors emphasized the significance of utilizing multiple GANs to increase the variety of generated data. Although a single GAN can produce seemingly diverse image content, training on such data often results in significant overfitting. To examine the effects of ensembled GANs on synthetic 2D data, the authors conducted tests and concluded that combining independently trained GANs through ensembling could offer advantages over boosting strategies, yielding comparable or superior performance while being less susceptible to overfitting. Aforementioned techniques for data augmentation is limited in its ability to generate new data and can be difficult to train. GANs for data augmentation can be unstable and the generated synthetic samples may not be representative of real-world \cite{make5010019,10.1145/3548785.3548793}. Figure~\ref{fig:gm} shows the fundamental differences in the framework pipelines for VAEs, GANs and DM based architectures.

            \textbf{Diffusion Model based Augmentation}
            \begin{figure}[!ht]
                \centering
                \includegraphics[width=\textwidth]{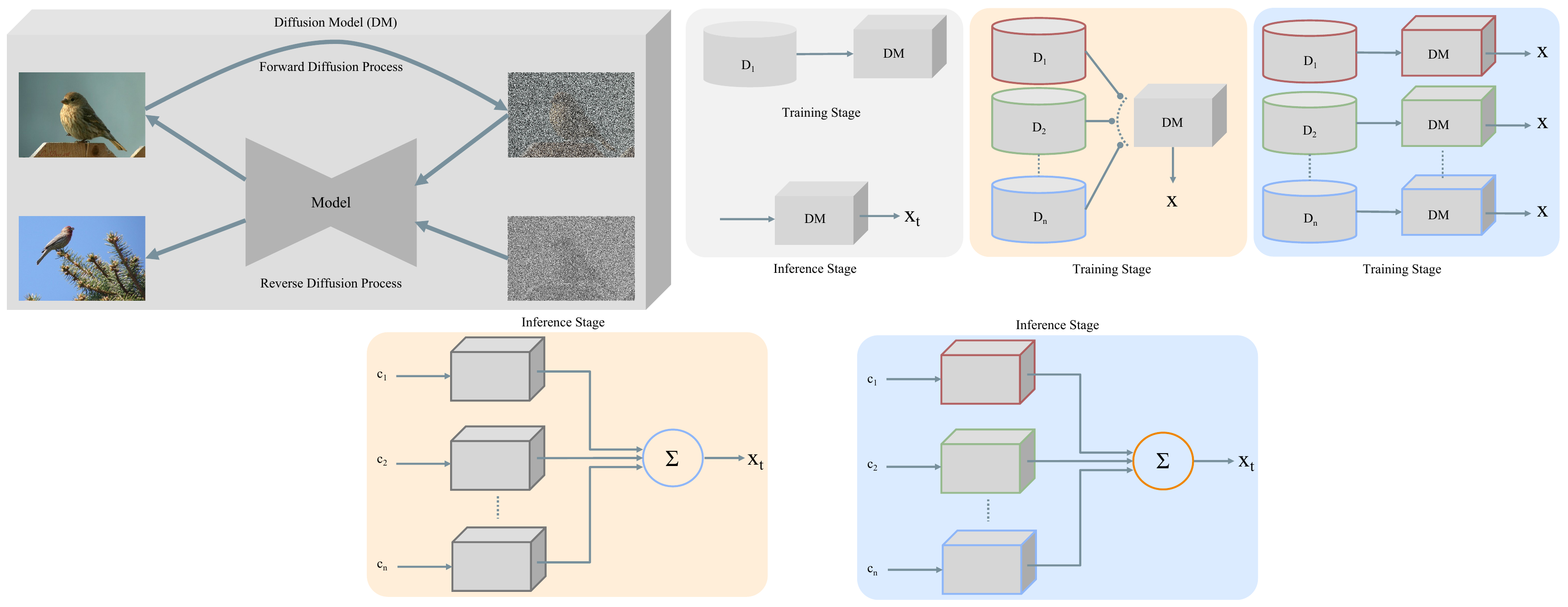}
                \caption{\label{fig:diffusion-model}\textit{A visual comparison representing different styles of multi-modal diffusion model frameworks.}}
            \end{figure}
            \begin{figure}[!ht]
                \centering
                \includegraphics[width=\textwidth]{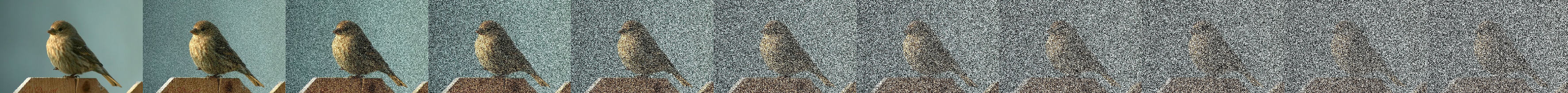}
                \caption{\label{fig:diffusion-model-noise}\textit{Visualization of the diffusion model process, where noise is gradually added to the latent space over time steps to generate new images.}}
            \end{figure}

            Traditional synthetic data is often generated by randomly sampling from a distribution of real images, but this can lead to data that is unrealistic and not representative of the real world. On the other hand, Diffusion models are a type of generative model that can be used to generate images by gradually adding noise to a latent space shown in Figure~\ref{fig:diffusion-model-noise}. This allows the models to generate images that are more realistic and diverse than traditional synthetic data. 
            Among the recent models diffusion models towards generating synthetic datasets, \citeauthor{DBLP:journals/corr/abs-2304-08466} shows that their method can generate more realistic and diverse data than traditional synthetic data and has improved the performance of ImageNet classification models by up to 1.5\%. One limitation of the method \cite{DBLP:journals/corr/abs-2304-08466} is that it can be computationally expensive to generate large amounts of synthetic data. However, the authors believe that the benefits of using synthetic data outweigh the costs. \citeauthor{DBLP:journals/corr/abs-2301-13188} show that diffusion models memorize individual images from their training data and emit them at generation time. They use a ``\textit{generate-and-filter}'' pipeline to extract over a thousand training examples from state-of-the-art models, ranging from photographs of individual people to trademarked company logos. They also train hundreds of diffusion models in various settings to analyze how different modeling and data decisions affect privacy. In addition to their findings on memorizing training data with high fidelity, they raise concerns as a fundamental limitation of diffusion models, and that it is not possible to completely eliminate the risk of privacy leakage and that it is possible for an attacker to extract sensitive information from a diffusion model \cite{DBLP:journals/corr/abs-2301-13188}. The authors' method, called Unite and Conquer \cite{nair2023unite}, is based on diffusion models, which are a type of generative model that can be used to generate images by gradually adding noise to a latent space. Authors' \citeauthor{nair2023unite} key insight is that diffusion models can be used to generate images that satisfy multiple constraints by combining the predictions of multiple diffusion models. Figure~\ref{fig:diffusion-model} illustrates method \cite{Xia_2021_CVPR}, that require paired data training across all modalities, while \cite{nair2023unite} approach allows separate training for different modalities or utilizing separate diffusion models. During sampling, independent processing of conditioning strategies and combination of outputs preserve diverse conditions.\newline

    In general, data augmentation is more effective at mitigating small distribution shifts than large distribution shifts. This is because small distribution shifts can be more easily accommodated by the CNN's, GAN's, etc. The amount of data augmentation used also affects its effectiveness. More data augmentation is generally better at mitigating distribution shifts, but it can also make the training process more difficult and time-consuming. Finally, the architecture of the CNN also affects its ability to generalize to unseen distribution shifts. Model architectures with more capacity are generally more robust to distribution shifts, but they can also be more difficult to train. Overall, data augmentation is an important technique for improving the robustness of CNNs to unseen distribution shifts. However, it is important to note that data augmentation is not a silver bullet and it cannot completely eliminate the effects of unseen distribution shifts.

    \subsubsection{Test time Augmentation (TTA)}\hfill

    Data augmentation has been abundantly studied for training the model and has been effective. In case of TTA, we augment the test data for the model that is ready for deployment. Usually, a deployed model encounters each data instance (in this instance: an image) just once, meaning there's only a solitary chance for the model to make an accurate prediction. Through TTA (Test-Time Augmentation), the deployed model receives numerous augmented variations of each instance of test data, resulting in multiple probabilities. This allows the model an increased likelihood of correctly predicting the output. Moreover, it can help to improve the robustness and accuracy of the model's predictions. However, as we can see it comes with a inference computational cost that might become more challenging for the time-sensitive applications. We formulate a general TTA equation to an image-classification task:
    \begin{equation}
        \hat{y} = \mathbb{E}(f(T(x_1)), f(T(x_2)), f(T(x_3)), \dots, f(T(x_n)) )
    \end{equation}
    Here, $\hat{y}$ is th epredicted label for the test image $x$. $f$ is the deep learning model, $T$ is a transformation/augmentation function that applies a data augmentation or a set of data augmentations to the test image $x$, and lastly, $n$ is the number of data augmentations used. The ability of TTA to address unseen distribution shift relies on several factors, such as the nature of the distribution shift, the model's complexity, and the quantity and nature of data augmentation transformations employed. In general, TTA proves more effective in mitigating minor distribution shifts compared to significant ones.
    
    Numerous studies such as Radosavovic et al. \cite{radosavovic2018data}, Wang et al. \cite{wang2018testtime} have shown the augmentation of test time in their choice of design and its benefits. In \cite{radosavovic2018data} paper, authors evaluate the effectiveness of test-time augmentation on the COCO dataset. They show that test-time augmentation can improve the performance of the model by up to 1.1\% AP. The authors believe that test-time augmentation helps to improve the performance of the model by two main mechanisms. First, it helps to improve the robustness of the model's predictions. Second, it helps to learn new knowledge from the extra unlabeled data. However, TTA has not only helped classification of unseen data, a kind of unseen data is also known as anomaly. \citeauthor{COHEN2023821} show how unsupervised diverse test-time enhancement helps in boosting anomaly detection. The authors used a technique called test-time augmentation to apply a variety of transformations to the test images, such as flipping, rotation, and scaling. They then used a technique called clustering to create a diverse set of synthetic anomalies. A recent method TeSLA \cite{Tomar_2023_CVPR} uses a technique called self-learning in test time to generate adversarial examples, and uses these examples to train a model that is more robust to adversarial attacks. Due to the modular nature of TTA, it has also recently been tested for domain adaption, MEMO \cite{DBLP:conf/nips/ZhangLF22}. The method is again proposed to improve robustness against unseen data, where the authors propose a combination of test time enhancement and domain adaptation with applying a variety of transformations to the test images, such as flipping, rotation, and scaling. In segmentation tasks for Melanoma cancer authors in \cite{melanoma-tta} used TTA transformations with conditional random fields to model the spatial dependencies between pixels in the images. Although test time enhancement has shown some promising results such as in the case of YOLOv5x-TTA \cite{yolov5} having a mean average precision of mAP 55.8 to YOLOv5x with mAP 50.7 as reported in \cite{yolov5}.

    Hence, it is worth noting that TTA can only help to mitigate unseen distribution shift that is caused by factors that are known to the model. For example, if the model is trained on images of buildings and houses, TTA can help to improve the model's ability to classify images of buildings and houses that are taken from different angles or with different lighting conditions. TTA can only help to improve the model's accuracy to a certain extent. If the distribution shift is too large, TTA may not be able to prevent the model from making errors. TTA can be computationally expensive, especially if a large number of data augmentation transformations are used.


    \begin{figure}[H]
        \centering
        \includegraphics[scale=0.30, angle=90]{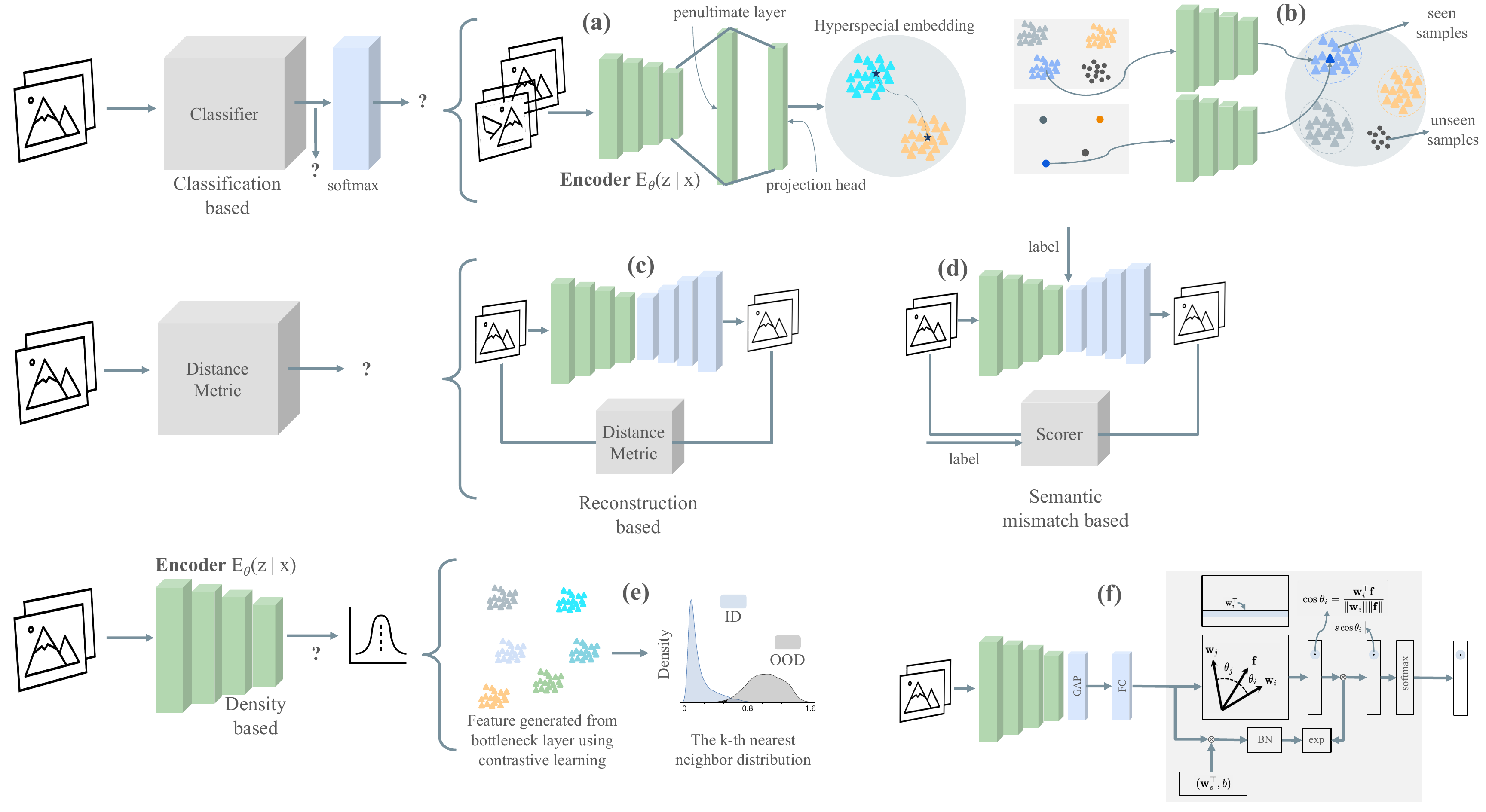}
        \caption{\label{fig:ood-detection}\textit{Comprehensive Framework for Out-of-Distribution (OOD) Detection: This figure illustrates the multi-faceted approaches to OOD detection in machine learning models, addressing unseen distribution shifts. \textbf{[Top]} classifier based, \textbf{[Middle]} distance based, \textbf{[Bottom]} density based.}}
    \end{figure}

    \subsection{Out of Distribution (OODs)}\hfill

        Out of Distribution shift is an important aspect of the unseen distribution shift, where the data distribution differs from the training data distribution $\mathcal{D}_{\text{train}}$. OOD shift can lead to incorrect or dangerous decisions, as deep learning models often exhibit overconfidence in their predictions on OOD data \cite{hendrycks2017a}. As \citeauthor{yang2021oodsurvey} describe in \cite{yang2021oodsurvey}, In OOD detection, the test samples are sourced from a distribution that displays a semantic shift relative to the in-distribution (ID). This shift is manifested through a discrepancy in the probability distribution of the labels, denoted as $P(Y) \neq P'(Y)$. It is crucial to highlight that the training dataset typically encompasses multiple classes, and the objective of OOD detection is to identify out-of-distribution samples while preserving the model's capability to accurately classify in-distribution samples. We categorize OOD detection and mitigation into three distinct approaches as shown in Figure~\ref{fig:ood-detection}. and are described below.

        \begin{definition} \cite{JMLR:v13:gretton12a} \label{ood-def}
            Let $(X, Y)$ be a random variable valued in a space $X \times Y$ with unknown probability density function (pdf) $p_{XY}$ and probability distribution $P_{XY}$. Here, $X \subseteq \mathbb{R}^d$ represents the covariate space and $Y = \{1, \dots, C\}$ corresponds to the labels attached to elements from $X$. The training dataset $S_N = \{(x_i, y_i)\}_{i=1}^N$ is defined as independent and identically distributed (i.i.d) realizations of $P_{XY}$.
        \end{definition}

    \subsubsection{Approaches based on Distance metrics}\hfill

        In the context of OOD detection and mitigation, distance metrics can be employed to measure the dissimilarity between input data and the training distribution. These approaches aim to identify and handle OOD samples based on their distance from the training data. The methods use feature embeddings to calculate the distance between test samples and in-distribution (ID) data. Methods assume that OOD samples will be relatively far away from the ID data, and they flag these samples as OOD.

        Traditional methods utilize distance metrics, such as Maximum Mean Discrepancy (MMD) \cite{JMLR:v13:gretton12a} or Kullback-Leibler (KL) divergence \cite{Kullback51klDivergence}, to detect OOD samples. These metrics quantify the dissimilarity between the training and testing distributions. By setting a threshold on the distance measure, samples that exceed the threshold are flagged as OOD. Below, we divide recent articles into parametric and non-parametric methods:

        \paragraph{Parametric Methods}\hfill
        
        Parametric methods for out-of-distribution (OOD) detection assume that the in-distribution data follows a known distribution. These methods typically calculate the distance between a test sample and the mean of the in-distribution data using a covariance matrix that is estimated from the in-distribution data. The distance score is then used to classify the test sample as in-distribution or OOD.

        The following proposed methods \cite{LeeLLS18, abs-2106-09022, AmersfoortSTG20, HuangLWCZD21} aim to detect out-of-distribution (OOD) samples, and three of them also aim to detect adversarial attacks. Authors in \cite{LeeLLS18} uses a Mahalanobis distance based approach to calculate an OOD score for each test sample, which is then used to classify the sample as in-distribution or OOD. The novelty they achieved was that their model can detect both OOD and adversarial examples, however, it is too sensitive to hyper-parameters and computationally expensive. On the other hand, \citeauthor{abs-2106-09022} proposes a simple fix to the Mahalanobis distance method \cite{abs-2106-09022} by adding a regularization term to improve its performance on near-OOD samples. \citeauthor{AmersfoortSTG20} used a single deep deterministic neural network to estimate both the class label and the uncertainty of the prediction based on Bayesian interpretation of the uncertainty estimates, which can be used to detect OOD samples. 
        \begin{equation}
            M(x) = \max_{y_i \in \mathcal{Y}} ~ - \left( (x - \mu_{y_i})^T \Sigma_y^{-1} (x - \hat{\mu}_{y_i}) \right)
        \end{equation}
        where, $M(x)$ is the \textbf{Mahalanobis distance-based confidence score}, $y$ is the class label, $\mu_{y_i}$ is the mean of the in-distribution data for class $y_i$ and $\Sigma_y^{-1}$ is the covariance matrix of the in-distribution data for class $y_i$. 
        Feature Space Singularity \cite{HuangLWCZD21} detects OOD samples by identifying feature space singularities, which are regions of the feature space where the density of the in-distribution data is very low. \cite{AmersfoortSTG20} explains `Why Sensitivity can be at odds with Classification?' with a toy experiment involving two features $x_1$ and $x_2$ from Gaussian distribution, and output $y$ is determined by $y = sign(x_1) \cdot \epsilon$, where $\epsilon$ represents noise with occasional label flips. While the optimal decision function for classification prioritizes empirical risk, it fails to account for OOD behavior. The challenge emerges when encountering inputs like $x_1, x_2 = (1,10000)$, significantly deviating from observed data.

        One key difference between the methods \cite{LeeLLS18,abs-2106-09022,AmersfoortSTG20,HuangLWCZD21} is the approach used to detect OOD samples. The two methods \cite{LeeLLS18,abs-2106-09022} use a Mahalanobis distance based approach, while the method \cite{AmersfoortSTG20} uses a deep deterministic neural network, and the method \cite{HuangLWCZD21} uses feature space singularities. 

        \paragraph{Non-parametric Methods}\hfill

        Non-parametric strategies utilized for OOD detection do not presume any inherent structure within the distribution of in-distribution or OOD samples. This unique attribute makes these methods more resilient to alterations in the distribution of OOD samples. Opting for non-parametric approaches provides a significant advantage as they demand no prior knowledge of the underlying distribution and are often computationally more efficient than their parametric counterparts.

        \begin{definition}
            From the formulation in definition~\ref{ood-def}, detecting out-of-distribution (OOD) samples boils down to building a binary rule $g : X \to \{0, 1\}$ through a soft scoring function $s : X \to \mathbb{R}$ and a threshold $\gamma \in \mathbb{R}$. Namely, a new observation $x \in X$ is then considered as in-distribution, i.e., generated by $P_{XY}$, when $g(x) = 0$ and as OOD when $g(x) = 1$. Finding this rule $g$ from $X$ can become intractable when the dimension $d$ is large. Thus, previous work rely on a multi-layer pre-trained classifier $f_\theta : X \to Y$ defined as:
            $$f_\theta(x) = h \circ f_L \circ f_{L-1} \circ \dots \circ f_1(x)$$
            where, $h$ is a final layer and $f_i$ are non-linear functions.
        \end{definition}

        Several non-parametric studies have introduced innovative techniques for enhancing OOD detection by leveraging concepts such as autoencoders, Wasserstein distance, cross-entropy, cosine similarity, and more. A noteworthy study by \citeauthor{DBLP:journals/corr/abs-1812-02765} recommends incorporating autoencoder reconstruction error ($L_{recon}$) with Mahalanobis distance ($L_{maha}$) within the latent space for improved detection of specific types of OOD samples \cite{DBLP:journals/corr/abs-1812-02765}. The study posits that some OOD data may closely align with the latent manifold of in-distribution data but remain distant from in-distribution samples within the latent space. In order to address this situation, they suggest the application of Mahalanobis distance in tandem with reconstruction error, represented as $L = L_{recon} + \alpha L_{maha}$, where $\alpha$ serves as a weighting factor for the Mahalanobis distance term.

        Meanwhile, \citeauthor{DBLP:conf/cvpr/Zhou22}, in their research, examine autoencoder-based OOD detection and identify two preconditions for valid OOD metrics, which pertain to the latent feature domain and decoder reconstructive power \cite{DBLP:conf/cvpr/Zhou22}. Their proposition includes maximizing the compression of the latent space with cross-entropy regularization ($L_reg$) and reconstructing activations rather than images. The OOD score is calculated using the normalized $L_{2}$ reconstruction error (Equation~\ref{eq:zhou-ae-ood}), converted to a probability with a function $\psi$.

        \citeauthor{DBLP:conf/eccv/ChenLSZ20} present a boundary-based OOD classifier for generalized zero-shot learning, wherein class-specific manifolds are learned \cite{DBLP:conf/eccv/ChenLSZ20}. The classifier creates a bounded manifold for each observed class on a unit hyper-sphere by utilizing von Mises-Fisher distributions within a shared latent space (shown in Figure~[\ref{fig:ood-detection}.b]). Samples that fall outside all in-distribution class manifolds are categorized as OOD.

        \citeauthor{DBLP:journals/corr/abs-1905-10628} advocate the usage of softmax of scaled cosine similarity in the last layer of a classifier for OOD detection, deploying a data-dependent scale factor \cite{DBLP:journals/corr/abs-1905-10628}. The key motivation driving this approach is the recognition that hyperparameters, when tuned on assumed OOD samples, do not generalize efficiently to the actual OOD samples encountered in practical scenarios. The predicted class probabilities are obtained by applying the softmax of the scale factor, as indicated in Equation~\ref{eq:scale-factor}.
        \begin{equation}\label{eq:scale-factor}
            \mathcal{L} = -log \frac{e^{w_c^{\top}f + b_c}}{\sum_{i=1}^{C}e^{w_c^{\top}f + b_i}}
        \end{equation}
        Building on these methods, \citeauthor{DBLP:conf/iclr/MingSD023} introduced CIDER, a contrastive learning framework furnished with dispersion ($L_disp$) and compactness ($L_comp$) losses, intended to shape hyperspherical embeddings for enhanced OOD detection \cite{DBLP:conf/iclr/MingSD023} (shown in Equation~\ref{eq:zhou-ae-ood}). These losses promote inter-class dispersion and intra-class compactness, contributing to superior OOD detection performance compared to other methods such as SSD+ and KNN+. CIDER's prowess is demonstrated in experiments on CIFAR benchmarks, where it established new state-of-the-art results by explicitly optimizing the embedding space for ID-OOD separability.
        \begin{equation}\label{eq:zhou-ae-ood}
            P(x \in \mathcal{X}) = P(E(x) \in S_{ID})~\cdot\psi~(NL2(x, D(E(x))))
        \end{equation}
        \citeauthor{DBLP:journals/corr/abs-2306-02879} demonstrate that individual neurons can display different activation patterns in response to in-distribution (ID) versus out-of-distribution (OOD) inputs. This finding highlights the potential of analyzing neuron behavior as a way to characterize the model's state with respect to the OOD problem. 
        %
        Overall, the resilience of non-parametric methods to outliers and their independence from any assumptions about the in-distribution data distribution contribute to their wide use. However, it is noteworthy that in some instances, such as when the in-distribution data follows a known distribution and is well-behaved, parametric methods may prove more effective.

    \subsubsection{Approaches based on Density metrics}\hfill

    Density based metrics are a class of techniques for detecting out-of-distribution (OOD) or anomalous samples that rely on modeling the density of the data. The key idea is to train a probabilistic generative model to estimate the density $p(x)$ of the training data. Then, at test time, samples with low density under the learned model are considered likely to be OOD or anomalous.
    The density $p(x)$ could be modeled with likelihood-based generative models like PixelCNN or normalizing flows. Since these models can tractably compute densities, the density value itself or the negative log-likelihood $-log p(x)$ can be used as an anomaly score. However, density values alone are often not enough, as highlighted by papers \cite{choi2019generative,KobyzevPB21,SerraAGSNL20}. Various techniques have been proposed to adjust, improve or contextually interpret the densities, such as using ensembles, input complexity, optimizing latent parameters, or combining with compression models. The overall goal is to obtain a score that assigns low values to in-distribution samples and high values to OOD samples.

    \citeauthor{choi2019generative} in \cite{choi2019generative} shows that modeling density alone is insufficient, as likelihoods exhibit high variance across models. It proposes using ensembles of generative models to estimate the `Watanabe-Akaike Information Criterion' (WAIC) which combines density and model variance:
    \begin{equation}
        WAIC(x) ~=~ \mathbb{E}_{\theta}[log p_{\theta}(x)] - \text{Var}_{\theta}[log p_{\theta}(x)]
    \end{equation}
    The variance penalty reduces the score for OOD samples that are sensitive to model parameters. Experiments show WAIC significantly improves over likelihoods alone for anomaly detection. While  \cite{choi2019generative} focuses on variance across models, \cite{KobyzevPB21} provides a review of normalizing flows, a class of generative models that can tractably model complex densities:
    \begin{equation}
    \textit{log}~p(x) ~=~ \textit{log}~p(f(x)) + \textit{log}~\abs{det \frac{\partial f}{\partial x}}~
    \end{equation}
    It summarizes different types of flows and their performance on density modeling tasks. However, good densities alone do not guarantee good OOD detection.

    Some more recent studies have proposed innovative methods to enhance model robustness against unseen distribution shifts. Notably, \cite{SerraAGSNL20}, \cite{XiaoYA20}, and \cite{ZisselmanT20} present distinct approaches, each contributing uniquely to this field.
    
    \cite{SerraAGSNL20} highlights the influence of input complexity on likelihood-based generative models. The authors argue that conventional likelihood measures in generative models are significantly biased by the complexity of the input data. They propose an OOD score that accounts for this bias:
    \begin{equation}
        S(x) = -\ell_M(x) - L(x)
    \end{equation}
    where \(-\ell_M(x)\) represents the negative log-likelihood and \(L(x)\) is the complexity estimate. This approach suggests that by adjusting for input complexity, one can achieve a more reliable OOD detection, crucial for handling unseen data distributions.
    
    In contrast, \citeauthor{XiaoYA20} introduces the concept of Likelihood Regret (LR) for OOD detection in VAEs. LR is defined as the difference in log likelihood between the optimized and original model configurations for a given sample:
    \begin{equation}
        LR = \ell_{OPT}(x) - \ell_{VAE}(x)
    \end{equation}
    This metric effectively captures the discrepancy in model behavior for in-distribution and OOD samples, offering a nuanced view of model robustness. The LR framework is particularly insightful for understanding how VAEs can be misled by OOD samples, emphasizing the need for models that can adaptively discern between in-distribution and OOD data.

    \citeauthor{ZisselmanT20} takes a different approach by introducing a residual flow architecture for OOD detection. This method leverages the concept of normalizing flows, extending the Gaussian distribution model to capture more complex data distributions. The residual flow model is formulated as:
    \begin{equation}
        x = g(z) + r(z)
    \end{equation}
    where \(g(z)\) represents the base Gaussian distribution and \(r(z)\) is the learned residual component. This architecture demonstrates superior performance in detecting OOD samples, particularly in image datasets, by effectively modeling the intricate distributions of neural network activations.
    
    Comparatively, while \citeauthor{SerraAGSNL20} and \citeauthor{XiaoYA20} focus on modifying the likelihood-based metrics to enhance OOD detection, \citeauthor{ZisselmanT20} innovates at the architectural level. The residual flow model's ability to capture complex distributions offers a significant advantage in handling diverse and unforeseen data shifts. This diversity in approaches underscores the multifaceted nature of OOD detection and the importance of considering both data complexity and model architecture for robust machine learning systems.

    Despite the significant advancements made in \cite{SerraAGSNL20}, \cite{XiaoYA20}, and \cite{ZisselmanT20} in OOD detection, there remain inherent limitations when considering robustness against unseen data distributions. While the adjustment for input complexity \cite{SerraAGSNL20} and the introduction of Likelihood Regret \cite{XiaoYA20} offer improved detection metrics, these methods primarily rely on the assumption that deviations from learned distributions are indicative of OOD samples. This assumption may not always hold, especially in complex real-world scenarios where data distributions can overlap or exhibit subtle variations.

    Furthermore, the residual flow architecture proposed by \citeauthor{ZisselmanT20}, despite its effectiveness in capturing complex distributions, may face challenges in scalability and computational efficiency, particularly when applied to high-dimensional data. Moreover, the current approaches predominantly focus on individual model improvements without extensively addressing the collective behavior of models in an ensemble setting. Ensemble methods could potentially offer a more robust solution against diverse and dynamic data distributions.

    \subsubsection{Approaches based on output reconstruction}\hfill
    
    Reconstruction based methods technically fall under the density based methods, however there has been a wide range of study recently done in this category. The approaches have a underlined assumption that the ID can be well-reconstructed from trained generative model but OOD cannot as they are unseen during training and perhaps do not fall inside the cluster of learned representation \cite{denouden2018improving, zhou2022rethinking, yang2022out}. 
    
    \citeauthor{denouden2018improving} enhances the traditional autoencoder approach by integrating Mahalanobis distance in the latent space with the reconstruction error. This hybrid method addresses the issue where some OOD samples lie near the latent dimension manifold but are far from any known encoded training sample. The novelty score in this approach is given by:
    \begin{equation}
        \text{novelty}(x) = \alpha \cdot DM(E(x)) + \beta \cdot \ell(x, D(E(x)))
    \end{equation}
    where \(DM\) represents the Mahalanobis distance, \(E\) and \(D\) are the encoding and decoding functions, and \(\alpha, \beta\) are mixing parameters.
    
    \citeauthor{zhou2022rethinking} proposes a different strategy, focusing on maximizing the compression of the autoencoder's latent space while maintaining its reconstructive power. This approach involves semantic reconstruction and normalized L2 distance, aiming to address the limitations of traditional reconstruction-based methods.
    
    \citeauthor{yang2022out} introduces MoodCat, which uses a generative model to synthesize masked images based on classification results. The key innovation here is the calculation of semantic differences between the original and synthesized images, providing a novel perspective in OOD detection.
    
    Comparatively, while \citeauthor{denouden2018improving} and \citeauthor{zhou2022rethinking} focus on enhancing the traditional autoencoder framework, \citeauthor{yang2022out} adopts a more radical approach by synthesizing images to highlight semantic discrepancies. This diversity in methodologies underscores the complexity of OOD detection and the need for versatile strategies to handle unseen data distributions.
    
    In terms of robustness against unseen distribution shifts, these methods offer promising directions. However, they also come with limitations. The effectiveness of \cite{denouden2018improving} and \cite{zhou2022rethinking} might be constrained by the inherent limitations of autoencoders in capturing the full complexity of data distributions. On the other hand, \citeauthor{yang2022out}'s work in reliance on semantic synthesis could face challenges in scenarios where semantic information is ambiguous or not well-defined.
    
    In conclusion, these studies collectively advance our understanding of OOD detection using output reconstruction. They highlight the importance of considering both the reconstruction capability and the semantic content of the data, paving the way for more robust machine learning systems capable of handling the unpredictability of real-world data distributions.

\section{Discussion}

    The research on distribution shifts mentioned above has been rather limited, and the definition of robustness varies significantly depending on the application's sub-field. However, it is crucial to acknowledge that a robust model should encompass all the aforementioned categories. Real-world deployed models may encounter challenges like spurious correlations, perturbation attacks during testing, or unseen corruption shifts. Surprisingly, there have been no survey papers yet that comprehensively summarize and integrate all these distribution shift categories into a unified framework. Hence, this survey is pioneering in its aim to serve this purpose.
    
    Recent data-centric approaches published in the last couple of years (as shown in Table~\ref{tab:selected-articles}) have shown promise and provided valuable perspectives, as highlighted in this survey. We believe, it is essential for researchers to explore and extend their investigations to incorporate all three forms of robustness into a comprehensive framework.

    By delving deeper into the integration of robustness in a unified framework, new AI algorithms can make substantial progress in tackling distribution shifts and ensuring the reliability of AI systems across various applications. This notion of robustness and its critical role in the functionality of AI systems is exemplified in the development and application of advanced models such as the Dalle \cite{shi2020improving} \& GPT \cite{openai2023gpt4} series. The journey and implications of these models, driven by a data-centric AI approach, offer insightful perspectives on both the power and challenges inherent in modern AI technologies.
    
    \paragraph{GPT's Success and the Implications of Data-Centric AI} \hfill

    The GPT series \cite{openai2023gpt4} of language models are based on a deep learning architecture called a Transformer. These models have been trained using large amounts of textual data in a process called self-supervised learning. By training on a vast amount of data, the models can learn patterns and relationships in language that allow them to perform a range of natural language processing tasks, such as language translation and text summarization. The key to the success of these models is the use of data-centric AI. This approach involves collecting, preprocessing, and curating large amounts of data from diverse sources to ensure that the models have a broad and representative understanding of language. This data-centric approach has allowed the GPT series to achieve state-of-the-art performance on a range of language tasks.
    
    However, the use of data-centric AI is also relevant to the robustness and distribution shift of deep learning models. Deep learning models are known to be sensitive to changes in the distribution of the data they are trained on. If a model is trained on a particular type of data and then deployed in a different context with different data, its performance can degrade significantly. To mitigate this issue, data-centric AI can be used to ensure that the models are trained on a diverse range of data that covers a wide variety of contexts and situations. By doing so, the models can learn to generalize better and be more robust to distribution shifts. Additionally, by continuously updating the training data, models can be kept up-to-date with the latest developments in language and remain relevant over time.

    The data-centric approach used in models like GPT, while effective, introduces certain risks and challenges that highlight the importance of robustness in AI applications. This is particularly evident in healthcare settings. One such application addresses the complicated case of ethically using large language models in healthcare and medicine' \cite{Harrer2023-gr} discusses ethical concerns regarding the impact of LLMs on patient care. Additionally, BiomedGPT \cite{zhang2023biomedgpt} reveals performance limitations in the ChestXRay dataset due to spurious correlations causing hidden stratification. The recent exploration in `Extracting Training Data from ChatGPT' \cite{nasr2023scalable} illustrates the potential vulnerabilities in handling sensitive data. The concept of extractable memorization indirectly touches on distribution shift challenges. Extractable memorization can be seen as a form of unintended data leakage that may lead to model biases or vulnerabilities, potentially impacting the model's performance when faced with distribution shifts. This issue underscores the need for robust models that can maintain performance and reliability despite potential data distribution shifts, especially in the context of adversarial scenarios.
    Similarly, ChatGPT and Physicians’ Malpractice Risk'' \cite{10.1001/jamahealthforum.2023.1938} illustrates the legal and ethical complexities in medical applications, pointing to the necessity of models being robust against adversarial distribution shifts, where slight perturbations or misrepresentations in data could lead to significant consequences. Furthermore, ``Challenging the appearance of machine intelligence: Cognitive bias in LLMs'' \cite{talboy2023challenging} connects to the challenge of unseen distribution shifts, emphasizing the importance of unbiased and reliable AI applications capable of performing accurately in novel or unexpected scenarios. These studies collectively highlight the imperative of integrating robustness into AI models, addressing ethical, reliability, and bias aspects, especially in sensitive and high-stakes environments. This integration not only aligns with the broader objectives of our research but also reinforces the essential role of robustness in ensuring the safe and effective deployment of AI systems across various domains.
    
    Overall, the success of the GPT series of language models is due to the use of data-centric AI, which allows the models to be trained on a diverse and representative set of data. This approach is also relevant to the robustness and distribution shift of deep learning models more broadly, as it can help ensure that models are trained on a wide variety of data and are able to generalize well to new contexts.

\section{Conclusion}
\label{conclusion}

    Our survey has provided a comprehensive analysis of the various types of distribution shifts and their impact on computer vision models in real-world deployments. We've bridged the gap between theoretical assumptions and practical applications, shedding light on the critical role of data-centric methods in enhancing model robustness. Our unique contribution lies in the detailed comparison of diverse AI models and their approaches to handling different distribution shifts. This survey not only summarizes the latest innovations but also critically examines the limitations of existing models, offering new perspectives on robustness in AI. We've identified gaps in current research and proposed future directions, both long-term and short-term, to guide ongoing developments in this rapidly evolving field.

\section{Future Work}
\label{future-work}

    As we look ahead, the rapidly evolving landscape of AI and machine learning presents high-yielding ground for future research. Our study has laid a foundation, but much more remains to be discovered, especially with the recent rapid growth in AI driven by developments like Large Language Models (LLMs). We suggest a selected number of important research areas to further understand and improve AI robustness and how it handles changes in data.

    \textbf{Long-term Future Works}: \hfill
    \paragraph{\underline{Exploring Hallucinations in Vision-Language Models}:}~ Addressing the phenomenon of hallucination in vision-language models (VLMs) represents a significant future research direction. Hallucination in VLMs refers to the generation of incorrect or unrelated output, such as describing objects that are not present in an image. This problem directly impacts the robustness of models and their ability to handle subpopulation distribution shifts, especially in tasks like image captioning or visual reasoning.

    A starting point for this research could be inspired by approaches like the one in LURE \cite{zhou2023analyzing}, which offers a post-hoc algorithm to correct hallucinations. Future work could extend this by exploring preemptive strategies for minimizing hallucination during the training phase itself, enhancing model accuracy and reliability. Investigating the root causes of hallucination, such as data biases or model architecture limitations, will be crucial. This area of research not only improves model robustness but also ensures more reliable and trustworthy AI applications in diverse real-world scenarios. Some other efforts have been made on reducing object hallucination post training phases \cite{li2023m, liu2023aligning, liu2023visual}.

    \paragraph{\underline{Enhancing AI with Human-Centric UIs}:}~ A critical direction for future research involves developing user interfaces (UIs) that enhance human interaction in data-centric AI systems. Such UIs can play a pivotal role in integrating expert feedback to fine-tune AI models, particularly in complex tasks. Studies like \cite{carrera2023structured} that enable adaptive user interfaces and \cite{wu2022survey} illustrate the potential of UIs in augmenting AI robustness and handling distribution shifts. These interfaces can provide insightful human input. Further exploration in this domain may include designing intuitive UIs that incorporate expert feedback, thereby enhancing the model's resilience against unseen data samples or OOD variations and ensuring more reliable AI applications.

    \paragraph{\underline{Addressing Spurious Correlation widely}:} Spurious patterns where models incorrectly associate unrelated variables, continue to pose challenges across various domains, not just in medical imaging. The presence of spurious correlations can lead to biased decisions and reduced model robustness, especially in cases of subpopulation shifts. Therefore, broadening the study of these correlations beyond specific areas like medical imaging is crucial for developing more generalizable and robust AI systems.

    \paragraph{\underline{Compound Distribution Shifts: To attain universal robustness}:}~ The concept of compound distribution shifts, as discussed in \cite{dan2021statistical}, presents a complex yet realistic scenario where models encounter shifts due to a combination of factors. This type of shift, common in real-world applications like healthcare, challenges the transferability of fairness properties and robustness of models across different settings \cite{schrouff2023diagnosing}. For instance, new medical devices or different healthcare environments can introduce compound shifts, affecting model performance and fairness. These shifts often simultaneously impact various aspects of the data distribution, underscoring the limitations of algorithmic solutions that assume isolated shifts \cite{schrouff2023diagnosing}.

    Thus, future research should focus on understanding and addressing compound shifts, considering their interaction with fairness and robustness \cite{schrouff2023diagnosing}. This involves not only developing new algorithms but also rethinking the entire machine learning pipeline to ensure robust and fair performance across diverse and shifting real-world environments. Such an interdisciplinary approach, integrating insights from causal inference, fairness, and robustness, is crucial for creating universally robust AI models that can adapt to the multifaceted nature of real-world data \cite{schrouff2023diagnosing}.


    \textbf{Short-term Future Works}:\hfill

    \paragraph{\underline{Refining Distributionally Robust Optimization}:}~ Exploring methods to enhance the robustness of Distributionally Robust Optimization (DRO) against dataset outliers. One approach could be the development of preprocessing techniques for outlier removal, similar to iterative trimming. However, compared to such methods, a more dynamic solution like DORO (Distributionally Robust Optimization with Outlier Robustness) \cite{DORO} offers the advantage of not discarding any data and being adaptable to online data streams. Research in this area would focus on refining DRO \cite{DRO-example-1} to handle real-time data effectively, without the need for retraining models.

    \paragraph{\underline{Algorithm Design for Subpopulation Shifts}:}
    Building upon the principles of DORO, future work should delve into expanding these concepts across a wider array of algorithms that address subpopulation shifts. The focus could be on innovating within the realm of risk function refinement in techniques like static re-weighting \cite{shimodaira2000improving}, adversarial re-weighting \cite{lahoti2020fairness}, and group DRO \cite{GDRO}. The objective is to cultivate algorithms that are not only resilient to outliers but also capable of adjusting to unpredictable data shifts. One novel direction could involve the integration of real-time data analytics and AI-driven predictive models. These models could dynamically adjust the weighting mechanisms in response to emerging data trends and shifts. This approach aims to foster algorithms that are not just reactive but proactively adapt to changing data landscapes. The ultimate goal is to pave the way for the development of machine learning models that are robust, versatile, and capable of consistent performance across a spectrum of real-world scenarios. This exploration holds the potential to revolutionize the adaptability and applicability of machine learning models in diverse and evolving environments.

    \paragraph{\underline{Addressing decision-boundary bias}:}~ This research area aligns with the challenge of Adversarial Distribution Shift. As explored in \cite{utrera2021adversariallytrained}, the bias in decision boundaries in adversarially-trained models showcases a key aspect of this distribution shift. Future efforts should focus on developing strategies to counteract biases induced by adversarial manipulations. This includes innovating in training methodologies and regularization techniques that specifically target the biases in decision boundaries. By doing so, the aim is to enhance the models' fairness and equity, especially when subjected to adversarial scenarios. This work is pivotal in ensuring that AI systems remain reliable and unbiased in the face of adversarial distribution shifts.

    \paragraph{\underline{Model Selection in Domain Oblivious setting}:}~ Directly related to Unseen Distribution Shift, this future work is inspired by the insights from \cite{gulrajani2020search}. It revolves around optimizing model selection in situations where the test domain's distribution remains unknown during the training phase. This research should concentrate on developing adaptive algorithms and theoretical frameworks tailored for domain-generalization. The ultimate objective is to elevate the adaptability and generalizability of machine learning models, enabling them to perform reliably across a range of unforeseen and diverse real-world settings. By addressing this aspect of unseen distribution shifts, the research will contribute significantly to the robustness and versatility of AI applications in domain-agnostic environments.

    \paragraph{\underline{Addressing Complexity and Specificity in AI for Enhanced Robustness}:}~ As AI systems increasingly encounter complex and varied data, we see a need to develop models that can efficiently process this diversity while avoiding over-reliance on misleading features that might cause spurious effects. One way this challenge can be handled is by applying a dual-focused approach: firstly, addressing the issue of overparametrization \cite{overparameterization} by developing regularization techniques or model architectures that discern between genuine and spurious correlations. This might involve novel training methodologies that penalize overreliance on misleading features, and secondly, enhancing the ability of these models to interpret and learn from instance-specific data labels, as per the principles of LISA \cite{LISA}. This could be achieved by integrating adaptive annotation techniques that allow models to learn from label-specific nuances, potentially through tailored loss functions or instance-specific learning rates. Future research could concentrate on creating models that integrate the aforementioned aspects. The aim will be to develop systems that are both adaptable and accurate across different data contexts. This approach promises AI models that are not only robust against misleading correlations and distribution shifts but are also equipped to handle diverse, real-world data effectively.


\bibliographystyle{ACM-Reference-Format}
\bibliography{references} 

\end{document}